\documentclass[acmsmall]{acmart}

\usepackage{graphicx}
\usepackage{svg}
\usepackage{multirow}
\usepackage{makecell}
\usepackage{multirow}
\usepackage{amsmath}
\usepackage{breqn}
\usepackage{lineno,hyperref}
\usepackage{rotating}
\usepackage{longtable}
\usepackage{makecell}
\modulolinenumbers[5]
\hypersetup{pdfauthor={Name}}

\newcommand*\rot{\rotatebox{90}}

\AtBeginDocument{%
  \providecommand\BibTeX{{%
    \normalfont B\kern-0.5em{\scshape i\kern-0.25em b}\kern-0.8em\TeX}}}

\setcopyright{acmcopyright}
\copyrightyear{2021}
\acmYear{2021}
\acmDOI{10.1145/xxxxx.xxxxx}

\acmVolume{00}
\acmNumber{00}
\acmArticle{000}
\acmMonth{00}

\begin{document}

\title{Graph Neural Networks: Methods, Applications, and Opportunities}

\author{Lilapati Waikhom}
\authornote{Both authors contributed equally to this research.}
\email{lilapati\_rs@cse.nits.ac.in}
\author{Ripon Patgiri}
\orcid{0000-0002-9899-9152}
\authornotemark[1]
\email{ripon@cse.nits.ac.in}
\affiliation{%
  \institution{National Institute of Technology Silchar}
  \streetaddress{Department of Computer Science \& Engineering, National Institute of Technology Silchar}
  \city{Silchar}
  \state{Assam}
  \country{India}
  \postcode{788010}
}

\renewcommand{\shortauthors}{Waikhom and Patgiri}

\begin{abstract}
In the last decade or so, we have witnessed deep learning reinvigorating the machine learning field. It has solved many problems in the domains of computer vision, speech recognition, natural language processing, and various other tasks with state-of-the-art performance. The data is generally represented in the Euclidean space in these domains. Various other domains conform to non-Euclidean space, for which graph is an ideal representation. Graphs are suitable for representing the dependencies and interrelationships between various entities. Traditionally, handcrafted features for graphs are incapable of providing the necessary inference for various tasks from this complex data representation. Recently, there is an emergence of employing various advances in deep learning to graph data-based tasks. This article provides a comprehensive survey of graph neural networks (GNNs) in each learning setting: supervised, unsupervised, semi-supervised, and self-supervised learning. Taxonomy of each graph based learning setting is provided with logical divisions of methods falling in the given learning setting. The approaches for each learning task are analyzed from both theoretical as well as empirical standpoints.  Further, we provide general architecture guidelines for building GNNs. Various applications and benchmark datasets are also provided, along with open challenges still plaguing the general applicability of GNNs.
\end{abstract}



\maketitle


\section{Introduction}
A graph is a data structure defining a set of nodes and their relationships.  We observe them everywhere, starting from social networks \cite{qiu2018deepinf} to physical interactions  \cite{zuo2012network}. Graphs can also be used to represent inconceivable structures like atoms, molecules, ecosystems, living creatures, planetary systems \cite{duvenaud2015convolutional}, and so on. So, graph structures are found in our surroundings and perception of the world. It comprises entities and inter-relationships to establish concepts like reasoning, communication, relations, marketing, etc. 

As the technologies nowadays have become advanced, the Internet (a giant graph) usage is rapidly growing. Other massive graphs are also found nowadays in the social networks, knowledge databases of search engines, street maps, even molecules, high-energy physics, biology, and chemical compounds. A graph-structured representation in these environments is common; therefore, effective and novel techniques are required to solve graph-based tasks. Many traditional machine learning techniques have been proposed on top of the extracted features using various predefined processes from the original data form. The extracted features could be pixel statistics in image data or word occurrence statistics in natural language data. In the last decade, Deep learning (DL) techniques have gained massive popularity, tackling the learning problems efficiently, learning representation from raw data, and predicting using the learned representation simultaneously. Usually, this is accomplished by exploring many different non-linear transformations (performed by layers) and end-to-end training of such models using gradient descent-based learning methods. Even though DL has recently advanced in several fields such as computer vision, natural language processing, biomedical imaging, bioinformatics, and so on, it still lacks relational and causal reasoning, intellectual abstraction, and various other human abilities. Structuring the computations and representations in a deep neural network (DNN) in the form of a graph is one of the ways to address these problems, which is characterized as Graph Neural Network (GNN). 

\footnotesize
\begin{table*}[!ht]
    \centering
    \caption{Existing surveys on graph neural network and their comparison with our work.}
    \begin{tabular}{p{1.5cm}p{12cm}}
    \hline
          \textbf{Papers} & \textbf{Difference and novelty of ours article}
   \\ \hline
       
     Wu \textit{et al.} \cite{Wu} & Wu \textit{et al. } presented a new taxonomy by dividing existing GNNs into four categories: recurrent, convolutional, spatial-temporal, and graph autoencoders GNNs. Nevertheless, the paper has not explained each learning setting separately. On the contrary, our article presents the new taxonomies for each type of GNNs' learning settings. We further provide many more available datasets related to various fields.  \\ 
     
     R. Sato \cite{Sato} & R. Sato focuses more on the power of GNNs, and also presented a comprehensive overview of the powerful variants of GNNs, but has not focused on taxonomy. Therefore, we present not only the taxonomy but also diverse features of GNN. \\ 
     
     Zhou \textit{et al.} \cite{zhou2020graph}  &  Zhou \textit{et al.} provided a broad design pipeline of the GNNs and discussed each of the module's variants of GNNs. The article analyzed the GNNs both theoretically and empirically. The paper presented applications of GNNs by dividing them into structural scenarios and non-structural scenarios. The paper also introduced four open problems of GNNs and gave future directions too. But again, the paper has not provided a separate taxonomy for each of the learning settings. Our article provides the new taxonomies for each type of GNNs' learning settings. We introduced various applications of GNNs with the existing works, and we also discussed several datasets for GNNs in many different domains.  \\ 
     
     Abdal \textit{et al.} \cite{Abadal}  & Abdal \textit{et al.} presented a comprehensive review of GNNs from the computing perspective. The paper also provided a detailed examination of current software and hardware acceleration schemes, from which a graph-aware, hardware-software, and communication-centric vision for GNN accelerators is derived. On the contrary, our paper focuses on different learning settings of GNNs by providing proper taxonomy for each of the setting types. \\ \hline
     
    \end{tabular}
    \label{tab1}
\end{table*}
\normalsize

GNNs are successful on graph-structured datasets in various domains with many learning settings: supervised, semi-supervised, self-supervised, and unsupervised. Most graph-based methods fall under unsupervised learning and are often based on Auto-encoders, contrastive learning, or random walk concepts. A few recent works on the graph Auto-encoders are: feature extraction in hyperspectral classification by Cao et al. \cite{cao2020unsupervised}; for preventing over smoothing of message passing by Yang et al. \cite{yang2020toward}; using message passing Auto-encoders for hyperbolic representation learning by Park et al. \cite{park2021unsupervised};  for addressing a limitation of current methods for link prediction by Wu et al. \cite{wu2021deepened}. Recently contrastive learning-based methods are also successful, as shown in many works done by researchers. Okuda et al. \cite{okuda2021unsupervised} is a recent unsupervised graph representation learning for discovering common objects and localization method for a set of particular objects in images. Random walks have been coupled with current representation learning methods for language modeling to provide exceptional representations of vertices. The learned representation can be utilized for downstream learning tasks like node classification and edge prediction as shown in Du et al. \cite{du2018dynamic}, and Perozzi et al. \cite{perozzi2014deepwalk}. Subgraph embeddings are also captured using expanded random walks in Adhikari et al. \cite{adhikari2018sub2vec}, and vertex representations in heterogeneous graphs as in Dong et al. \cite{dong2017metapath2vec}. 

This paper classifies the graph semi-supervised learning methods based on their embedding characteristics, such as shallow graph embedding and deep graph embedding. We divided the shallow Graph embedding as factorization, random walk, and the deep graph embedding as Auto-encoder embedding and GNN. A further explanation of each method is provided, along with the categories of the GNN. The graph-based self-supervised learning methods are classified based on pretext tasks and training strategies. Most of the existing survey papers in GNNs focus either on the single learning setting or on general GNNs, as shown in Table \ref{tab1}. These surveys have not explained each learning setting separately. One of the most recent works is done by Zhou et al. \cite{zhou2020graph} that focuses on various machine learning algorithms on graphs. 

In this paper, we explored each graph-based learning setting and divided it into several categories. The key contributions of the article are outlined below-
\begin{itemize}
    \item Basic terminologies and variants of graphs are defined along with various graph-based tasks.
    \item A comprehensive review of GNNs is presented. Our work focus on all the learning settings, contrary to various surveys that concentrate on a single learning setting.
    \item Further, each graph-based learning setting is explored and divided into required categories.
    \item A general guideline for designing GNN architectures is presented.
     \item We provide many GNN resources, including SOTA models, popular graph-based datasets, and various applications. 
    \item We analyze theoretical and empirical aspects of GNNs, assess the challenges of current techniques, and propose possible future research routes in terms of model depth, scalability, higher-order, and complex structures, and robustness of the techniques.
\end{itemize}

\noindent
\textit{Organization of the paper: } Section \ref{GNN} introduces the basic terminologies and concepts of GNN followed by variants of graph and tasks based on graph-structured data in Section \ref{variants} and Section \ref{tasks} respectively. Section \ref{methods} explains the GNN based methods for each learning setting and further breakdown of methods and learning settings into logical divisions. Section \ref{supervised} briefly explains the existing graph supervised learning methods. Graph-based unsupervised learning methods are explained in Section \ref{unsupervised} along with subdivision of the existing techniques in terms of learning methods. Then we present the graph semi-supervised learning methods in Section \ref{semi} along with subdivision of the methods by the embedding methods. Section \ref{self} explains the graph self-supervised learning methods, dividing each method in terms of pretext tasks and the training strategies. The general step-wise architecture of the GNN is given in Section \ref{general}. Section \ref{analysis} analyzes the GNN methods in both theoretical and empirical aspects. We present in Section \ref{data} several commonly available datasets used in research for GNNs followed by Section \ref{application}, presenting a few popular applications of GNN. Section \ref{open} summarises the unresolved issue still plaguing the GNN based solutions for graph-based tasks. Finally, in Section \ref{conc}, we conclude this work.

\section{Graph Neural Networks}
\label{GNN}


Graph Neural Network (GNN) is a type of DNN that is suitable for analyzing graph-structured data. The mathematical notations used by us throughout this article are given in Table \ref{tab2}. A graph is a set of $V$ vertices (nodes) and a set of $E$ edges (links).  Using the notation for vertices and edges, a graph is represented as $G=(V, E)$, where a vertex $v_i \in V$ and a directed edge between $v_i$ and $v_j$ is represented by an arrow as $i \rightarrow j$, it forms an ordered pair of nodes $(v_i, v_j) \in V \times V$. Undirected edges are assumed to have equal weightage from both directions. An instance of a graph $G$  where nodes are associated with feature vectors $ x_i $ with or without edges associated with feature vectors $ x_{(i,j)} $ is fed to a GNN as the inputs. $ h_i $ and $ h_{(i,j)} $ denote the hidden representations in the neural network for nodes and edges, respectively. As an initial node representation, we can use $ h_i = x_i $. The message passing updates are dictated by the structure of the graph $G$, which are carried out to get updated node and edge representations $ h'_i $ and $ h'_{(i,j)}$.

\begin{equation}
\label{eq1}
 h_{(i,j)} = f_{edge}(h_i, h_j, x_{(i,j)}) 
\end{equation}

\begin{equation}
\label{eq2}
h'_i = f_{node}(h_i,\sum_{j \in Ni} h_{(j,i)}, x_i)
\end{equation}

Where $ N_i $ is the set of neighboring nodes with an incoming edge to $i_{th}$ node, and $f_{node}$ and $f_{edge}$ are two or three-layer Multi-layer Perceptrons (MLPs) that accept a concatenation of the function parameters as input, although other various options are also possible. By changing $ h_i \leftarrow h'_i $ after each node update defined by Equation \ref{eq2}, multiple message passing updates can also be chained. message passing updates do not require the shared $ f_{edge}$ and $ f_{node} $ parameters. 


Kipf and Welling \cite{kipf2016semi} generalize and unite prior models, known as the GCN (Graph convolution network) or the interaction network presented by Battaglia et al. \cite{battaglia2016interaction}, this version of GNN was presented by Gilmer et al. \cite{gilmer2017neural} with the name message passing neural network. Gori et al. \cite{gori2005new} typically gave the first GNN model. The model incorporates many basic principles like a recurrent neural network trained in back-propagation over time and identified a contraction map to get an on-demand general description of GNNs. Moreover, an explicit edge representation $ h_{(i,j)}$ is not learned by this type of GNNs, and the node update function $f_{node}$ was based on adjacent states $ h_{j,i} $ and $ j \in N_i $ (along with initial node feature vectors $ x_i $). Scarselli et al. \cite{scarselli2008graph} expanded the work of updating intial edges $ x_{(i,j)} $ by conditioning the messages.

\subsection{Variants of Graph}
\label{variants}
Various variants of the graph are found in nature and science. Undirected graphs are the most commonly found type. Based on the structure and scale of the graph type present in the data, problem settings are defined. Below we present the variants of graphs.
\subsubsection{Directed Graph}
An undirected edge indicates that the two nodes have a link with no directional information, but a directed edge can provide more information about the relationship between the nodes. For example, suppose we have a class hierarchy. In that case, we can describe that class hierarchy-data using a directed graph, with the head representing the child and the tail representing the parent, or vice versa.

        

\subsubsection{Heterogeneous Graph}
Heterogeneous graphs are the types of graphs that consist of different node types. The computations in this type of graph are achieved by altering the data using one-hot encoding to make each node representation identical. The ability of various nodes, the method Graph Inception was created, which took use of this feature by grouping and clustering different neighbors to be utilized as a whole. These clusters, also known as sub-graphs, are being used to do parallel calculations. The Heterogeneous Graph Attention Networks \cite{linmei-etal-2019-heterogeneous} have been developed with the same heterogeneous characteristic in mind.

\subsubsection{Dynamic Graph}
The dynamic graph structure is a type of graph with changes over time, and their inputs might be dynamic. The node and edge are updated in this type of graph. Nodes are added or deleted, and the corresponding edges between the nodes are either created or updated. This allows the adaptive structures or algorithms that require internal structures to be dynamic to be applied to graphs. 

\subsubsection{Attributed Graph}
Edges in Graphs with edge information include additional information such as weights or type of edges. This knowledge can aid in the development of architectures such as G2S encoders and Relational-GCN (R-GCN). Particularly in R-GCN, which is a GCN extension, here, R stands for relational. As a result, having a graph with edges that can store extra information, such as the relationship between nodes, becomes more manageable when working with relational data.

\subsection{Classification of Tasks based on hierarchies of Graph-Based data}
\label{tasks}
\begin{figure}[!ht]
    \centering
    \includegraphics[width=0.4\textwidth]{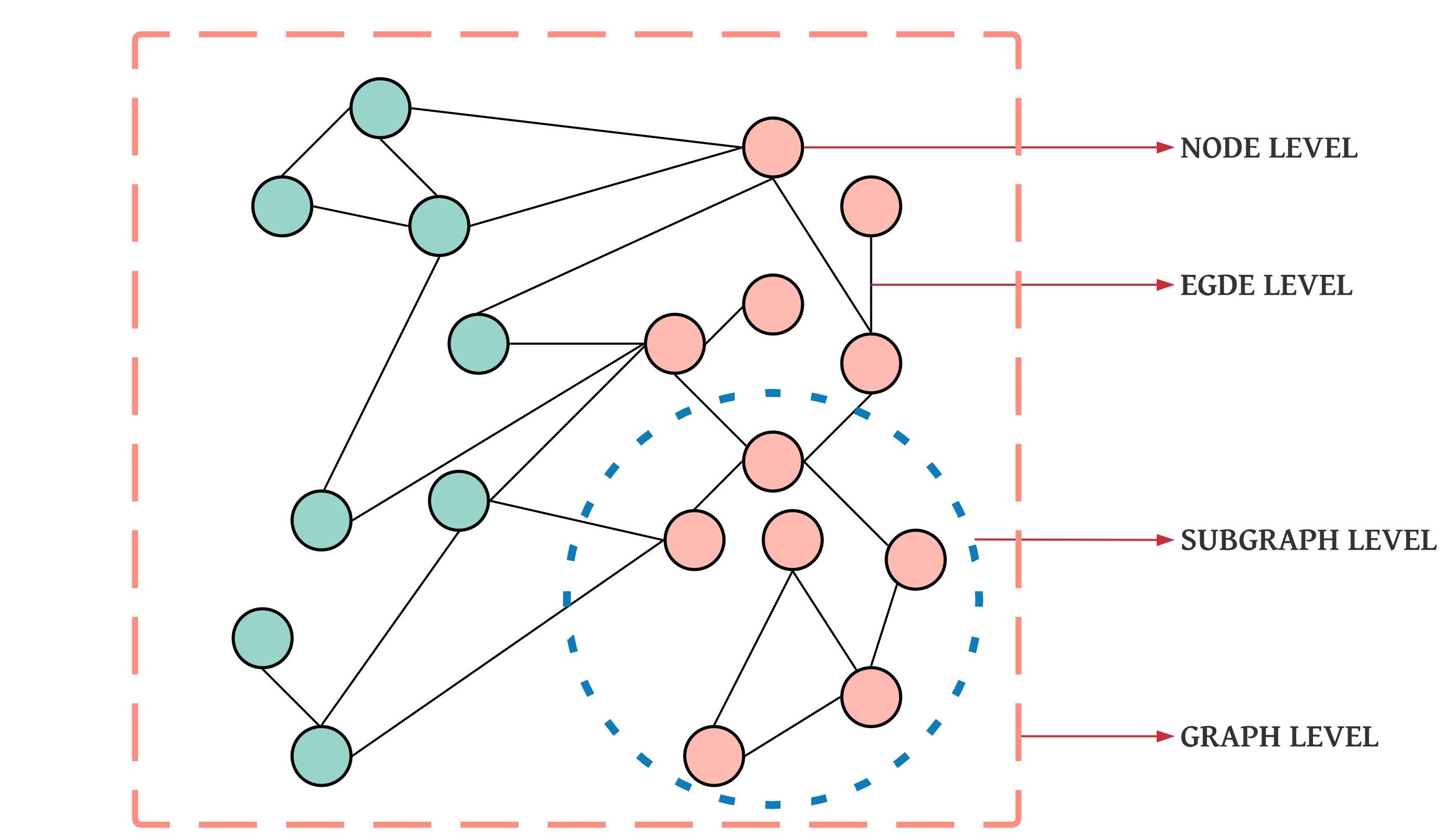}
    \caption{Types of tasks in Graph. Node level based tasks works on vertices, Edge level based tasks works on relationship among the vertices, Subgraph level tasks forms different set of smaller graphs within a given graphs and Graph level tasks works on independent graphs.}
    \label{fig1}
\end{figure}

Graph-based data has knowledge embed at various hierarchies of the structure shown in Figure \ref{fig1}. At the node level, various node-based tasks are defined. Similarly, edge-level tasks are defined. Further, graph-level tasks encompassing whole graphs or sub-graph are also defined based on various applications. 
\subsubsection{Node Level Task}
Tasks like node classification, node clustering, node regression, etc., focus on node-level taxonomy. Node classification attempts to classify nodes into different groups, whereas regression task for nodes predicts a real value for each node. The node clustering intends to divide nodes into many distinct classes, with related nodes are grouped together. One famous example of node-level tasks is the protein folding problem, where amino acids in a protein sequence are treated as the nodes and proximity between amino acids as the edges. By using information propagation/graph convolution, recurrent GNNs and convolutional GNNs can extract high-level node representations. GNNs can execute node-level tasks in an end-to-end manner using a multi-perceptron or softmax layer as the output layer.

\subsubsection{Edge Level Task}
The link prediction and edge classification tasks are edge-level tasks. In short, link prediction and edge classification are the tasks that require the model to predict whether or not there is an edge between two nodes or categorize edge types. One of the primary examples of edge-level tasks is the recommendation system (recommend the items users might like). Here, items and users act as the nodes and user-item interactions as the edges. Another important example is biomedical graph link prediction. Here, drugs and proteins represent the nodes, and interactions between them represent the edges. The task is such as how likely Simvastatin and Ciprofloxacin will break down muscle tissue. Using the hidden representations of two nodes from GNNs as inputs, a similarity function or a DNN is used to determine the connection strength of an edge.

\subsubsection{Graph Level Task}
Graph classification, regression, and matching tasks require graph-level representation to be modeled. The graph classification task is linked to graph-level results. The popular graph level tasks are a) Drugs discovery (such as antibiotics are treated as the small molecular graphs). Here, atoms represent nodes and chemical bonds between them as the edges. b) Physics simulation: Particles as the nodes and interaction between particles as the edges, etc. The graph level tasks can be of subgraph level. One of the most famous examples of subgraph level tasks is the traffic prediction by considering road network as a graph; road segments represent the nodes and connectivity between road segments as the link/edges. GNNs are frequently coupled with pooling and readout operations to get a compact representation on the graph level.

\section{Methods of Graph based Deep learning}
\label{methods}
We observe from the literature and divide graph-based learning tasks into three distinct training settings from the perspective of supervision: graph supervised learning, graph Unsupervised learning, and graph semi-supervised learning.

\subsection{Graph-Based Supervised Learning}
\label{supervised}
The present prevailing graph construction techniques are unsupervised during the building phase, i.e., unsupervised learning does not employ any specific label information. However, the labeled samples are utilized to improve the graph created for downstream learning tasks as a type of prior knowledge. Dhillon et al. \cite{dhillon2010learning} investigate the use of labeled points to determine the similarities between node pairs. Rohban et al. \cite{rohban2012supervised} provide another supervised technique of graph building, demonstrating that the best solution for a neighborhood graph is considered as a subgraph of a KNN graph as long as the manifold sampling rate is big enough. Based on earlier research presented by Ozaki et al. \cite{ozaki2011using}, a novel technique, Graph-based on the informativeness of labeled instances (GBILI), is proposed by Berton et al. \cite{berton2014graph}, which also uses the label information. GBILI achieves a reasonable classification accuracy level, but it also has quadratic time complexity. Furthermore, Lilian et al. \cite{berton2017rgcli} have improved the method for creating more resilient graphs by addressing an optimization issue with the Robust graph that considers labeled instances (RGCLI) algorithm, which is based on GBILI \cite{berton2014graph}. Recently, a low-rank semi-supervised representation by Zhuang et al. \cite{zhuang2017label} has been suggested as a novel semi-supervised learning technique that includes labeled data into the Low-rank representation (LRR). The produced similarity graph can significantly aid the subsequent label inference process by including extra supervised information.

\subsection{Graph-Based Unsupervised Learning}
\label{unsupervised}
In supervised and semi-supervised learning settings, the ground truth labels are present along with the corresponding data samples. There are no labeled samples in the unsupervised setting; thus, loss functions must rely on information extracted from the graph itself, such as node, edge attributes, and graph topology. In this part, we mostly describe unsupervised training variations. Most graph unsupervised learning methods are often based on the concepts of contrastive learning, Auto-encoders, or random walk, as shown in Table \ref{tab4}. 

\subsubsection{Graph-Based Auto-encoders}
For the graph-structured data, Kipf and Welling \cite{kipf2016variational} used the extended Auto-encoder and called it Graph Auto-encoder (GAE). GAE \cite{kipf2016variational} gets the initial representation of nodes using GCNs. The training is done using the loss computed using the similarity between the reconstructed adjacency matrix and the original one. The learning process follows the variational style of training, called Variational graph Auto-encoder (VGAE).  Wang et al. \cite{wang2017mgae}, and Park et al. \cite{park2019symmetric} attempt to reconstruct the feature matrix rather than the adjacency matrix. To provide resilient node representation, Marginalized Graph Auto-encoder (MGAE) \cite{wang2017mgae} use a marginalization denoising Auto-encoder. Park et al. \cite{park2019symmetric} proposed a technique called Graph convolutional Auto-encoder using Laplacian smoothing and sharpening (GALA), which is a Laplacian sharpener that performs the inverse operation of smoothing and decode hidden states to create a symmetric graph Auto-encoder.

\subsubsection{Contrastive Learning}
In addition to GAE, contrastive learning is used for graph representation learning in the unsupervised learning setting.  Velickovic et al. \cite{velickovic2019deep} proposed Deep Graph Infomax (DGI), which is extended from deep InfoMax presented by Hjelm et al. \cite{hjelm2018learning}. DGI maximizes the mutual information across graph and node representations. Infograph presented by Sun et al. \cite{sun2019infograph}, maximizes the mutual information between graph level representations and subgraph level representations of various sizes, such as nodes, edges, and triangles, to learn graph representations. Hassani and Khasahmadi's multi-view \cite{hassani2020contrastive} compares first-order adjacency matrices representation with graph diffusion, achieving SOTA results on numerous graph learning challenges.  Okuda et al. \cite{okuda2021unsupervised} is a recent unsupervised graph representation learning for discovering common objects and localization method for a set of particular object images. 

\subsubsection{Random Walk}
The random walks have been proven to be scalable for the large networks to capture the graph structure efficiently by Perozzi et al. \cite{perozzi2014deepwalk} proposed method call Deepwalk. Moreover, random walks were demonstrated to be capable of compensating structural equivocation (vertices with comparable local structures with similar embeddings) and equally (vertices with similar embedding belonging to the same communities) \cite{du2018dynamic}. Random walks have been coupled with current language modeling representation learning methods to provide high-quality representations of vertices are utilized for downstream learning tasks like vertex and edge prediction as shown in Du et al. \cite{du2018dynamic}, and Perozzi et al. \cite{perozzi2014deepwalk}. In addition, random walk-based techniques have been expanded to capture subgraph embeddings in Adhikari et al. \cite{adhikari2018sub2vec}, and vertex representations in heterogeneous graphs as in Dong et al. \cite{dong2017metapath2vec}.  

\footnotesize
\begin{longtable}{p{.8cm}p{1.5cm}p{1.5cm}p{1cm}p{7cm}}
    \caption{Summary of existing GNNs based unsupervised learning methods. Abbreviations: NC=Node Classification, LP=Link Prediction, GC=Graph Classification, GCN=Graph convolutional network, NN=Neural Network, GAT=Graph Attention Network, GAE=Graph Auto-encoder, RW=Random walk.}\\
    \hline
    \textbf{Paper} & \textbf{Feature extraction} & \textbf{Technique} &   \textbf{Task} & \textbf{Key functionality}\\ \hline
     \endfirsthead
     \multicolumn{5}{c}{{Table \thetable.\ Continued from the previous page}} \\
\hline
\centering 
 \textbf{Paper} & \textbf{Feature extraction} & \textbf{Technique} &   \textbf{Task} & \textbf{Key functionality}\\ \hline
\endhead
     
     
    
    \cite{okuda2021unsupervised} &  RW & CNN & NC  & Discover common object and localization method for a set of particular object images. \\ 
     
    \cite{velickovic2019deep} &  Contrastive & CNN &  NC & Maximizing the mutual information between the graph representations and node representations. \\ 
     
    \cite{sun2019infograph} & Contrastive & K-layer GCN & NC, LP, GC & Graph-level representations. \\ 
     
    \cite{hassani2020contrastive}& Contrastive & GCN & NC,GC & Learning node and graph level representations. \\ 
     
    \cite{hjelm2018learning} & Contrastive & NN & NC & New avenue for unsupervised learning of representations. \\ 
     
     
    \cite{tang2015line} & RW & NN & NC, LP &  Low dimensional node embeddings in the huge graphs. \\ 
     
     
    \cite{hamilton2017inductive} & RW & GCN \& LSTM & NC, GC &  Low dimensional node embeddings in the huge graphs. \\ 
     
     
     
    \cite{dong2017metapath2vec} & RW & NN & NC & Node Representation Learning for Heterogeneous Networks. \\
     
    \cite{adhikari2018sub2vec} & RW & NN & NC & Formulate subgraph embedding problem. \\ 
     
    \cite{park2021unsupervised} & GAE & NN & NC, LP & Hyperbolic Representation Learning via Message Passing Auto-encoders. \\ 
     
      
       
         
    \cite{li2021unsupervised} & GAE & NN & NC & Learning prerequisite chains in both the known and unknown domains for acquiring knowledge \\ 
     
    \cite{kipf2016variational} & GAE & GCN & LP & Learning the interpretable latent representations for the undirected graphs.  \\ 
     
    \cite{PanAdversarially} & GAE & GCN & LP, GC & Representing graph-structured data in a low dimensional space for graph analytics. \\ 
     
    \cite{wang2017mgae} & GAE & GCN & GC & Marginalized graph Auto-encoder algorithm for graph clustering. \\ 
     
    \cite{park2019symmetric} & GAE & GCN & NC, LP, GC & Extracted low-dimensional latent representations from a graph in irregular domains. \\ 
     
    \cite{CuiAdaptive} & GAE & GCN & NC, LP & Graph embedding for learning the vector representations from node features and graph topology. \\
     
    \hline

    \label{tab4} 
\end{longtable}
\normalsize

\subsection{Graph-Based Semi-Supervised Learning}
\label{semi}
Semi-supervised learning has been around for many years. It applies to the scenario where only a few labeled samples are available, and the rest of the data samples are unlabeled. Actual labels of labeled data samples are computationally heavy to compute, so unlabeled data samples have to be utilized using novel ways to solve the problems. The manifold assumption in this field of study suggests that nodes closer to each other in the low-dimensional manifold are similar and should have the same label. Over the years, many methods are employed to do semi-supervised learning. Graph-based semi-supervised learning is an emerging subfield that is a good fit as graph structure fits manifold assumptions in semi-supervised learning. The node represents the data samples in graph-based semi-supervised learning, and edges give the similarity between the nodes. Nodes having large edge weights represent high similarity and belong to the same label class, corresponding to the manifold assumption. Graph structures are intuitive to understand and highly expressive, leading to success in graph semi-supervised learning-based methods falling under the manifold assumption. Several semi-supervised survey articles \cite{prakash2014survey} are focused on traditional ways of dealing with semi-supervised settings. A few of the new works, such as \cite{van2020survey} study graph construction and graph regularization, which focuses on a general overview of semi-supervised learning. We focus on the advances in graph semi-supervised learning in this section, specifically advances in graph embedding. Table \ref{tab5} shows the most recent methods for graph semi-supervised learning.

\subsubsection{Graph Embedding}
In graph-based semi-supervised learning methods, there are two levels of embedding that are observed. One is for the entire graph, and the second is for a single node \cite{hamilton2017representation}. The objective of both embeddings is to represent the given object in a low-dimensional space. Node embedding is generally the focus of graph-based semi-supervised learning tasks. It aims to represent nodes in a low-dimensional space where the local structure of nodes is preserved. Given a graph $G = (V,E)$, the node embedding on graph G is denoted as a mapping $h_{\textbf{z}}:p\rightarrow\textbf{z}_p \in \mathbb{R}^d, \forall p \in V$ such that $d \ll |V|$. The proximity measure for nodes in graph G is preserved by $h_{\textbf{z}}$. Graph embedding methods' loss function can be defined by the generalized Equation \ref{eq3}:

\begin{equation}
\begin{split}
    \mathbb{C}(f) & = \sum_{(x_i, y_i) \in \mathbb{D}_i} \mathbb{C}_s(h(h_{\textbf{z}}(x_i)), y_i) 
    + \mu \sum_{x_i\in \mathbb{D}_i + \mathbb{D}_u} \mathbb{C}_r(h(h_{\textbf{z}}(x_i)))
\end{split}
\label{eq3}
\end{equation}

Here $h_{\textbf{z}}$ represents the embedding function. This function is similar to the one used in graph regularization. The only difference is that models are trained using node embedding results instead of node attributes in graph regularization. All the graph embedding methods are generalized under the framework of encoder-decoder. The encoder part generates low dimensional embeddings of input nodes, while the decoder reconstructs the original information related to each node from the embeddings.

\textit{Encoder: }
The encoder can be formally regarded as mapping of $p \in V$ nodes into $z_p \in \mathbb{R}^d$ vector embedding functions. The generated embeddings in the latent space with additional dimensions are more discriminatory. In addition, the following decoder module is easier to convert back to the original function vector. We have encoder: $V \rightarrow \mathbb{R}^d$ from a mathematical point of view.

\textit{Decoder: }
The principal aim of the decoder module is to rebuild specific graph statistics from the node embedded in the previous phase. For example, if $z_u$ is a node embedding of a node $u$, the decoder might try to predict the neighbor $ \mathit{N}(u) $ or its row $ A[u] $ in the adjacency matrix of that node. Often decoders are defined in a pair form that is shown to predict the similarity of each node. We have, Decoder: $ \mathbb{R}^d \times \mathbb{R}^d \rightarrow \mathbb{R}^+ $ from a mathematical form.

\begin{figure}[!ht]
    \centering
    \includegraphics[scale=0.25]{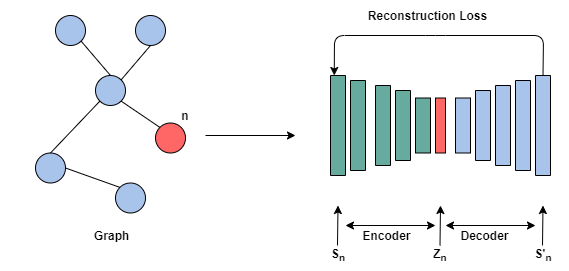}
    \caption{A high-dimensional vector $ S_n $ is extracted and fed into the Auto-encoder based methods for generating a low-dimensional $ z_i $ embedding.}
    \label{fig2}
\end{figure}

\subsubsection{Shallow Graph Embedding}
Many specific matrix factorization optimization methods are used as a determining means of solving the optimization issues. The entire task is generally taken to develop a low-dimensional approximation of an S-Matrix using matrix factorization techniques, where S encodes the information relating to the original adjacency matrix or other matrix measurements. In recent years, various methods have successfully developed that overlap stochastic neighborhood measures to provide low embedding, unlike the deterministic factorization techniques. The essential new feature is the node embedding which optimizes two network nodes to coincide with a high probability on short random walks, and the embeddings are also the same \cite{spitzer2013principles}.

\paragraph{Factorization: }A matrix that specifies the relationship among each node pair is factorized for the category of factorization-based techniques to get the node embedding. This matrix generally includes certain basic structural information, such as an adjacency matrix and a normalized Laplacian matrix, regarding the similarity graph, is built. Various matrix characteristics can lead to the factorization of these matrices in different ways. The normalized Laplacian Matrix is obviously positive semi-definite; hence, its eigenvalue decomposition is a natural match.

\paragraph{Random Walk: }The random walk is a valuable tool for obtaining approximation outcomes on specific characteristics of the particular graph, such as node centrality \cite{newman2005measure} and similarity \cite{fouss2007random}. Thus, random node embedding methods for particular circumstances are successful when just part of the graph is accessible or too wide for efficient management.

\paragraph{Limitations of Shallow Embedding: }Although shallow embedding methods have demonstrated remarkable performance on various semi-supervised tasks notably, shallow embedding also has certain significant disadvantages that researchers have found difficult to overcome; for instance, it can only generate embeddings for a single fixed graph. Moreover, it does not focus on nodes' features.

 \textit{Lack of shared parameters: }Few parameters are shared. Parameters are not shared among nodes in the encoder module, as the encoder directly creates a unique vector for each node. The lack of shared parameters means necessary to increase the number of parameters as $ O( \; | \; V \; | \; ) $, which are intractable in massive graphs.
 
  \textit{No use of node features: }Another critical issue in shallow embedding methods is the exclusion of the node features. However, the encoding procedure might possibly contain rich feature information. This applies notably to semi-supervised learning activities because each node provides significant information about features.
 
  \textit{Failure in inductive applications: }Methods of shallow embedding are intrinsically transductive \cite{hamilton2017representation}. It is not possible to generate embeddings for additional nodes discovered after the training phase. Shallow embedding methods can not be employed for inductive applications because of this constraint.
   
\subsubsection{Deep Graph Embedding}
Many deep embedding techniques have been developed in recent years to address the limitations highlighted above. The shallow embedding methods are different from the deep embedding ones. Here, the encoder module would take into account both the graph's structural and attribute information. A top-level classifier must be trained to predict class labels for unlabelled nodes under the transductive setup for semi-supervised learning tasks based on the node embeddings.

\footnotesize
\begin{longtable}{p{1cm}p{1.8cm}p{1.5cm}p{4cm}p{1.5cm}p{1.5cm}}
    \caption{Summary of existing GNNs based semi-supervised learning methods. Abbreviations: AE=Auto-encoder, Dec=Decoder, Fact= Factorization, RW=Random Walk.}\\
      \hline
    \textbf{Paper} & \textbf{Embeddi-ng Architecture} & \textbf{Embeddi-ng Methods} & \textbf{Loss Function} &  \textbf{Decoder} & \textbf{Similari-ty Measure} \\ \hline
     \endfirsthead
     \multicolumn{5}{c}{{Table \thetable.\ Continued from the previous page}} \\
\hline
\centering  
\textbf{Paper} & \textbf{Embeddi-ng Architecture} & \textbf{Embeddi-ng Methods} & \textbf{Loss Function} &  \textbf{Decoder} & \textbf{Similari-ty Measure} \\ \hline
\endhead
     
     \cite{roweis2000nonlinear} & Shallow & Fact & $\sum_p ||\textbf{z}_p - \sum_q \mathit{A}_{pq}\textbf{z}_q||^2$ & $\textbf{z}_p - \sum_q \mathit{A}_{pq}\textbf{z}_q$ & $A_{pq}$\\
     
    \cite{belkin2001laplacian} & Shallow & Fact & $\text{Dec}(\textbf{z}_p, \textbf{z}_q).\textbf{S}[p,q]$ & $||\textbf{z}_p - \textbf{z}_q||_2^2$ & $\mathit{A}_{pq}$\\
     
    \cite{ahmed2013distributed,cao2015grarep} & Shallow & Fact & $||\text{Dec}(\textbf{z}_p, \textbf{z}_q)-\textbf{S}[p,q]||_2^2$ & $\textbf{z}_p^T\textbf{z}_q$ & $\mathit{A}_{pq}$\\
     
     
    \cite{ou2016asymmetric} & Shallow & Fact & $||\text{Dec}(\textbf{z}_p, \textbf{z}_q)-\textbf{S}[p,q]||_2^2$ & $\textbf{z}_p^T\textbf{z}_q$ & Similarity martrix \textbf{S}\\
     
    \cite{perozzi2014deepwalk} & Shallow & RW & $-\textbf{S}[p, q]\log(\text{Dec}(\textbf{z}_p, \textbf{z}_q))$ & $\frac{e^{{\textbf{z}_p}^T\textbf{z}_p}}{\sum_{k \in V}e^{{\textbf{z}_p}^T\textbf{z}_k}}$ & $\mathcal{PG}(p|q)$\\
     
    \cite{yang2016revisiting} & Shallow & RW & $\mathbb{E}_{p_n \sim P_n(V)}[\log(-\sigma(\textbf{z}_p^T\textbf{z}_{q_n}))]$ & $\frac{e^{{\textbf{z}_p}^T\textbf{z}_p}}{\sum_{k \in V}e^{{\textbf{z}_p}^T\textbf{z}_k}}$ & $\mathcal{PG}(p|q)$\\
     
    \cite{grover2016node2vec} & Shallow & RW & $\sum_{(p, q) \in \mathcal{D}}- \log(\sigma(\textbf{z}_p^T\textbf{z}_{q_n})) - \gamma\mathbb{E}_{p_n} \sim P_n(V)[\log(-\sigma(\textbf{z}_p^T\textbf{z}_{q_n}))]$ & $\frac{e^{{\textbf{z}_p}^T\textbf{z}_p}}{\sum_{k \in V}e^{{\textbf{z}_p}^T\textbf{z}_k}}$ & $\mathcal{PG}(p|q)$\\

    \cite{tang2015line} & Shallow & RW & $\sum_{(p, q) \in \mathcal{D}}- \log(\sigma(\textbf{z}_p^T\textbf{z}_{q_n})) - \gamma\mathbb{E}_{p_n} \sim P_n(V)[\log(-\sigma(\textbf{z}_p^T\textbf{z}_{q_n}))]$ & $\frac{1}{1-e^{-\textbf{z}_p^T\textbf{z}_k}}$ & $\mathcal{PG}(p|q)$\\
     
    \cite{tang2015pte}& Shallow & RW & $-\textbf{S}[p, q]\log(\mathcal{PG}(p|q))$  & $\frac{1}{1-e^{-\textbf{z}_p^T\textbf{z}_k}}$ & $\mathcal{PG}(p|q)$\\

   \cite{wang2016structural,cao2016deep} & Deep & AE & $\sum_{p \in V} || \text{Dec}(\textbf{z}_p) - \textbf{s}_p||_2^2$ & MLP & $\textbf{s}_p$\\
    
    
    \cite{taheri2018learning} & Deep & AE & $\sum_{p \in V} || \text{Dec}(\textbf{z}_p) - \textbf{s}_p||_2^2$ & LSTM & $\textbf{s}_p$\\
    
    \cite{tu2018deep} & Deep & AE & $\sum_{p \in V} ||(\textbf{z}_p) - \sum_{p\in\mathcal{N}_(p)}\text{LSTM}(\textbf{z}_p)||_2^2$ & LSTM & $\textbf{s}_p$\\
    
    
    \cite{kipf2016variational}& Deep & AE & $\mathbb{E}_{u(\textbf{Z}|X,A)}[\log \mathit{p}(A|\textbf{Z})] - KL[u(\textbf{Z}|X, A)||\mathit{p}(\textbf{Z})]$ & $\textbf{z}_p^T\textbf{z}_q$ & $\mathit{A}_{pq}$\\
    
    \cite{pan2019learning} & Deep & AE & $\min_{\mathcal{G}}\max_{\mathcal{D}} \mathbb{E}_{\textbf{z}\sim \mathit{u}_z} [\log\mathcal{D}(\textbf{Z})] + \mathbb{E}_{x \sim \mathit{u}(x)} [\log(1 - \mathcal{D}(\mathcal{G}(X, A)))]$ & $\textbf{z}_p^T\textbf{z}_q$ & $\mathit{A}_{pq}$\\
    
     \hline
    \label{tab5} 
\end{longtable}
\normalsize

\paragraph{Auto-encoder:} Besides using DL models, Auto-encoder based methods differ from shallow embedding methods, and Auto-encoder uses a unary decoder rather than a pairwise one. In Auto-encoder based techniques, every node, $i$, is represented by a high-dimensional vector derived from a row in the similarity matrix, specifically, $s_i = i^{th}$ row of S, where $ S_{i,j} = s_{\mathcal{G}}(i, j)$. The Auto-encoder based techniques seek to encode each node based on the associated vector $s_i$, then reconstruct it from the embedding results, with the requirement that the reconstructed one is as similar to the original as much possible which is demonstrated in Figure \ref{fig2}.

From Equation \ref{eq1}, the encoder module actually relies on the provided $S_i$ vector. This enables deep embedding techniques based on Auto-encoders to embed local structural information inside the encoder, while the low embedding methods can simply not. Several recent deep embedding techniques aim to address the primary disadvantages of shallow embedding methods by building certain particular functions dependent on a node's neighborhood. 

\subsection{Graph-Based Self-Supervised Learning}
\label{self}
Self-supervised learning is new emergence in the advances of DL. It addresses the reliance issue of semi-supervised learning on manual labels, computationally heavy access to ground truth labels, overfitting, and poor performance against adversarial attacks \cite{liu2021self}. Self-supervised learning comes over the mentioned shortcoming by training a model to solve well-designed ``Pretext Tasks". Self-supervised learning learns more generalized representations from unlabelled, which performs better on the desired ``Downstream tasks" \cite{you2020does} such as node, edge, and graph level tasks. Table \ref{tab6} shows the existing methods for graph self-supervised learning.

\footnotesize
\begin{longtable}{p{0.8cm}p{1.4cm}p{1.2cm}p{1.2cm}p{1cm}p{6.2cm}} 
    \caption{Summary of existing GNNs based self-supervised learning methods. Abbreviations: NC=Node Classification, LP=Link Prediction, GC=Graph Classification, CSSC=  Context-Based, ASSC=Augmentation-Based, MFR=Masked Feature Regression, RAPP=Regression-based, CAPP=Classification-based, PT\&FT=Pre-training and Fine-tuning, JL=Joint Learning, URL=Unsupervised Representation Learning.}
\\ 
     \hline
    \textbf{Paper} & \textbf{Pretext Task Category} & \textbf{Training Scheme} &  \textbf{Technique} & \textbf{Task} & \textbf{Key functionality}\\ \hline
     \endfirsthead
     \multicolumn{5}{c}{{Table \thetable.\ Continued from the previous page}} \\
\hline
\centering 
\textbf{Paper} & \textbf{Pretext Task Category} & \textbf{Training Scheme} & \textbf{Technique} & \textbf{Task} & \textbf{Key functionality}\\ 
\hline
\endhead

     
    

    \cite{manessi2021graph} & MFR & JL & GCN & NC &  Train GNN models in a multi-task fashion. \\
     
    \cite{hu2019strategies}  & MFR & PT\&FT & GCN & NC &  A novel strategy for pre-training GNNs on both the individual nodes and on entire graph.  \\
     
    \cite{sun2020multi} & CAPP & JL & GCN & NC & Iteratively train an encoder architecture for assigning pseudo labels to unlabeled nodes. \\
     
    \cite{you2020does} & CAPP & PT\&FT/JL & GCN & NC & Node clustering by using pre-computed cluster index. \\
     
    \cite{hu2019pre}  & CAPP & PT\&FT & GCN & NC, LP, GC & A generic structural feature extraction by pretrain the GNNs. \\
     
     
     \cite{rong2020self}  & CAPP & PT\&FT & NN & NC, LP, GC & Learning rich structural and semantic information of molecules from enormous unlabelled molecular data. \\
     
     
     
    \cite{hamilton2017inductive} & CSSC & URL & SAGE\# & NC & Low-dimensional nodes embedding for the huge type of graphs.  \\
     
     
    \cite{kipf2016variational}& CSSC & URL & GCN & LP & Able to learn interpretable latent representations in the undirected graphs. \\
      
    \cite{jin2020self} & CSSC & PT\&FT/JL & GCN & NC & Alleviating the limitation of DL (requiring larger amounts of costly annotated data) by building the domain specific pretext tasks on unlabeled data. \\
     
     
    \cite{kim2020find}  & CSSC & JL & GAT & NC &  Graph attention model for noisy graphs. \\
      
    \cite{peng2020self}  & CSSC & JL & NN & NC, LP &  Introduced a subordinate task to predict meta-paths by employing node embeddings. \\
       
    \cite{qiu2020gcc}  &  ASSC & URL/ PT\&FT & GIN & NC, GC & Using random walk as augmentations over subgraphs and artificially designed positional node embeddings are used as node features. \\

         
    \cite{zhu2021graph}  & ASSC & URL & GCN & NC & Contrastive learning of the graph with adaptive augmentation. \\
     
     \cite{zeng2020contrastive}  & ASSC & URL/ PT\&FT/JL & NN & GC & Alleviated overfitting in graph classification. \\
     
     \cite{zhang2020iterative} & ASSC & URL/JL & GCN/GIN & GC &  Iteratively performed self distillation with graph augmentations.  \\
     
     \cite{jiao2020sub}  & CSC & URL & GCN & NC & Capturing the regional structure information in graph representations learning by using the strong interaction between the central nodes and their sampled subgraphs.  \\
      
    \cite{velickovic2019deep}  & CSC & URL & GCN & NC & Learning node representations within the graph-structured data.  \\
  
    \cite{hassani2020contrastive}  & CSC & URL & GCN & NC,GC & Graph level representation learning by maximizing the mutual information between the nodes.  \\
     
     
     
    \cite{sun2019infograph}  & CSC & URL & NN & GC & Elevating the graph level tasks by augmenting the  mutual  information  between  substructures  of different levels and graph representations. \\
     
     \cite{Subramonian}  & CSC & URL & GCN & GC & Finding pattern motifs iteratively and try  to enhance the similarity of the embeddings between motifs and a graph.  \\
    
     \cite{sun2021sugar}  & CSC & JL & GCN & GC & Employing the representation of local subgraph and global graph to discriminate the representation of the subgraph between graph. \\
     
    \cite{ren2020hdgi}  & CSC & URL & GCN/GAT & NC & Maximized combination of local and  global  mutual  information  for  representation  learning  in heterogeneous graphs.  \\
     
     \cite{wang2021self}  & CSC & PT\&FT & Shallow NN & LP & Prediction task of contextual nodes in Heterogeneous Networks.  \\
     
    \cite{park2020unsupervised}  & CSC & URL & GCN & NC,LP & Attributed multiplex network embedding.  \\
     
     \cite{cao2021bipartite} & CSC & URL & NN & NC,LP & Bipartite graph embedding by maximizing the mutual information between the graphs. mutual information maximization.  \\
     
     \cite{opolka2019spatio} & CSC & URL & GCN & NC & Tackling the challenging task of node-level regression by training embeddings. \\
     
     
    
    \cite{zhang2020graph} & CSSC & PT\&FT & NN & NC & Graph structure recovery is used to pretrain a transformer based model for graphs.  \\
     
     \cite{wan2020contrastive}  & Hybrid & JL & GCN/ HGCN & NC & Contrastive and generative graph convolutional network.  \\
     
    \cite{jin2021automated}  & Hybrid & JL & GCN/ HGCN & NC, GC & Automatically leveraging multiple pretext tasks effectively. \\
     
    \cite{wu2021self} & ASSC & PT\&FT & GCN & LC & Improving the accuracy and robustness of GCNs for recommendation. \\
     
    \cite{bui2021infercode} & Hybrid & PT\&FT & CNN & NC, LP, GC &  Code Representations learning by Predicting Subtrees. \\
     
    \cite{che2021self} & ASSC & PT\&FT & GCN & NC, LP, GC & Graph representation learning. \\
     
    \cite{jin2021multi}  & ASSC & URL & GCN & NC, LP, GC & Automatically leveraging multiple pretext tasks effectively. \\
     
    \cite{lin2021multi}  & Hybrid & URL & GCN & NC, LP, GC &  Multi-label classification of fundus images. \\
     
    \cite{li2021self}  & Hybrid & URL & GCN & NC & Ethereum Phishing Scam Detection. \\
     
    \cite{choudhary2021self}  & ASSC & URL & GCN & NC, LP, GC & Automatically leveraging multiple pretext tasks effectively. \\
     \hline
    \label{tab6}
\end{longtable}
\normalsize

\subsubsection{Pretext Tasks}
Pretext task is critical in self-supervised learning, and its construction is crucial to the model's performance for the downstream tasks. We divide the pretext task into the following categories: Masked Feature Regression (MFR), Auxiliary Property Prediction (APP), Same-Scale Contrasting (SSC), Cross-Scale Contrasting (CSC), and Hybrid Self-supervised Learning (HSL).

\paragraph{Masked Feature Regression}
The Masked Feature Regression (MFR) category of pretext task comes from the field of computer vision, precisely the task of the image inpainting inspired it \cite{yu2018generative}. The idea behind the method is to mask the feature of nodes/edges with zero or a specific value. Then the pretext task is to recover the original information of nodes/edges before GNNs mask the data. You et al. \cite{you2020does} provided a node-based MFR technique that lets GNN extract features from the surrounding environment information. The tasks, namely reconstructing raw features from noisy input data, raw reconstruction of features from ideal input data, and reconstruction of feature embeddings from noisy feature embeddings, are used as pretext tasks for learning well-generalized representations.

\paragraph{Auxiliary Property Prediction}
Besides the methods mentioned falling in the category of MFR, other methods explore the underlying attribute information of nodes/edges or even the graph structure to design new pretext tasks for providing better learning signals to the self-supervision models. These methods, both regression and classification, fall into the category of auxiliary property prediction \cite{liu2021graph}.

\textbf{Regression-based technique}:
The regression-based technique is similar to MFR, but it diverges to numerical structure and attributes property prediction of graphs by focusing on pretext tasks. A node degree-based local structure-aware pretext task was introduced by Jin et al. \cite{jin2020self} called NodeProperty, and global structure information is also considered in the computation.  Jin et al. \cite{jin2020self} proposed a method, called Distance2Cluster, to compute the distance between pre-defined clusters in the graph to unlabeled nodes. This technique forces the node representation to consider the global positioning for training. PairwiseAttrSim, another proposed method by Jin et al. \cite{jin2020self} focuses on closing the gap between similarity value for pair of nodes to the feature similarity of the pair of nodes on representation distribution. It comes from the idea of increasing feature transformation for the local structures considering over-smoothing.

\textbf{Classification-based technique}: Contrary to the Regression-based technique, the methods based on classification take the task of constructing pseudo labels to help model training. Sun et al. \cite{sun2020multi} present a technique called  Multi-Stage Self-Supervised (M3S) by leveraging the DeepCluster \cite{caron2018deep} to iteratively train an encoder architecture for assigning pseudo labels to unlabeled nodes during each iteration of the training process. Similarly, You et al. \cite{you2020does} introduced a method (Node Clustering) for node clustering by using pre-computed cluster index as self-supervised labels. 


\paragraph{Same-Scale Contrasting}
Unlike the above-mentioned two types of methods that focus on building on a single element (e.g., a single node), methods based on contrastive learning learn by training on the agreement between two graph elements. Positive pairs denoting the agreement between samples with similar semantic data are maximized in this method, while negative pairs denoting samples with unrelated semantic data are minimized. Same-Scale Contrasting (SSC) subdivides two elements of a graph by contrasting them in a similar or equal scale, e.g., graph-graph contrasting and node-node contrasting. Further SSC-based techniques are divided into two categories based on the definition of positive and negative pairs.

\textbf{Context-based}: The theory behind the Context-based same-scale contrasting (CSSC) gives closer locations in the embedding space to the contextual nodes. The contextual nodes are mostly adjacent positions in the structure of the graph. The intuition behind the idea of entities with similar semantic data to interconnected is based on the Homophily hypothesis \cite{mcpherson2001birds}. An effective method to define context is using a random walk to generate sets of nodes with similar semantic data. Closer node pairs in the walk are denoted as positive pairs, while those acquired using negative sampling are denoted negative pairs. 

\textbf{Augmentation-Based}: Contrastive visual feature learning has witnessed many advances in the last few years \cite{he2020momentum}. Augmentation-based same-scale contrasting (ASSC) is also motivated by these advances, and it generates new augmented examples for actual data samples. Defining the data augmentation process is a crucial factor in ASSC. Two augmented samples from actual data are considered positive pairs, while augmented samples from different actual data are treated as negative pairs. The techniques falling in this category are based on mutual information (MI) estimation \cite{hjelm2018learning} and using InfoNCE for estimation \cite{oord2018representation}. Qiu et al. \cite{qiu2020gcc} presented  a method called Gragh contrastive coding (GCC) for node-level tasks; focusing on universal unattributed graphs. This technique uses random walk as augmentations over subgraphs with Restart for each node, and then artificially designed positional node embeddings are used as node features. Zhu et al. \cite{zhu2020deep} presented a technique, called Graph contrastive representation learning (GRACE), that evolves two augmentations strategies by removing masking node features and edges for generating an augmented representation of the graph. For contrast purposes, both inter and intra view negative pairs are taken into account.


\paragraph{Cross-Scale Contrasting}
The Cross-Scale Contrasting (CSC) technique is different from SSC, and it learns different scale graph (node-subgraph, node-graph contrasting) representations by contrasting. Summary of the graph/subgraph is generally acquired by adopting a readout function. These techniques also inherit the idea of mutual information maximization, as is the case in ASSC. Hjelm et al. \cite{hjelm2018learning} proposed a method using Jensen Shannon divergence as mutual information estimator. Velickovic et al. \cite{velickovic2019deep} presented Deep Graph Infomax (DGI) to learn representations of the nodes by maximizing the mutual information between the top-level summary of graphs and the corresponding patch representations. DGI pollutes the original graph using node features via random sampling for getting negative samples of each graph. The structure of the graph is preserved in DGI. Similarly, Hassani and Khasahmadi \cite{hassani2020contrastive} devised a method, called Multi-view representation learning on graphs (MVGRL), to consider multi-view contrasting. This method takes graph diffusion and the original graph structure as two different views. The goal is to maximize the mutual information between large-scale graphs and cross-view representations. Jiao et al. \cite{jiao2020sub} presented another method, called a Sub-graph Contrast (Subg-Con), to contrast subgraphs context and the node embeddings for learning local structure information of the graph.

\paragraph{Hybrid Self-supervised Learning} 
Few proposed methods combine pretext tasks from different categories in multitask learning strategies to take leverage their advantages. Hu et al. \cite{hu2020gpt} proposed MFR-based Generative pretraining GNN (GPT-GNN) that drops edge prediction task (CSSC) to a graph generation task for pretraining the GNN. Peng et al. \cite{peng2020graph} gave  Graphical mutual Information (GMI) method, which jointly maximizes feature mutual information between the raw feature of neighboring node and node embedding. It also maximizes the edge mutual information (node embedding of two adjoining nodes) for learning graph representation.  Node feature reconstruction (MFR) is utilized by Zhang et al. \cite{zhang2020graph} in their proposed method, Bert. Along with MFR, graph structure recovery (CSSC) is used to pre-train a transformer-based graphs model. Wan et al. \cite{wan2020contrastive} took context-based SSCs and augmentation as self-supervised learning signals and learned them simultaneously by using downstream node classification tasks.

\subsubsection{Self-supervised Training Strategies}

Depending on the relationship of pretext tasks, downstream tasks, and graph encoders, training schemes for self-supervised learning methods are categories into three types: Joint Learning (JL), Pre-training and Fine-tuning (PT \& FT), and Unsupervised Representation Learning (URL).

\paragraph{Joint Learning}
In this scheme, the pretext and downstream tasks are jointly trained with the encoder.  The combined loss function takes error from downstream task loss function and self-supervised learning process. A trade-off hyperparameter controls the percentage of contribution of each error into total error. It is taken as a multitask learning or regularization of the downstream task used as a pretext task.

\paragraph{Pre-training and Fine-tuning} In this scheme, pretext tasks along with encoder are pre-trained at the start. This is treated as the initialization of parameters in the encoder. Further, the prediction head and pre-trained encoder are fine-tuned simultaneously under the guidance of particular downstream tasks.

\paragraph{Unsupervised Representation Learning}
Like PT \& FT, this scheme also performs the pretext tasks, and the encoder is pre-trained at the start.  However, the second stage differs; the encoder parameters are locked when training the model using the downstream task. URL is more challenging compared to other training schemes as encoder training is performed under no supervision.

\section{General Design Architecture of GNN}
\label{general}
\begin{figure}[!ht]
    \centering
    \includegraphics[width=0.4\textwidth]{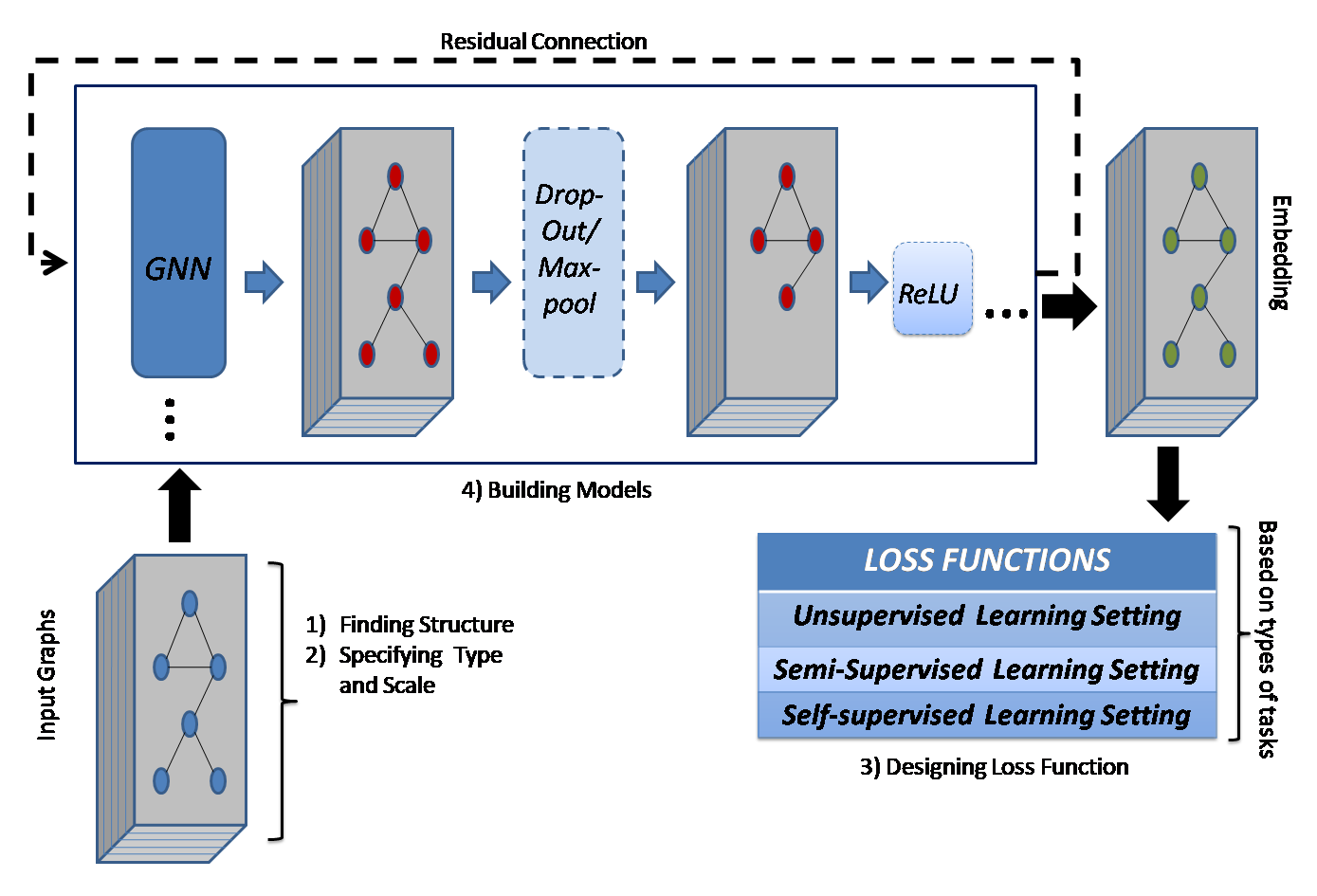}
    \caption{General Architecture of GNN models.}
    \label{fig3}
\end{figure}

In this section, we present GNN models from the perspective of a designer. We will first explain the overall design workflow for creating a GNN model as shown in the figure \ref{fig3}. Then, in Sections \ref{struc}, \ref{types}, \ref{loss} and \ref{computational}, we go through each stage in-depth. The general design process of a GNN model for a given task on a specific graph type consists of four steps- a) To identify graph structure in the application, b) To describe type and scale of the graph, c) designing appropriate loss function, and d) To create model using computational modules. In this part, we cover broad design concepts as well as background information. The design specification of these phases is covered in subsequent sections.

\subsection{Structure of Graph}
\label{struc}
First, we must determine the graph structure of the application.
There are often two types of scenarios- structural and non-structural. The graph structures are apparent in various systems such as physical systems, molecules, knowledge networks, etc. Graphs are also implicitly present in non-structural contexts. Therefore we must first create the graph from the specific task, such as generating a fully connected ``word” graph for text or a picture's scene graphs. Then we have the graph, and the subsequent process involves finding the best GNN model for this graph.

\subsection{Types of Graph}
\label{types}
 After finding out the structure of the graph in the application under consideration, we must determine the graph type and scale. Complex graphs have the innate ability to hold more information about the nodes and edges. Graphs are often classified as follows:
 a) Graphs that are directed or undirected. Directed graphs have directed edges from one node to the next, giving added information than the edges present in undirected graphs. Undirected edges are treated as two directed edges in an undirected graph. b) Graphs that are homogeneous or heterogeneous. Homogeneous networks have the same type of nodes and edges, but nodes and edges in heterogeneous graphs have different types. The types of nodes and edges are crucial and should be investigated thoroughly for heterogeneous graphs. c) Graphs that are static or dynamic. When the input characteristics or the graph topology changes over time, the graph is said to be dynamic. In dynamic graphs, temporal information should be carefully examined. This temporal information can be mixed because these categories are orthogonal, resulting in a dynamic, directed heterogeneous graph. Other graph forms, such as signed graphs and hypergraphs, are also tailored for specific applications. We will not list all of them here, but the main point is to examine the additional information offered by these graphs. For the design process, the extra information provided by the specification of the graph type should be explored. In terms of scale, there is no apparent distinction between ``small” and ``large” graphs. The criteria are still evolving in tandem with the advancement of computing equipment (GPUs). We consider the graph a large-scale graph; when the graph Laplacian or adjacency matrix of a graph ($O(n^2)$ space complexity) cannot be processed and stored on the device.
 
\subsection{Design Loss Function}
\label{loss}
The loss function is designed depending on the task type and training environment. There are typically three types of graph learning tasks:
node level tasks concentrate on characteristics of nodes such as node classification, node clustering, node regression, and so on. Node classification attempts to classify nodes into different groups. Node clustering aims to divide the nodes into many distinct groups, with comparable nodes in the same group. Furthermore, the node regression task is modeled to predicts a real value for each node. Edge level tasks include link prediction and edge classification that require the model to predict whether an edge exists between two specified nodes or identify edge types. Furthermore, the graph level tasks incorporate graph matching, graph classification, and graph regression, which all rely on the learning graph representations by the model.

From the supervision point of view, the graph learning tasks can also be divided into three distinct training settings: a) In the supervised scenario, labeled data is provided for training. b) The semi-supervised setup provides many unlabeled nodes and a small number of labeled nodes for the training process. 

\subsection{Build Model Using Computational Modules}
\label{computational}
We begin to construct the model using the computational modules widely used, which follows as:
a) Propagation Module: The propagation module transmits data across nodes, allowing the aggregated data to incorporate with both feature and topological data. The convolution and recurrent operators are often employed in propagation modules for collecting neighbors' information. At the same time, the skip connection is typically used to acquire information from previous node representations and alleviate the problem of over-smoothing.
b) Sampling Module: When graphs are big, graph propagation is generally required to be carried out by sampling modules. The sampling and propagation modules are frequently combined.
c) Pooling Module: Information is extracted by pooling modules when high-ranking subgraphs or graphs are required to be represented.

\section{Dataset}
\label{data}
Various graph-structured datasets are openly available for research, and experimental purposes for GNNs \cite{snapnets}. The datasets are hosted in the diverse platform, including Citation Network dataset, Webpage Citation Networks, Social Networks, Co-purchase Networks, Bio-chemical, Image, and Knowledge graph, which are described in Table \ref{tab7}. Table \ref{tab7} describes the popular datasets for GNNs, which belong to one of the categories mentioned above. We describe the datasets with their number of nodes, edges, and the number of classes.  

\subsection{Citation Network}
\textit{Cora} \cite{mccallum2000automating} dataset includes 2708 scientific papers in a class of seven. There are 5429 links in the citation network. Each publication in the dataset includes an 0/1 word-vector, which indicates the absence or presence of the corresponding word in the dictionary.  And the 1433 distinct words are contained in the dictionary. In addition, \textit{Citeseer} \cite{giles1998citeseer} collection contains 3312 scientific papers that are divided into six categories. There are 4732 links in the citation network. A 0/1-valued word vector describes each publication in the dataset, indicating the existence or absence of the associated term from the dictionary. There are 3703 distinct terms in the dictionary. Moreover, \textit{Pubmed}  \cite{sen2008collective} includes 19,717 scientific papers about diabetes from the PubMed database, which are divided into 3 categories. There are 44,338 links in the citation network. A TF/IDF weighted word vector from a dictionary of 500 unique words is used to characterize each publication in the dataset. Furthermore, \textit{DBLP} \cite{tang2008dblp} is a network of citations dataset. Citation datasets are collected from the databases such as DBLP, MAG (Microsoft Academic Graph), and others. There are 29,199 publications and 133,664 citations in the initial edition. Each article has an abstract, authors, year, location, and title. The dataset is utilized for clustering using network and side information, analyzing citation network influence, locating the most influential articles, topic modeling analysis, etc. The publications are categorized into 4 classes. Citation network are mostly directed, homogeneous and unweighted but the \textit{PubMed} and \textit{DBLP} are exception to it as shown in Table \ref{tab7} \textit{Patents} \cite{leskovec2005graphs} dataset include U.S. patent from 1/1/1963 to 12/30/1999 spanning 37 years. There are 3,774,768 nodes and 16,518,948 edges in this citation network. The \textit{HepPh} \cite{leskovec2005graphs} is a dataset collected from Arxiv, and these dataset cover citation on high energy physics theory and phenomenon. The dataset is a moderately big dataset with 34,556 edges and 22,770 nodes.

\subsection{Web Graphs}
\textit{Cornell dataset} \cite{zhang2020community} is a citation network of web pages obtained from Cornell University, with nodes representing webpages and edges representing connections or webpage accesses. The underlying objects are corresponding persons who can visit particular web pages to get course information or entertainment news. If two people are friends or classmates, they can share similar interests and online webpage accesses, explaining the networks' closed triangles. The dataset consists of 195 nodes, 286 edges, and the dataset is classified into 5 different classes. \textit{Texas dataset} \cite{zhang2020community} is also a citation network of webpages obtained from Texas University, with nodes representing webpages and edges representing connections or webpage accesses. Also, it consists of 187 nodes, 298 edges and is categorized the dataset into 5 classes. The another similar citation networks of webpages is \textit{Washington dataset} \cite{zhang2020community}. The dataset is collected from Washington University, with nodes representing web pages and edges representing connections or webpage accesses. It consists of 230 nodes, 417 edges and is categorized the dataset into 5 classes. Web graph networks are directed and heterogeneous in nature. As shown in Table \ref{tab7}, web graph datasets are unweighted. Several other datasets like \textit{BerkStan}, \textit{Google}, \textit{Stanford} \cite{leskovec2008community} , \textit{Notredame} \cite{albert1999diameter} are webpage link datasets collected from various domains, these datasets contain huge number of nodes and edges.

\footnotesize
\begin{longtable}{p{1.8cm}p{3.2cm}p{1cm}p{1cm}p{1cm}ccc}
\caption{Summary of available benchmark datasets for GNN experimentation and their characteristics. Directed: $\rightarrow$, Undirected: $\leftrightarrow$, Homogeneous: $\circ$, Heterogeneous: $\bullet$, Weighted: $\blacksquare$ and Unweighted: $\Box$.}

\\
    \hline
  \centering   \textbf{\rot{Category}} &  \textbf{\rot{Dataset}}&\textbf{\rot{Edge Type}}& \textbf{\rotatebox{45}{Heterogeneity}}& \textbf{\rot{Nature}} &\textbf{\rot{\#Nodes}} & \textbf{\rot{\#Edges}} &  \textbf{\rot{\#Classes}} \\ \hline
     \endfirsthead
     \multicolumn{5}{c}{{Table \thetable.\ Continued from the previous page}} \\
\hline
\centering 
\textbf{\rot{Category}} & \textbf{\rot{Dataset}}&\textbf{\rot{Edge Type}}& \textbf{\rot{Heterogeneity}}& \textbf{\rot{Nature}} &\textbf{\rot{\#Nodes}} & \textbf{\rot{\#Edges}} &  \textbf{\rot{\#Classes}} \\ \hline
\endhead

\multirow{7}{*}{\textbf{\makecell{Citation\\Networks}}}         & \textbf{Cora} \cite{mccallum2000automating}                                                     & $\rightarrow$ & $\circ$ & $\Box$ & 2,708 & 5,429 & 7\\
            & \textbf{Citeseer} \cite{giles1998citeseer} & $\rightarrow$ & $\circ$ & $\Box$ & 3,327 & 4,732 & 6\\
            & \textbf{PubMed}  \cite{sen2008collective}  & $\rightarrow$  & $\circ$ & $\blacksquare$ & 19,717  & 44,338   & 3 \\
            & \textbf{DBLP} \cite{tang2008dblp} & $\rightarrow$ & $\circ$ & $\blacksquare$ & 29,199 & 133,664 & 4\\
            & \textbf{Patents} \cite{leskovec2005graphs} & $\rightarrow$ & $\circ$ & $\Box$ & 3,774,768 & 16,518,948	& - \\
            & \textbf{HepPh} \cite{leskovec2005graphs} & $\rightarrow$ & $\circ$ & $\Box$ & 34,546 & 421,578	& - \\\\ \hline
\multirow{8}{*}{\textbf{\makecell{Web\\ graphs}}} & \textbf{Cornell} \cite{zhang2020community}& $\rightarrow$ & $\bullet$ & $\Box$ & 195 & 286 & 5 \\
            & \textbf{Texas} \cite{zhang2020community}& $\rightarrow$ & $\bullet$ & $\Box$ &  187 & 298 & 5 \\
            & \textbf{Washington} \cite{zhang2020community}& $\rightarrow$ & $\bullet$ & $\Box$ & 230  & 417 &  5 \\ 
            & \textbf{BerkStan} \cite{leskovec2008community}& $\rightarrow$ & $\bullet$ & $\Box$ & 685,230 & 7,600,595 & - \\
            & \textbf{Google} \cite{leskovec2008community} & $\rightarrow$ & $\bullet$ & $\Box$ & 875,713 & 5,105,039 & -  \\
            & \textbf{NotreDame} \cite{albert1999diameter} & $\rightarrow$ & $\bullet$ & $\Box$ & 325729 & 1497134 & -\\
            & \textbf{Stanford} \cite{leskovec2008community} & $\rightarrow$ & $\bullet$ & $\Box$ & 281,903 &	2,312,497 & - \\ \hline
                                                    
\multirow{6}{*}{\textbf{\makecell{Social\\Networks}}}       & \textbf{Karate club} \cite{zachary1977information} & $\leftrightarrow$ & $\bullet$ & $\Box$ & 34            & 77         & 2        \\
            & \textbf{Reddit} \cite{hamilton2017inductive} & $\rightarrow$   & $\bullet$ & $\blacksquare$  & 232965 & 11606919 & 41 \\
            & \textbf{BlogCatalog} \cite{tang2012unsupervised} & $\rightarrow$ & $\bullet$ & $\Box$ & 10312 & 333983 & 39 \\
            & \textbf{Flickr} \cite{mislove2008flickr}   &  $\leftrightarrow$   & $\bullet$ & $\blacksquare$ & 1,715,256 & 22,613,981 & 5 \\
            & \textbf{Facebook} \cite{mcauley2012learning}  & $\leftrightarrow$  & $\bullet$ & $\Box$& 4039 & 88,234 & - \\
            & \textbf{Youtube} \cite{yang2015defining}  & $\leftrightarrow$ & $\bullet$ & $\Box$  & 1,138,499 & 2,990,443 & 8,385 \\
            \hline 
\multirow{6}{*}{\textbf{\makecell{Co-purchase\\Networks}}}      & \textbf{Amazon Computers} \cite{shchur2018pitfalls}  & $\rightarrow$   & $\bullet$ & $\blacksquare$ & 13,752 & 245,861   & 10  \\
            & \textbf{Amazon Photo} \cite{shchur2018pitfalls} & $\rightarrow$   & $\bullet$ & $\blacksquare$ & 7,650             & 119,081 & 8 \\
            & \textbf{Amazon0601}\cite{leskovec2007dynamics} &   $\rightarrow$ & $\bullet$ & $\Box$ & 403,394 & 3,387,388	 & - \\
            & \textbf{Coauthor CS} \cite{shchur2018pitfalls}  &  $\rightarrow$ & $\circ$ & $\Box$ & 18,333 & 81,894 & 15  \\
            & \textbf{Coauthor Physics} \cite{shchur2018pitfalls}  & $\rightarrow$ & $\circ$ & $\Box$ & 34,493 & 247,962 & 5  \\ \hline 
\multirow{4}{*}{\textbf{Bio-chemical}}  & \textbf{MUTAG}  \cite{debnath1991structure} & $\leftrightarrow$ & $\circ$ & $\Box$ & 97900 & 202500 & 2 \\
           & \textbf{PROTEINS}\cite{borgwardt2005protein} & $\leftrightarrow$ & $\circ$ & $\Box$ & 39                & 72             & 2  \\
           & \textbf{PPI} \cite{li2021deep} & $\rightarrow$ & $\bullet$ & $\blacksquare$ & 56944             & 818716             & 121                 \\
           & \textbf{NCI-1} \cite{zhang2020structure} & $\leftrightarrow$ & $\circ$ & $\Box$ & 29 & 32 & 2  \\\hline
\multirow{7}{*}{\textbf{\makecell{Temporal\\Networks}}}   & \textbf{RedditHyperlinks} \cite{kumar2018community}  & $\rightarrow$  &$\bullet$ & $\blacksquare$& 55,863  & 858,490 & -        \\
            & \textbf{stackoverflow} \cite{paranjape2017motifs} & $\rightarrow$ &  $\bullet$ & $\Box$  & 2,601,977 & 63,497,050   & -  \\
            & \textbf{mathoverflow} \cite{paranjape2017motifs}  & $\rightarrow$ & $\bullet$ & $\Box$  & 24,818   & 506,550  & -  \\
            & \textbf{superuser} \cite{paranjape2017motifs}   & $\rightarrow$ & $\bullet$ & $\Box$  & 194,085   & 1,443,339  & -  \\
            & \textbf{askubuntu} \cite{paranjape2017motifs}  & $\rightarrow$ & $\bullet$ & $\Box$  & 159,316  & 964,437  & -  \\
            & \textbf{wiki-talk-temporal} \cite{paranjape2017motifs}     & $\rightarrow$ & $\bullet$ & $\Box$  & 1,140,149 & 7,833,140  & -  \\
            & \textbf{mooc} \cite{kumar2019predicting} & $\rightarrow$ &  $\bullet$ & $\blacksquare$ & 7,143 & 411,749  & -  \\  \hline              
\multirow{4}{*}{\textbf{\makecell{Communicat\\ion Networks}}} & \textbf{email-EuAll} \cite{leskovec2010signed}  & $\rightarrow$  &  $\bullet$ & $\Box$  & 265,214  & 420,045 & -        \\ 
            & \textbf{Enron} \cite{leskovec2009community}      & $\leftrightarrow$ &  $\bullet$ & $\Box$ & 36692 & 183831   & -\\
            & \textbf{wiki-Talk} \cite{leskovec2010signed}     & $\rightarrow$ &   $\bullet$ & $\Box$  & 2394385   & 5021410  & -  \\
            & \textbf{f2f-Resistance} \cite{bai2019predicting}     & $\rightarrow$ &   $\bullet$ & $\blacksquare$ & 451 & 3,126,993  & -  \\ \hline
\multirow{5}{*}{\textbf{\makecell{Autonomous\\ systems\\ graphs}}} & \textbf{as-733} (733 graphs) \cite{leskovec2005graphs}  & $\leftrightarrow$  &  $\circ$ & $\Box$ & 103-6,474 & 243-13,233 & - \\ 
            & \textbf{Skitter} \cite{leskovec2005graphs} & $\leftrightarrow$ &  $\circ$ & $\Box$ & 1,088,092 & 1,541,898 & - \\
            & \textbf{Caida} (122 graphs) \cite{leskovec2005graphs}     &  $\rightarrow$ &    $\circ$ & $\Box$ & 1,379,917 & 1,921,660  & -  \\
            & \textbf{Oregon-1} (9 graphs) \cite{leskovec2005graphs}     &  $\leftrightarrow$ &    $\circ$ & $\Box$ & 1,379,917 & 1,921,660  & -  \\
            & \textbf{Oregon-2} (9 graphs) \cite{leskovec2005graphs}     &  $\leftrightarrow$ &    $\circ$ & $\Box$ & 1,379,917 & 1,921,660  & -  \\ \hline
\multirow{3}{*}{\textbf{\makecell{Road\\ Networks}}} & \textbf{roadNet-CA} \cite{leskovec2009community}  & $\leftrightarrow$  &  $\circ$ & $\Box$ & 1,965,206  & 2,766,607 & - \\
            & \textbf{roadNet-PA} \cite{leskovec2009community} & $\leftrightarrow$ &  $\circ$ & $\Box$ & 1,088,092 & 1,541,898 & - \\
            & \textbf{roadNet-TX} \cite{leskovec2009community}     &  $\leftrightarrow$ &    $\circ$ & $\Box$ & 1,379,917 & 1,921,660  & -  \\ \hline
\label{tab7}
\end{longtable}
\normalsize

\subsection{Social Networks}
\textit{Karate club} \cite{zachary1977information} dataset is a social network of a karate club that was observed for three years from the year 1970 to 1972. The network includes 34 karate club members, with relationships between pairs of members who interacted outside of the club. The dataset includes 77 edges (connections between the pairs of members). The data are shortened into a list of integer pairs. Each number symbolizes a karate club member, and a pair denotes the interaction between the two. The dataset is classified into 2 classes. \textit{Reddit} \cite{hamilton2017inductive} dataset is also one of the social network datasets that include Reddit posts from September 2014. The community, or subreddit, to which a post belongs is the node label in this scenario. A post-to-post graph was created by sampling 50 major communities and connecting posts where the same person commented on both. There are 232,965 postings in this dataset, with an average degree of 492. The dataset has 232965 nodes, 11606919 edges and is divided into 41 classes. One important dataset for the social network is \textit{BlogCatalog} \cite{tang2012unsupervised} dataset, which is a social network of bloggers from the Blogcatalog website, which manages the bloggers and their blogs. This dataset has 10,312 bloggers with unique ids starting from 1 to 10,312 and 333,983 friendship pairs. Each blogger belongs to multiple groups. There are 39 groups with indices ranging from 1 to 39. \textit{Flickr} \cite{mislove2008flickr} dataset is a sparse social network of Flickr collected in a specific time interval, where persons represent the nodes, the friend status represents the edges, and labels represent user interests. The dataset has 1,715,256 nodes, 22,613,981 edges, and the dataset is classified into 5 classes. \textit{Facebook} \cite{mcauley2012learning} dataset is a undirected graph along with \textit{Flickr}, it contains friend list of user as nodes and connection between them as edges. It has 4,039 nodes and 88,234 edges between them. \textit{Youtube} \cite{yang2015defining} dataset is also a popular video-based social network dataset where users can friend with each other, and users can also create groups in which other users can join. Here, the genre interests indicate the labels. The dataset consists of 1,128,499 nodes and 2,990,443 edges.

\subsection{Co-Purchase Networks}
 \textit{Amazon Computers}\cite{shchur2018pitfalls} dataset is an Amazon-created network of co-purchase connections, where nodes represent items and the relation of often purchased goods together. Each node is labeled with its category and includes a sparse bag-of-words feature encoding product reviews. The dataset includes 13,752 nodes (items), 245,861 edges (items that are purchased together), and 10 classes (product categories). Another Co-purchase network dataset is \textit{Amazon Photo}\cite{shchur2018pitfalls} dataset is also created by Amazon. Similar to the Amazon Computers dataset, The Amazon Photo dataset is also a network created of co-purchase connections, where nodes represent items. The edge represents the purchased items together. Each node is labeled with its category and includes a sparse bag-of-words feature encoding product reviews. The dataset includes 7,650 nodes (items), 119,081 edges (items that are purchased together), and 8 classes (product categories). \textit{Amazon0601}\cite{leskovec2007dynamics} is also the popular amazon product co-purchasing network collected from June 1 2003. The dataset contains 403,394 nodes and 3,387,388 edges. \textit{Coauthor CS}\cite{shchur2018pitfalls} dataset is an academic network that contains graphs made up of co-authorship based on the Microsoft Academic Graph, which is collected from the KDD Cup 2016 competition. The nodes in these graphs represent authors, and the edges represent co-authorship connections. Two nodes are considered as connected if the nodes co-authored a publication. Each node contains a sparse bag-of-words feature that is based on the authors' paper keywords. The authors' label refers to their most active research area. The dataset contain 18,333 nodes, 81,894 edges, and 15 label/classes. Another important Co-purchase network is \textit{Coauthor Physics} \cite{shchur2018pitfalls} dataset contains co-authorship graphs based on the Microsoft Academic Graph from the KDD Cup 2016 competition. The nodes in these graphs represent authors, and the edges represent co-authorship connections; two nodes are connected if the nodes co-authored a publication. Each node contains a sparse bag-of-words feature that is based on the authors' paper keywords. The authors' label refers to their most active research area. The dataset includes 34,493 nodes, 247,962 edges, and 5 label/classes.

\subsection{Bio-Chemical}
\textit{MUTAG}\cite{debnath1991structure} dataset is a collection of nitroaromatic compounds created for the prediction task of mutagenicity in Salmonella typhimurium. Here, input graphs are used to represent chemical compounds, with vertices representing atoms and edges between vertices representing bonds between the corresponding atoms (encoded using one-hot encoding). It contains 97900 nodes, 202500 edges and is classified into 2 classes. \textit{PROTEIN} \cite{borgwardt2005protein} dataset is a collection of proteins categorized as either enzymes or non-enzymes. Nodes indicate amino acids, and if two nodes are less than 6 Angstroms away, the nodes are connected by an edge. The dataset consists of 39 nodes, 72 edges and is classified into 2 classes.\textit{Protein-Protein Interactions (PPI) }\cite{li2021deep} dataset contains yeast protein interactions which are collected from the Molecular Signatures Database. In various PPI graphs, in which each graph is corresponding to different human tissue. The positional gene sets represent nodes (56944), the gene ontology sets are given as the classes or labels (121 in total), and interactions between the proteins represent edges (818716). \textit{NCI1}\cite{zhang2020structure} is also a dataset that originates from the chemical field, in which each input graph represents a chemical compound, and each vertex is a molecular atom, and its edges indicate links between atoms. This data relates to anti-cancer screenings, in which chemicals are identified as positive or negative for lung cancer cells. Each vertex has an input label representing the associated atom type, encoded into a vector of 0/1 elements using a one-hot encoding method. The dataset contains 29 nodes, 32 edges and is categorized into 2 classes. 

\subsection{Temporal networks}
\textit{RedditHyperlinks} \cite{kumar2018community}, \textit{stackoverflow} \cite{paranjape2017motifs}, \textit{mathoverflow} \cite{paranjape2017motifs}, \textit{superuser} \cite{paranjape2017motifs}, \textit{askubuntu} \cite{paranjape2017motifs}, \textit{wiki-talk-temporal} \cite{paranjape2017motifs}, and \textit{mooc} \cite{kumar2018community} are the popular temporal networks. The \textit{RedditHyperlinks} is a network of Hyperlinks between subreddits on Reddit, and it consists of 55,863 nodes and 858,490 edges. The \textit{stackoverflow} is a network of Comments, questions, and answers on Stack Overflow. The \textit{mathoverflow} is a network for Comments, questions, and answers on Math Overflow consisting 24,818 nodes and 506,550 edges. The \textit{superuser} is a temporal network for Comments, questions, and answers on Super User containing 194,085 nodes and 1,443,339 edges. The \textit{askubuntu} is a network for Comments, questions, and answers on Ask Ubuntu, consisting 159,316 nodes and 964,437 edges. The \textit{wiki-talk-temporal} is a temporal network of Users editing talk pages on Wikipedia, containing 1,140,149 nodes and 7,833,140 edges. The \textit{mooc} is a network of student actions on a MOOC platform, with student drop-out binary labels, it consist of 7,143 nodes and 411,749 edges. 

\subsection{Communication Networks}
The commonly used Communication networks datasets are \textit{email-EuAll} \cite{leskovec2010signed}, \textit{Enron} \cite{leskovec2009community}, \textit{wiki-Tal} \cite{leskovec2010signed}, and \textit{f2f-Resistance} \cite{bai2019predicting}. The \textit{email-EuAll} communication network is the email network from a EU research institution, which consists of 265,214 nodes and edges 420,045. The \textit{Enron} is the email communication network from Enron, consisting 36692 nodes and 183831 edges. The \textit{wiki-Talk} is the wikipedia talk communication network having 2394385 nodes and 5021410 edges. The \textit{f2f-Resistance} is the dynamic face-to-face interaction network between group of people, consisting 451 nodes and 3,126,993 edges.

\subsection{Autonomous Systems Graphs}

\textit{as-733} \cite{leskovec2005graphs}, \textit{Skitter} \cite{leskovec2005graphs}, \textit{Caida} \cite{leskovec2005graphs}, \textit{Oregon-1} \cite{leskovec2005graphs}, and \textit{Oregon-2} \cite{leskovec2005graphs} are the commonly used available datasets for autonomous systems graphs. The \textit{as-733} is the autonomous systems graphs of 733 daily instances(graphs) from November 8 1997 to January 2 2000, consisting 103-6,474 nodes and 243-13,233 edges. The \textit{Skitter} is the autonomous systems internet topology graph, collected from trace routes run daily in 2005. The dataset consists of 1,088,092 nodes and 1,541,898 edges. The \textit{Caida} is the CAIDA autonomous relationships datasets, collected from January 2004 to November 2007. The dataset consists of 1,379,917 nodes and 1,921,660 edges. The \textit{Oregon-1} is the autonomous system peering information inferred from Oregon route-views between March 31 and May 26 2001. The dataset consists of 1,379,917 nodes and 1,921,660 edges. The \textit{Oregon-2} is the autonomous system peering information inferred from Oregon route-views between March 31 and May 26 2001. The dataset consists of 1,379,917 nodes and 1,921,660 edges.

\subsection{Road Networks}
\textit{roadNet-CA} \cite{leskovec2009community}, \textit{roadNet-PA} \cite{leskovec2009community}, and \textit{roadNet-TX} \cite{leskovec2009community} are the three commonly used road networks datasets. The \textit{roadNet-CA} is the dataset collected from road network of California, consisting 1,965,206 nodes and 2,766,607 edges. The \textit{roadNet-PA} is the road network of Pennsylvania, consisting 1,088,092 nodes and 1,541,898 edges. The \textit{roadNet-TX} is the road network of Texas, consisting 1,379,917 nodes and 1,921,660 edges.

\section{Theoretical and Empirical aspects of GNNs}
\label{analysis}
This section discusses the analytical aspect of various methods presented above from both theoretical and empirical standpoints.
    \subsection{Theoretical aspect}
    Here we overview the articles from diverse viewpoints on the theoretical underpinnings of GNN.
        \subsubsection{Graph Signal Processing}
        The convolution operations by the GCNs are done in the spectral domain that follows the signal processing theory for graphs and is performed over the input data.

        Li et al. \cite{li2018deeper} use graph convolution, which performs Laplacian smoothing over the feature matrix that results in similar hidden representations of neighboring nodes. Laplacian smoothing is based on Homophily's assumption that nearby nodes have similar labels. The smoothing is performed as a low pass filter applied over the input feature matrix. Wu et al. \cite{wu2019simplifying} confirmed it by removing the non-linearity between layers and weight matrices, showing that GNNs work only because of smoothing.
        
        Using the idea of a low-pass filter over the feature matrix is used by many researchers to get new insight by applying different filters \cite{zhang2019attributed}. Nt and Maehara \cite{nt2019revisiting} state that graph convolution is essentially a process of denoising for the input feature. The model performance depends significantly upon the quantity of noise in the matrix. \cite{chen2020measuring} presents two metrics to measure the smoothness of representation for nodes and the excess smoothness of GNN models to relieve the over-smoothing problem. The authors conclude that the key to over-smoothing is the noise-to-information ratio.
        
        \subsubsection{Generalization}
        Recent emphasis has also been paid to GNN's capacity for generalization. The VC dimensions of a limited class of GNNs are shown by Scarselli et al. \cite{scarselli2018vapnik}.  Garg et al. \cite{garg2020generalization} also offer stricter generalization limits for neural networks based on Rademacher constraints.

        The stability and widespread characteristics of single-layer GNNs with various convolutional filters examine Verma and Zhang \cite{verma2019stability}. The authors find that GNN stability depends on the filters' greatest own value. Knyazev et al. \cite{knyazev2019understanding} concentrate on the capacity of the GNN mechanism for generalization. Their result indicates that GNNs generalize attention in bigger and noisier graphs.
        
        \subsubsection{Expressivity}
        With regard to the expressiveness of GNNs, Xu et al. \cite{xu2018powerful}, Morris et al. \cite{morris2019weisfeiler} indicate that the graph isomorphism testing technique GCNs and GraphSAGE is not so distinctive as Weisfeiler Leman's (WL) testing. Xu et al. \cite{xu2019graph} offer more expressive GNN GINs as well. Barcelo et al. \cite{barcelo2019logical} explore whether GNNs express $FOC_2 $, which is a first-order logic fragment, go beyond a WL-test. Existing GNNs are barely suitable for writers. Garg et al. \cite{garg2020generalization} show that local GNN varieties are not able to acquire global graphic characteristics, including diameters, larger/smallest cycles, or motifs in order to learn about graph topologies.

        Loukas \cite{loukas2019graph} and Dehmamy et al. \cite{dehmamy2019understanding} claim that present efforts only consider expressiveness when GNNs contain endless units and layers. Their work examines the depth and width of representing the power of GNNs. Oono and Suzuki \cite{oono2019graph} discuss the asymptotic behavior of GNNs as the model expands them and represents them as dynamic systems.
        
        \subsubsection{Invariance}
        GNN output embedding should be permutation invariant or similar to the input features since node orderings are present in graphs. Maron et al. \cite{maron2018invariant} describe linear or equivalent permutation layers to construct invariant GNNs. Maron et al. \cite{maron2019universality} further demonstrate the finding that a higher-order tensorization of the universal invariant GNN is possible. The alternate proof by Keriven and Peyre \cite{keriven2019universal} extends the result to the equivalent case. Chen et al. \cite{chen2019equivalence} establish linkages between graph isomorphism testing and permutation invariance testing. The author uses $\sigma$ algebra to show the expression of GNNs to demonstrate their equivalence.
        
        \subsubsection{Transferability}
        Untied parameterization with graphs is one deterministic feature of GNNs. It indicates the possibility of transferring learning between graphs with performance guarantees (so-called transferability). The transferability of spectrum filters is investigated by Levie et al. \cite{levie2019transferability}, and it is shown that these filters are transferred on charts of the same dominance. The GNN behavior on graphs is analyzed by Ruiz et al. \cite{ruiz2020graphon}. The authors conclude that GNN's can be transmitted from the same graph with various sizes across graphs.
        
        \subsubsection{Label Efficiency}
        Self-supervised learning for GNNs requires a significant quantity of labeled information to achieve success. Given self-learning, an improvement in label marking efficiency was explored. Labels are chosen from informative nodes with the help of an oracle for training GNNs. Cai et al. \cite{cai2017active}, Gao et al. \cite{gao2018active}, Hu et al. \cite{hu2020graph} show that the labeling efficiency can be substantially increased by choosing high-degree nodes and uncertain nodes (informative nodes).
 
    \subsection{Empirical Aspect}
    In order to properly compare and assess, theoretical analyses and empirical research of GNNs are also necessary. We include many empirical research and standards for GNN assessment. There are various open-source codes and frameworks available to conduct experiments on the GNN models, which are mentioned in Table \ref{tab8}.

\footnotesize
\begin{table}[!ht]
      \centering
    \caption{Openly available source code for graph neural networking. The links are cited in the first column.}
        \begin{tabular}{p{3cm}p{1cm}p{2cm}p{1.5cm}p{5cm}} \\ \hline 
        \textbf{Platforms} &  \textbf{Source Code} & \textbf{Library} & \textbf{Language} & \textbf{Solution} \\ \hline 
        Euler \cite{src2} &  $\surd$ & Tensorflow & Python & Node level, and graph level tasks\\ 
        Paddle Graph Learning \cite{src3} & $\surd$ & Tensorflow & Python & Graph learning framework\\ 
        Graph-Learn \cite{src4} & $\surd$ & Tensorflow and Pytorch & Python & Focuses on portability and scalability\\ 
        Deep Graph Library \cite{src5} & $\surd$ & Tensorflow and Pytorch & Python & Ease the DL on graph\\
        OpenNE \cite{src6} & $\surd$ & PyTorch & Python & Self-supervised/unsupervised graph embedding\\ 
        CogDL \cite{src7} & $\surd$ & Pytorch & Python  & Node classification, Graph classification, and other important tasks in the graph domain\\  
        GNN \cite{src8} & $\surd$ & TensorFlow & Python & Molecular applications \\ 
        Spektral \cite{src9} & $\surd$ & Tensorflow and Keras & Python & Social Networking, molecular, GAN etc. \\
        QGNN \cite{src11} & $\surd$ &TensorFlow and PyTorch & Python & Provides quaternion GNNs \\ 
        GraphGym \cite{src12} & $\surd$ & PyTorch & Python & Parallel GNN Library\\ 
        GNN-NLP \cite{src13} & $\surd$ & PyTorch and TensorFlow & Python & GNN for NLP \\
        Stellargrph \cite{StellarGraph} & $\surd$ & TensorFlow and Keras & Python & Provides algorithms for graph classification \\
        Autogl\cite{guan2021autogl} & $\surd$ & Pytorch & Python & To conduct autoML on graph datasets and tasks easily and quickly \\
        Jraph \cite{jraph2020github} & $\surd$ & Jax & Python & A lightweight library for working with GNNs in jax \\
        Pytorch_geometric \cite{feyLenssen2019} & $\surd$ & Pytorch & Python & A geometric DL extension library for PyTorch \\
        Ktrain \cite{maiya2020ktrain} & $\surd$ & Tensorflow and Keras & Python & Making DL and AI more accessible and easier to apply \\
        PyTorch Geometric Temporal \cite{rozemberczki2021pytorch} & $\surd$ & Pytorch & Python & Making dynamic and temporal GNNs implementing quite easy \\
        Deeprobust \cite{li2020deeprobust} & $\surd$ & Pytorch & Python & An adversarial library for attack and defense methods on images and graphs \\
        Graphein \cite{Jamasb2020} & $\surd$ & Pytorch & Python & Provides functionality for producing a number of types of graph-based representations of proteins \\
        Graph Nets \cite{wang2019graph} & $\surd$ & Tensorflow & Python & A deepmind's library for building graph networks \\
        \hline
    \end{tabular}
     \label{tab8}
\end{table}
\normalsize

        \subsubsection{Evaluation}
        A crucial stage in research is the evaluation of machine learning models. Over the years, concerns have been expressed regarding experimental reproductivity and replicability. Which GNN models function and to what extent? In which sections of the models the ultimate performance contributes? Studies on proper assessment techniques are critically needed to examine such fundamental issues. Shchur et al. \cite{shchur2018pitfalls} are looking at how the GNN models perform with the same training techniques and hyperparameter tuning in semi-supervised node classification tasks. Different divides in their data sets lead to significantly different model rankings. In addition, simple models might exceed sophisticated models in suitable conditions. The structural information is not completely used in graph categorization based on a thorough assessment. You et al. \cite{you2020design}  discusses the architectural designs of GNN models and others, such as the number of layers and the aggregation function. This study gives extensive instructions for the designation of GNN for different purposes with a vast number of tests.
        
        \subsubsection{Benchmarks}
        In machine study research, high-quality and big benchmark data sets such as ImageNet are essential. However, widely adopted benchmarks are arduous in graph learning. For example, most categorization node data sets are tiny compared to real-world charts, with just 3000 to 20,000 nodes. In addition, the experimental procedures in each study, which are dangerous to the literature, are not harmonized.  This problem is alleviated by providing scalable and reliable graph learning benchmarks, as shown by Dwivedi et al. \cite{dwivedi2020benchmarking}, and Hu et al. \cite{hu2020open}. In various domains and tasks, Dwivedi et al. \cite{dwivedi2020benchmarking} construct medium scale datasets, whereas Hu et al. \cite{hu2020open} offers large-scale datasets. In addition, these two papers analyze existing GNN models and provide guidelines for further comparison.

\section{Applications}
\label{application}
Standard neural networks work on an array, whereas GNN works on graphs. In recent years, graphs have gained tremendous popularity because of their capability of representing real-world problems in connected ways. The applications are structured data. The data are utilized as an unstructured form of data, such as testing, and pictures are modeled as graphs for analysis in social networks, molecular structures, web-link data, etc. GNNs have several applications across various activities and areas. Although tasks are addressed directly by each GNN category, including node classification, graph classification, graph generation, network embedding, and spatial-temporal graph forecasting. We outline several applications based on the following fields of research. 

\subsection{Network}
\subsubsection{Social Networks}
DeepInf by Qiu et al. \cite{qiu2018deepinf} incorporates user-specific features and network structures in graph convergence and attention processes that can anticipate the social impact. SEAL by Zhang et al. \cite{zhang2018link} is a link prediction framework. SEAL extracts a local enclosing subgraph for each target link and learns generic graph-structured characteristics using a GNN. MCNE (Multiple Conditional Network Embedding) is an embedding method by Wang et al. \cite{wang2019mcne}. The method introduced the binary mask, followed by an attentive network, combining with a GNN based on the message passing. Liu et al. \cite{liu2019single} observed that one single vector representation does not suffice to integrate a network, for instance when the client has purchased products of different kinds on an online shopping website. 

\subsubsection{Physical Networks}
Neuroscience research has become a major research field involving structural and functional connectivity \cite{fornito2013graph} and centralized actions that characterize the important area of a region in the network \cite{page1999pagerank}. Functional connectivity centrality has been used to show differences in age and sex \cite{zuo2012network}, bipolar disorder \cite{deng2019abnormal}, retinitus pigmentosa \cite{lin2021altered}, diabetic optic neuropathy \cite{xu2020altered}, and genotype \cite{wink2018functional}. 

\subsubsection{Natural Language Processing}
Text categorization in natural language processing is one of the popular applications of GNNs. GNNs infer the document labels \cite{kipf2016semi} from the interrelationships of documents or words. Natural language data can also have an internal graph structure like a syntactic dependency tree despite having a sequential sequence. A syntactic dependency tree defines the syntactic relationships between words in a sentence. The Syntactic GCN is proposed by Marcheggiani et al. \cite{marcheggiani2017encoding}, which operates on top of a CNN/RNN sentence encoder. The GCN Syntactic gathers hidden words from the syntactical dependence tree of a sentence. The Syntactic GCN is used by Bastings et al. \cite{bastings2017graph} to do neural machine translation. 

\subsubsection{Traffic}
In an intelligent transport system, predicting traffic speed accurately, road volume, or density in traffic networks is essential. Many works such as Zhang et al. \cite{zhang2018gaan}, Li et al. \cite{li2017diffusion}, and Yu et al. \cite{yu2017spatio} uses spatial-temporal GNNs (STGNNs) for making models for addressing various traffic network problems. In such works, the authors take the traffic network as a spatial-temporal graph, the sensors installed on the roads as the nodes, and the distance between the pairs of sensors as the edges. Each node is a dynamic input feature with average traffic speed during a frame. 

\subsubsection{E-Commerce}
Discovering item relationships has recently gotten a lot of attention \cite{wang2018path}. The items' relationships are mostly lie in the content information of the items (such as description and reviews of the items) \cite{mcauley2015inferring}. Mcaulet et al. \cite{mcauley2015inferring} proposed Sceptre that uses the latent dirichlet allocation (LDA) for learning the content features of items and fits a logistic function over the Scepter. It is further extended by using Variational Auto-encoder (VAE) \cite{he2016learning} to generate noisy word clusters by avoiding overfitting.
Another area of research is utilizing the picture of the items to infer the visual level relationships between items. To reveal linkages at the visual level, Mcauley et al. \cite{mcauley2015image} and He et al. \cite{he2016ups} used the picture of the items for style matching. 

\subsection{Visualisation}
\subsubsection{Image Classification}
The early document image classification algorithms used optical character recognition (OCR) to draw content information. Many advanced techniques such as picture characteristics, text features, and document layout information for document image classification are emerged successfully in recent decades, including. Deep Convolutional Neural Networks (DCNN) is one of the successful models to provide new tools for document picture categorization \cite{kang2014convolutional}, as it can extract salient and hierarchical visual feature representations. DCNN can partially mirror the hierarchical structure of document layout. Several DCNN training techniques are suggested and fully examined for document image classification \cite{kang2014convolutional}. Variants of VGG-16 \cite{das2018document} have been upgraded in Tobacco data sets that are accessible to the public \cite{lewis2006building}. 

\subsubsection{Text}
GNNs have gained dramatic attention recently, and becoming more popular in text categorization \cite{yao2019graph}. Except for a recent paper, \cite{li-etal-2020-learn} that utilizes meta-learning and GNNs for cross-lingual sentiment classification but only employs GNN as a tool for meta-learning, most previous work focuses on monolingual text classification. The biggest challenge is to bridge the semantic and syntactic gap between languages. The majority of available techniques look for semantic similarities between languages and learn a language-agnostic representation for documents written in many languages \cite{chen2018adversarial}. This includes SOTA multilingual pre-trained language models \cite{devlin-etal-2019-bert}, which use large-scale multilingual corpora to pre-train transformer-based neural networks. 

\subsection{Miscellaneous}
\subsubsection{Chemistry}
Researchers also use GNNs in chemistry to characterize compounds or molecules where atoms are represented as nodes in a compounds or molecules graph, and chemical bonds between them are represented as edges. The primary tasks working by researchers in the compounds or molecules graphs are the Node classification, graph generation, and graph classification. Various works are done such as molecular fingerprints learning \cite{duvenaud2015convolutional}, to infer protein interfaces \cite{fout2017protein}, predicting molecular properties \cite{gilmer2017neural}, and synthesizing chemical compounds \cite{li2018learning}.

\subsubsection{Predict Side-Effect Due To Drug Interaction}
Recent works suggest that it is challenging to identify side effects caused by drug-drug interactions manually as the interactions are infrequent \cite{bansal2014community}. Moreover, practically testing all the possible drug combinations is impossible, and side effects in small clinical trials are not detected normally.
Feng et al. \cite{feng2020dpddi}, for example, utilize a two-layer GCN to learn node embeddings and then a DNN to predict drug-drug interaction. 

\subsubsection{Computer Vision}
GNN applications for computer vision include generating scene graphs, point clouds classification, and action identification. Recognizing semantic connections between things makes it easier for the visual scene to grasp its meaning. Models used to produce scene graphs such as \cite{li2018factorizable}, etc., aims to analyze a picture in a graphing graph consisting of objects and their semantic relations. In another application, realistic images are generated by using scene chart \cite{johnson2018image}. 

\subsubsection{Recommender Systems}
GNNs are successfully used in recommender systems problems. Graph-based recommender systems treat things and users as nodes. The graph-based recommending systems can make high-quality suggestions by exploiting the relationships between users and users, items and items, users and items, and the content information. The key to a system recommender is to determine if an item is important to the user. It can therefore be framed as an issue with the prediction of relations. The Works \cite{berg2017graph,ying2018graph} employs Convolytional GNNs as the encoders to prevent missing linkages between users and objects. Monti et al. \cite{monti2017geometric} integrate RNNs with graph convolutions to understand the mechanism behind which the known ratings are generated. 


\section{Open problems}
\label{open}
Although GNNs have had a lot of success in several domains, it is unable to provide acceptable answers for graph structures in several situations. There are several open problems that need to be investigated further. 

\subsection{Complex Graph Structures}
In real-world applications, graph topologies are both flexible and complicated. Many studies are offered to deal with complex graph structures such as heterogeneous graphs or dynamic graphs. With the fast expansion of social networks on the Internet, new issues, difficulties, and application scenarios will undoubtedly emerge, demanding more powerful models.
While graphs are a typical technique of modeling complicated systems, an abstraction is sometimes too basic to use as dynamic, time-changing systems in the real world. Sometimes it is the temporal behavior of a system that provides critical insights. Despite recent advances, developing GNN models capable of coping with continuous-time graphs represented as a stream of the node or edge-wise events remains an open subject of study.

\subsection{Model Depth}
Deep neural architectures are the key to DL's success. However, by using many graph convolutionary layers, the performance of the ConvGNN can decline drastically. When graph convolutions drive neighboring nodes closer to each other, all nodes' representations converge, in principle, with an unlimited number of graph convolutionary layers. All this raises the question if a DL approach is still a suitable technique for learning graph data.

\subsection{Scalability}
Scalability is a significant limitation for industrial applications that often need to deal with huge graphs. The Twitter social network has millions of nodes with billions of edges and low latency restrictions. Many models presented in the literature are utterly inappropriate for large-scale contexts, and the academic research community has virtually completely disregarded this feature until lately. Furthermore, graphics hardware (GPU) is not always the most significant choice for graph-structured data. In the long term, it requires graph-specific hardware.

\subsection{Higher-Order Structures}
In complex networks, higher-order structures such as motifs, graphlets, and simplicial complexes are considered to be important, such as in characterizing protein-protein interactions in biological applications. GNNs, on the other hand, are mostly confined to nodes and edges. Including such components in the message, transmission mechanism might give graph-based models additional expressive capability.

\subsection{Robustness and Guaranteed Performance}
Another important and mostly new research topic is the robustness of GNNs when data are noisy or exposed to opponent assaults. GNNs are equally subject to adversarial attacks as a family of neural network models. In contrast to adversarial attacks on pictures or text that solely focus on characteristics, graph attacks take structural information into account. Several works have been proposed to attack existing graph models, and more robust models are proposed to defend.

\section{Conclusion}
\label{conc}
In this article, we surveyed the classification GNNs, the state-of-the-art GNNs techniques, their applications, challenges, and possible solutions. We classify the GNNs based on their architecture and tasks. Then a comprehensive survey is presented for each of the learning settings: unsupervised, semi-supervised, and self-supervised learning. Furthermore, each learning setting is divided into blocks depending on the employed learning methods. We also present a general guideline for designing GNN architectures, along with popular datasets and useful applications of GNNs. In addition, we present quantitative and comparative analysis on the state-of-the-art GNNs techniques. Finally, we have presented open challenges and future directions for GNNs.

\bibliographystyle{ACM-Reference-Format}
\bibliography{acm-sample-base}


\begin{thebibliography}{209}


\ifx \showCODEN    \undefined \def \showCODEN     #1{\unskip}     \fi
\ifx \showDOI      \undefined \def \showDOI       #1{#1}\fi
\ifx \showISBNx    \undefined \def \showISBNx     #1{\unskip}     \fi
\ifx \showISBNxiii \undefined \def \showISBNxiii  #1{\unskip}     \fi
\ifx \showISSN     \undefined \def \showISSN      #1{\unskip}     \fi
\ifx \showLCCN     \undefined \def \showLCCN      #1{\unskip}     \fi
\ifx \shownote     \undefined \def \shownote      #1{#1}          \fi
\ifx \showarticletitle \undefined \def \showarticletitle #1{#1}   \fi
\ifx \showURL      \undefined \def \showURL       {\relax}        \fi
\providecommand\bibfield[2]{#2}
\providecommand\bibinfo[2]{#2}
\providecommand\natexlab[1]{#1}
\providecommand\showeprint[2][]{arXiv:#2}

\bibitem[\protect\citeauthoryear{Abadal, Jain, Guirado, López-Alonso, and
  Alarcón}{Abadal et~al\mbox{.}}{2021}]%
        {Abadal}
\bibfield{author}{\bibinfo{person}{Sergi Abadal}, \bibinfo{person}{Akshay
  Jain}, \bibinfo{person}{Robert Guirado}, \bibinfo{person}{Jorge
  López-Alonso}, {and} \bibinfo{person}{Eduard Alarcón}.}
  \bibinfo{year}{2021}\natexlab{}.
\newblock \bibinfo{title}{Computing Graph Neural Networks: A Survey from
  Algorithms to Accelerators}.
\newblock
\newblock
\showeprint[arxiv]{2010.00130}~[cs.LG]


\bibitem[\protect\citeauthoryear{Adhikari, Zhang, Ramakrishnan, and
  Prakash}{Adhikari et~al\mbox{.}}{2018}]%
        {adhikari2018sub2vec}
\bibfield{author}{\bibinfo{person}{Bijaya Adhikari}, \bibinfo{person}{Yao
  Zhang}, \bibinfo{person}{Naren Ramakrishnan}, {and} \bibinfo{person}{B~Aditya
  Prakash}.} \bibinfo{year}{2018}\natexlab{}.
\newblock \showarticletitle{Sub2vec: Feature learning for subgraphs}. In
  \bibinfo{booktitle}{\emph{Pacific-Asia Conference on Knowledge Discovery and
  Data Mining}}. Springer, \bibinfo{publisher}{Springer, Cham},
  \bibinfo{address}{Melbourne, Australia}, \bibinfo{pages}{170--182}.
\newblock


\bibitem[\protect\citeauthoryear{Ahmed, Shervashidze, Narayanamurthy,
  Josifovski, and Smola}{Ahmed et~al\mbox{.}}{2013}]%
        {ahmed2013distributed}
\bibfield{author}{\bibinfo{person}{Amr Ahmed}, \bibinfo{person}{Nino
  Shervashidze}, \bibinfo{person}{Shravan Narayanamurthy},
  \bibinfo{person}{Vanja Josifovski}, {and} \bibinfo{person}{Alexander~J
  Smola}.} \bibinfo{year}{2013}\natexlab{}.
\newblock \showarticletitle{Distributed large-scale natural graph
  factorization}. In \bibinfo{booktitle}{\emph{Proceedings of the 22nd
  international conference on World Wide Web}}. \bibinfo{publisher}{Association
  for Computing Machinery}, \bibinfo{address}{New York, NY, USA},
  \bibinfo{pages}{37--48}.
\newblock


\bibitem[\protect\citeauthoryear{Albert, Jeong, and Barab{\'a}si}{Albert
  et~al\mbox{.}}{1999}]%
        {albert1999diameter}
\bibfield{author}{\bibinfo{person}{R{\'e}ka Albert}, \bibinfo{person}{Hawoong
  Jeong}, {and} \bibinfo{person}{Albert-L{\'a}szl{\'o} Barab{\'a}si}.}
  \bibinfo{year}{1999}\natexlab{}.
\newblock \showarticletitle{Diameter of the world-wide web}.
\newblock \bibinfo{journal}{\emph{nature}} \bibinfo{volume}{401},
  \bibinfo{number}{6749} (\bibinfo{year}{1999}), \bibinfo{pages}{130--131}.
\newblock


\bibitem[\protect\citeauthoryear{alibaba}{alibaba}{2021a}]%
        {src2}
\bibfield{author}{\bibinfo{person}{alibaba}.} \bibinfo{year}{2021}\natexlab{a}.
\newblock \bibinfo{title}{{euler}}.
\newblock
\newblock
\urldef\tempurl%
\url{https://github.com/alibaba/euler}
\showURL{%
\tempurl}


\bibitem[\protect\citeauthoryear{alibaba}{alibaba}{2021b}]%
        {src4}
\bibfield{author}{\bibinfo{person}{alibaba}.} \bibinfo{year}{2021}\natexlab{b}.
\newblock \bibinfo{title}{{graph-learn}}.
\newblock
\newblock
\urldef\tempurl%
\url{https://github.com/alibaba/graph-learn}
\showURL{%
\tempurl}


\bibitem[\protect\citeauthoryear{Bai, Kumar, Leskovec, Metzger, Nunamaker, and
  Subrahmanian}{Bai et~al\mbox{.}}{2019}]%
        {bai2019predicting}
\bibfield{author}{\bibinfo{person}{Chongyang Bai}, \bibinfo{person}{Srijan
  Kumar}, \bibinfo{person}{Jure Leskovec}, \bibinfo{person}{Miriam Metzger},
  \bibinfo{person}{Jay Nunamaker}, {and} \bibinfo{person}{VS Subrahmanian}.}
  \bibinfo{year}{2019}\natexlab{}.
\newblock \showarticletitle{Predicting the Visual Focus of Attention in
  Multi-Person Discussion Videos}. In \bibinfo{booktitle}{\emph{IJCAI 2019}}.
  International Joint Conferences on Artificial Intelligence.
\newblock


\bibitem[\protect\citeauthoryear{Bansal, Yang, Karan, Menden, Costello, Tang,
  Xiao, Li, Allen, Zhong, et~al\mbox{.}}{Bansal et~al\mbox{.}}{2014}]%
        {bansal2014community}
\bibfield{author}{\bibinfo{person}{Mukesh Bansal}, \bibinfo{person}{Jichen
  Yang}, \bibinfo{person}{Charles Karan}, \bibinfo{person}{Michael~P Menden},
  \bibinfo{person}{James~C Costello}, \bibinfo{person}{Hao Tang},
  \bibinfo{person}{Guanghua Xiao}, \bibinfo{person}{Yajuan Li},
  \bibinfo{person}{Jeffrey Allen}, \bibinfo{person}{Rui Zhong},
  {et~al\mbox{.}}} \bibinfo{year}{2014}\natexlab{}.
\newblock \showarticletitle{A community computational challenge to predict the
  activity of pairs of compounds}.
\newblock \bibinfo{journal}{\emph{Nature biotechnology}} \bibinfo{volume}{32},
  \bibinfo{number}{12} (\bibinfo{year}{2014}), \bibinfo{pages}{1213--1222}.
\newblock


\bibitem[\protect\citeauthoryear{Barcel{\'o}, Kostylev, Monet, P{\'e}rez,
  Reutter, and Silva}{Barcel{\'o} et~al\mbox{.}}{2019}]%
        {barcelo2019logical}
\bibfield{author}{\bibinfo{person}{Pablo Barcel{\'o}}, \bibinfo{person}{Egor~V
  Kostylev}, \bibinfo{person}{Mikael Monet}, \bibinfo{person}{Jorge P{\'e}rez},
  \bibinfo{person}{Juan Reutter}, {and} \bibinfo{person}{Juan~Pablo Silva}.}
  \bibinfo{year}{2019}\natexlab{}.
\newblock \showarticletitle{The logical expressiveness of graph neural
  networks}. In \bibinfo{booktitle}{\emph{International Conference on Learning
  Representations}}. \bibinfo{publisher}{ICLR}, \bibinfo{address}{Ethiopia},
  \bibinfo{pages}{1--21}.
\newblock


\bibitem[\protect\citeauthoryear{Bastings, Titov, Aziz, Marcheggiani, and
  Sima'an}{Bastings et~al\mbox{.}}{2017}]%
        {bastings2017graph}
\bibfield{author}{\bibinfo{person}{Jasmijn Bastings}, \bibinfo{person}{Ivan
  Titov}, \bibinfo{person}{Wilker Aziz}, \bibinfo{person}{Diego Marcheggiani},
  {and} \bibinfo{person}{Khalil Sima'an}.} \bibinfo{year}{2017}\natexlab{}.
\newblock \bibinfo{title}{Graph convolutional encoders for syntax-aware neural
  machine translation}.
\newblock
\newblock


\bibitem[\protect\citeauthoryear{Battaglia, Pascanu, Lai, Rezende, and
  Kavukcuoglu}{Battaglia et~al\mbox{.}}{2016}]%
        {battaglia2016interaction}
\bibfield{author}{\bibinfo{person}{Peter~W Battaglia}, \bibinfo{person}{Razvan
  Pascanu}, \bibinfo{person}{Matthew Lai}, \bibinfo{person}{Danilo Rezende},
  {and} \bibinfo{person}{Koray Kavukcuoglu}.} \bibinfo{year}{2016}\natexlab{}.
\newblock \bibinfo{title}{Interaction networks for learning about objects,
  relations and physics}.
\newblock
\newblock


\bibitem[\protect\citeauthoryear{Belkin and Niyogi}{Belkin and Niyogi}{2001}]%
        {belkin2001laplacian}
\bibfield{author}{\bibinfo{person}{Mikhail Belkin} {and}
  \bibinfo{person}{Partha Niyogi}.} \bibinfo{year}{2001}\natexlab{}.
\newblock \showarticletitle{Laplacian Eigenmaps and Spectral Techniques for
  Embedding and Clustering}. In \bibinfo{booktitle}{\emph{Proceedings of the
  14th International Conference on Neural Information Processing Systems:
  Natural and Synthetic}} (Vancouver, British Columbia, Canada)
  \emph{(\bibinfo{series}{NIPS'01})}. \bibinfo{publisher}{MIT Press},
  \bibinfo{address}{Cambridge, MA, USA}, \bibinfo{pages}{585–591}.
\newblock


\bibitem[\protect\citeauthoryear{Berg, Kipf, and Welling}{Berg
  et~al\mbox{.}}{2017}]%
        {berg2017graph}
\bibfield{author}{\bibinfo{person}{Rianne van~den Berg},
  \bibinfo{person}{Thomas~N Kipf}, {and} \bibinfo{person}{Max Welling}.}
  \bibinfo{year}{2017}\natexlab{}.
\newblock \bibinfo{title}{Graph convolutional matrix completion}.
\newblock
\newblock


\bibitem[\protect\citeauthoryear{Berton and De~Andrade~Lopes}{Berton and
  De~Andrade~Lopes}{2014}]%
        {berton2014graph}
\bibfield{author}{\bibinfo{person}{Lilian Berton} {and} \bibinfo{person}{Alneu
  De~Andrade~Lopes}.} \bibinfo{year}{2014}\natexlab{}.
\newblock \showarticletitle{Graph Construction Based on Labeled Instances for
  Semi-supervised Learning}. In \bibinfo{booktitle}{\emph{2014 22nd
  International Conference on Pattern Recognition}}. \bibinfo{publisher}{IEEE},
  \bibinfo{address}{Stockholm, Sweden}, \bibinfo{pages}{2477--2482}.
\newblock
\urldef\tempurl%
\url{https://doi.org/10.1109/ICPR.2014.428}
\showDOI{\tempurl}


\bibitem[\protect\citeauthoryear{Berton, de~Paulo~Faleiros, Valejo,
  Valverde-Rebaza, and de~Andrade~Lopes}{Berton et~al\mbox{.}}{2017}]%
        {berton2017rgcli}
\bibfield{author}{\bibinfo{person}{Lilian Berton}, \bibinfo{person}{Thiago de
  Paulo~Faleiros}, \bibinfo{person}{Alan Valejo}, \bibinfo{person}{Jorge
  Valverde-Rebaza}, {and} \bibinfo{person}{Alneu de Andrade~Lopes}.}
  \bibinfo{year}{2017}\natexlab{}.
\newblock \showarticletitle{Rgcli: Robust graph that considers labeled
  instances for semi-supervised learning}.
\newblock \bibinfo{journal}{\emph{Neurocomputing}}  \bibinfo{volume}{226}
  (\bibinfo{year}{2017}), \bibinfo{pages}{238--248}.
\newblock


\bibitem[\protect\citeauthoryear{Borgwardt, Ong, Sch{\"o}nauer, Vishwanathan,
  Smola, and Kriegel}{Borgwardt et~al\mbox{.}}{2005}]%
        {borgwardt2005protein}
\bibfield{author}{\bibinfo{person}{Karsten~M Borgwardt},
  \bibinfo{person}{Cheng~Soon Ong}, \bibinfo{person}{Stefan Sch{\"o}nauer},
  \bibinfo{person}{SVN Vishwanathan}, \bibinfo{person}{Alex~J Smola}, {and}
  \bibinfo{person}{Hans-Peter Kriegel}.} \bibinfo{year}{2005}\natexlab{}.
\newblock \showarticletitle{Protein function prediction via graph kernels}.
\newblock \bibinfo{journal}{\emph{Bioinformatics}} \bibinfo{volume}{21},
  \bibinfo{number}{suppl\_1} (\bibinfo{year}{2005}), \bibinfo{pages}{i47--i56}.
\newblock


\bibitem[\protect\citeauthoryear{Bui, Yu, and Jiang}{Bui et~al\mbox{.}}{2021}]%
        {bui2021infercode}
\bibfield{author}{\bibinfo{person}{Nghi~DQ Bui}, \bibinfo{person}{Yijun Yu},
  {and} \bibinfo{person}{Lingxiao Jiang}.} \bibinfo{year}{2021}\natexlab{}.
\newblock \showarticletitle{InferCode: Self-Supervised Learning of Code
  Representations by Predicting Subtrees}. In \bibinfo{booktitle}{\emph{2021
  IEEE/ACM 43rd International Conference on Software Engineering (ICSE)}}.
  IEEE, \bibinfo{publisher}{IEEE}, \bibinfo{address}{Madrid, ES},
  \bibinfo{pages}{1186--1197}.
\newblock


\bibitem[\protect\citeauthoryear{Cai, Zheng, and Chang}{Cai
  et~al\mbox{.}}{2017}]%
        {cai2017active}
\bibfield{author}{\bibinfo{person}{Hongyun Cai}, \bibinfo{person}{Vincent~W
  Zheng}, {and} \bibinfo{person}{Kevin Chen-Chuan Chang}.}
  \bibinfo{year}{2017}\natexlab{}.
\newblock \bibinfo{title}{Active learning for graph embedding}.
\newblock
\newblock


\bibitem[\protect\citeauthoryear{Cao, Lin, Guo, Liu, Liu, and Wang}{Cao
  et~al\mbox{.}}{2021}]%
        {cao2021bipartite}
\bibfield{author}{\bibinfo{person}{Jiangxia Cao}, \bibinfo{person}{Xixun Lin},
  \bibinfo{person}{Shu Guo}, \bibinfo{person}{Luchen Liu},
  \bibinfo{person}{Tingwen Liu}, {and} \bibinfo{person}{Bin Wang}.}
  \bibinfo{year}{2021}\natexlab{}.
\newblock \showarticletitle{Bipartite Graph Embedding via Mutual Information
  Maximization}. In \bibinfo{booktitle}{\emph{Proceedings of the 14th ACM
  International Conference on Web Search and Data Mining}} (Virtual Event,
  Israel) \emph{(\bibinfo{series}{WSDM '21})}. \bibinfo{publisher}{Association
  for Computing Machinery}, \bibinfo{address}{New York, NY, USA},
  \bibinfo{pages}{635–643}.
\newblock
\showISBNx{9781450382977}
\urldef\tempurl%
\url{https://doi.org/10.1145/3437963.3441783}
\showDOI{\tempurl}


\bibitem[\protect\citeauthoryear{Cao, Lu, and Xu}{Cao et~al\mbox{.}}{2015}]%
        {cao2015grarep}
\bibfield{author}{\bibinfo{person}{Shaosheng Cao}, \bibinfo{person}{Wei Lu},
  {and} \bibinfo{person}{Qiongkai Xu}.} \bibinfo{year}{2015}\natexlab{}.
\newblock \showarticletitle{GraRep: Learning Graph Representations with Global
  Structural Information}. In \bibinfo{booktitle}{\emph{Proceedings of the 24th
  ACM International on Conference on Information and Knowledge Management}}
  (Melbourne, Australia) \emph{(\bibinfo{series}{CIKM '15})}.
  \bibinfo{publisher}{Association for Computing Machinery},
  \bibinfo{address}{New York, NY, USA}, \bibinfo{pages}{891–900}.
\newblock
\showISBNx{9781450337946}
\urldef\tempurl%
\url{https://doi.org/10.1145/2806416.2806512}
\showDOI{\tempurl}


\bibitem[\protect\citeauthoryear{Cao, Lu, and Xu}{Cao et~al\mbox{.}}{2016}]%
        {cao2016deep}
\bibfield{author}{\bibinfo{person}{Shaosheng Cao}, \bibinfo{person}{Wei Lu},
  {and} \bibinfo{person}{Qiongkai Xu}.} \bibinfo{year}{2016}\natexlab{}.
\newblock \showarticletitle{Deep Neural Networks for Learning Graph
  Representations}. In \bibinfo{booktitle}{\emph{Proceedings of the Thirtieth
  AAAI Conference on Artificial Intelligence}} (Phoenix, Arizona)
  \emph{(\bibinfo{series}{AAAI'16})}. \bibinfo{publisher}{AAAI Press},
  \bibinfo{address}{Phoenix, Arizona}, \bibinfo{pages}{1145–1152}.
\newblock


\bibitem[\protect\citeauthoryear{Cao, Li, and Zhao}{Cao et~al\mbox{.}}{2020}]%
        {cao2020unsupervised}
\bibfield{author}{\bibinfo{person}{Zeyu Cao}, \bibinfo{person}{Xiaorun Li},
  {and} \bibinfo{person}{Liaoying Zhao}.} \bibinfo{year}{2020}\natexlab{}.
\newblock \bibinfo{title}{Unsupervised Feature Learning by Autoencoder and
  Prototypical Contrastive Learning for Hyperspectral Classification}.
\newblock
\newblock


\bibitem[\protect\citeauthoryear{Caron, Bojanowski, Joulin, and Douze}{Caron
  et~al\mbox{.}}{2018}]%
        {caron2018deep}
\bibfield{author}{\bibinfo{person}{Mathilde Caron}, \bibinfo{person}{Piotr
  Bojanowski}, \bibinfo{person}{Armand Joulin}, {and} \bibinfo{person}{Matthijs
  Douze}.} \bibinfo{year}{2018}\natexlab{}.
\newblock \showarticletitle{Deep Clustering for Unsupervised Learning of Visual
  Features}. In \bibinfo{booktitle}{\emph{Computer Vision -- ECCV 2018}},
  \bibfield{editor}{\bibinfo{person}{Vittorio Ferrari},
  \bibinfo{person}{Martial Hebert}, \bibinfo{person}{Cristian Sminchisescu},
  {and} \bibinfo{person}{Yair Weiss}} (Eds.). \bibinfo{publisher}{Springer
  International Publishing}, \bibinfo{address}{Cham},
  \bibinfo{pages}{139--156}.
\newblock
\showISBNx{978-3-030-01264-9}


\bibitem[\protect\citeauthoryear{Che, Yang, Zhang, Tao, and Liu}{Che
  et~al\mbox{.}}{2021}]%
        {che2021self}
\bibfield{author}{\bibinfo{person}{Feihu Che}, \bibinfo{person}{Guohua Yang},
  \bibinfo{person}{Dawei Zhang}, \bibinfo{person}{Jianhua Tao}, {and}
  \bibinfo{person}{Tong Liu}.} \bibinfo{year}{2021}\natexlab{}.
\newblock \showarticletitle{Self-supervised graph representation learning via
  bootstrapping}.
\newblock \bibinfo{journal}{\emph{Neurocomputing}}  \bibinfo{volume}{456}
  (\bibinfo{year}{2021}), \bibinfo{pages}{88--96}.
\newblock


\bibitem[\protect\citeauthoryear{Chen, Lin, Li, Li, Zhou, and Sun}{Chen
  et~al\mbox{.}}{2020}]%
        {chen2020measuring}
\bibfield{author}{\bibinfo{person}{Deli Chen}, \bibinfo{person}{Yankai Lin},
  \bibinfo{person}{Wei Li}, \bibinfo{person}{Peng Li}, \bibinfo{person}{Jie
  Zhou}, {and} \bibinfo{person}{Xu Sun}.} \bibinfo{year}{2020}\natexlab{}.
\newblock \showarticletitle{Measuring and Relieving the Over-Smoothing Problem
  for Graph Neural Networks from the Topological View}.
\newblock \bibinfo{journal}{\emph{Proceedings of the AAAI Conference on
  Artificial Intelligence}} \bibinfo{volume}{34}, \bibinfo{number}{04}
  (\bibinfo{date}{Apr.} \bibinfo{year}{2020}), \bibinfo{pages}{3438--3445}.
\newblock
\urldef\tempurl%
\url{https://doi.org/10.1609/aaai.v34i04.5747}
\showDOI{\tempurl}


\bibitem[\protect\citeauthoryear{Chen, Sun, Athiwaratkun, Cardie, and
  Weinberger}{Chen et~al\mbox{.}}{2018}]%
        {chen2018adversarial}
\bibfield{author}{\bibinfo{person}{Xilun Chen}, \bibinfo{person}{Yu Sun},
  \bibinfo{person}{Ben Athiwaratkun}, \bibinfo{person}{Claire Cardie}, {and}
  \bibinfo{person}{Kilian Weinberger}.} \bibinfo{year}{2018}\natexlab{}.
\newblock \showarticletitle{Adversarial deep averaging networks for
  cross-lingual sentiment classification}.
\newblock \bibinfo{journal}{\emph{Transactions of the Association for
  Computational Linguistics}}  \bibinfo{volume}{6} (\bibinfo{year}{2018}),
  \bibinfo{pages}{557--570}.
\newblock


\bibitem[\protect\citeauthoryear{Chen, Villar, Chen, and Bruna}{Chen
  et~al\mbox{.}}{2019}]%
        {chen2019equivalence}
\bibfield{author}{\bibinfo{person}{Zhengdao Chen}, \bibinfo{person}{Soledad
  Villar}, \bibinfo{person}{Lei Chen}, {and} \bibinfo{person}{Joan Bruna}.}
  \bibinfo{year}{2019}\natexlab{}.
\newblock \bibinfo{title}{On the equivalence between graph isomorphism testing
  and function approximation with gnns}.
\newblock
\newblock


\bibitem[\protect\citeauthoryear{Choudhary, Rao, Katariya, Subbian, and
  Reddy}{Choudhary et~al\mbox{.}}{2021}]%
        {choudhary2021self}
\bibfield{author}{\bibinfo{person}{Nurendra Choudhary}, \bibinfo{person}{Nikhil
  Rao}, \bibinfo{person}{Sumeet Katariya}, \bibinfo{person}{Karthik Subbian},
  {and} \bibinfo{person}{Chandan~K. Reddy}.} \bibinfo{year}{2021}\natexlab{}.
\newblock \showarticletitle{Self-Supervised Hyperboloid Representations from
  Logical Queries over Knowledge Graphs}. In
  \bibinfo{booktitle}{\emph{Proceedings of the Web Conference 2021}}
  (Ljubljana, Slovenia) \emph{(\bibinfo{series}{WWW '21})}.
  \bibinfo{publisher}{Association for Computing Machinery},
  \bibinfo{address}{New York, NY, USA}, \bibinfo{pages}{1373–1384}.
\newblock
\showISBNx{9781450383127}
\urldef\tempurl%
\url{https://doi.org/10.1145/3442381.3449974}
\showDOI{\tempurl}


\bibitem[\protect\citeauthoryear{Cui, Zhou, Yang, and Liu}{Cui
  et~al\mbox{.}}{2020}]%
        {CuiAdaptive}
\bibfield{author}{\bibinfo{person}{Ganqu Cui}, \bibinfo{person}{Jie Zhou},
  \bibinfo{person}{Cheng Yang}, {and} \bibinfo{person}{Zhiyuan Liu}.}
  \bibinfo{year}{2020}\natexlab{}.
\newblock \showarticletitle{Adaptive Graph Encoder for Attributed Graph
  Embedding}. In \bibinfo{booktitle}{\emph{Proceedings of the 26th ACM SIGKDD
  International Conference on Knowledge Discovery \& Data Mining}}
  \emph{(\bibinfo{series}{KDD '20})}. \bibinfo{publisher}{Association for
  Computing Machinery}, \bibinfo{address}{NY, USA}, \bibinfo{pages}{976–985}.
\newblock
\showISBNx{9781450379984}


\bibitem[\protect\citeauthoryear{daiquocnguyen}{daiquocnguyen}{2021}]%
        {src11}
\bibfield{author}{\bibinfo{person}{daiquocnguyen}.}
  \bibinfo{year}{2021}\natexlab{}.
\newblock \bibinfo{title}{{QGNN}}.
\newblock
\newblock
\urldef\tempurl%
\url{https://github.com/daiquocnguyen/QGNN}
\showURL{%
\tempurl}
\newblock
\shownote{[Online; accessed 7. Aug. 2021].}


\bibitem[\protect\citeauthoryear{danielegrattarola}{danielegrattarola}{2021}]%
        {src9}
\bibfield{author}{\bibinfo{person}{danielegrattarola}.}
  \bibinfo{year}{2021}\natexlab{}.
\newblock \bibinfo{title}{{Spektral}}.
\newblock
\newblock
\urldef\tempurl%
\url{https://github.com/danielegrattarola/spektral}
\showURL{%
\tempurl}


\bibitem[\protect\citeauthoryear{Das, Roy, Bhattacharya, and Parui}{Das
  et~al\mbox{.}}{2018}]%
        {das2018document}
\bibfield{author}{\bibinfo{person}{Arindam Das}, \bibinfo{person}{Saikat Roy},
  \bibinfo{person}{Ujjwal Bhattacharya}, {and} \bibinfo{person}{Swapan~K
  Parui}.} \bibinfo{year}{2018}\natexlab{}.
\newblock \showarticletitle{Document image classification with intra-domain
  transfer learning and stacked generalization of deep convolutional neural
  networks}. In \bibinfo{booktitle}{\emph{2018 24th International Conference on
  Pattern Recognition (ICPR)}}. IEEE, \bibinfo{publisher}{IEEE},
  \bibinfo{address}{Beijing, China}, \bibinfo{pages}{3180--3185}.
\newblock


\bibitem[\protect\citeauthoryear{Data61}{Data61}{2018}]%
        {StellarGraph}
\bibfield{author}{\bibinfo{person}{CSIRO's Data61}.}
  \bibinfo{year}{2018}\natexlab{}.
\newblock \bibinfo{title}{StellarGraph Machine Learning Library}.
\newblock
  \bibinfo{howpublished}{\url{https://github.com/stellargraph/stellargraph}}.
\newblock


\bibitem[\protect\citeauthoryear{Debnath, Lopez~de Compadre, Debnath,
  Shusterman, and Hansch}{Debnath et~al\mbox{.}}{1991}]%
        {debnath1991structure}
\bibfield{author}{\bibinfo{person}{Asim~Kumar Debnath}, \bibinfo{person}{Rosa~L
  Lopez~de Compadre}, \bibinfo{person}{Gargi Debnath}, \bibinfo{person}{Alan~J
  Shusterman}, {and} \bibinfo{person}{Corwin Hansch}.}
  \bibinfo{year}{1991}\natexlab{}.
\newblock \showarticletitle{Structure-activity relationship of mutagenic
  aromatic and heteroaromatic nitro compounds. correlation with molecular
  orbital energies and hydrophobicity}.
\newblock \bibinfo{journal}{\emph{Journal of medicinal chemistry}}
  \bibinfo{volume}{34}, \bibinfo{number}{2} (\bibinfo{year}{1991}),
  \bibinfo{pages}{786--797}.
\newblock


\bibitem[\protect\citeauthoryear{deepmind}{deepmind}{2021}]%
        {src12}
\bibfield{author}{\bibinfo{person}{deepmind}.} \bibinfo{year}{2021}\natexlab{}.
\newblock \bibinfo{title}{{graph{$\_$}nets}}.
\newblock
\newblock
\urldef\tempurl%
\url{https://github.com/deepmind/graph_nets}
\showURL{%
\tempurl}


\bibitem[\protect\citeauthoryear{Dehmamy, Barab{\'a}si, and Yu}{Dehmamy
  et~al\mbox{.}}{2019}]%
        {dehmamy2019understanding}
\bibfield{author}{\bibinfo{person}{Nima Dehmamy},
  \bibinfo{person}{Albert-L{\'a}szl{\'o} Barab{\'a}si}, {and}
  \bibinfo{person}{Rose Yu}.} \bibinfo{year}{2019}\natexlab{}.
\newblock \bibinfo{title}{Understanding the representation power of graph
  neural networks in learning graph topology}.
\newblock
\newblock


\bibitem[\protect\citeauthoryear{Deng, Zhang, Zou, Zhang, Cheng, Guan, Lin,
  Lao, Ye, Li, et~al\mbox{.}}{Deng et~al\mbox{.}}{2019}]%
        {deng2019abnormal}
\bibfield{author}{\bibinfo{person}{Wenhao Deng}, \bibinfo{person}{Bin Zhang},
  \bibinfo{person}{Wenjin Zou}, \bibinfo{person}{Xiaofei Zhang},
  \bibinfo{person}{Xiongchao Cheng}, \bibinfo{person}{Lijie Guan},
  \bibinfo{person}{Yin Lin}, \bibinfo{person}{Guohui Lao},
  \bibinfo{person}{Biyu Ye}, \bibinfo{person}{Xuan Li}, {et~al\mbox{.}}}
  \bibinfo{year}{2019}\natexlab{}.
\newblock \showarticletitle{Abnormal degree centrality associated with
  cognitive dysfunctions in early bipolar disorder}.
\newblock \bibinfo{journal}{\emph{Frontiers in psychiatry}}
  \bibinfo{volume}{10} (\bibinfo{year}{2019}), \bibinfo{pages}{140}.
\newblock


\bibitem[\protect\citeauthoryear{Devlin, Chang, Lee, and Toutanova}{Devlin
  et~al\mbox{.}}{2019}]%
        {devlin-etal-2019-bert}
\bibfield{author}{\bibinfo{person}{Jacob Devlin}, \bibinfo{person}{Ming-Wei
  Chang}, \bibinfo{person}{Kenton Lee}, {and} \bibinfo{person}{Kristina
  Toutanova}.} \bibinfo{year}{2019}\natexlab{}.
\newblock \showarticletitle{{BERT}: Pre-training of Deep Bidirectional
  Transformers for Language Understanding}. In
  \bibinfo{booktitle}{\emph{Proceedings of the 2019 Conference of the North
  {A}merican Chapter of the Association for Computational Linguistics: Human
  Language Technologies, Volume 1 (Long and Short Papers)}}.
  \bibinfo{publisher}{Association for Computational Linguistics},
  \bibinfo{address}{Minneapolis, Minnesota}, \bibinfo{pages}{4171--4186}.
\newblock


\bibitem[\protect\citeauthoryear{Dhillon, Talukdar, and Crammer}{Dhillon
  et~al\mbox{.}}{2010}]%
        {dhillon2010learning}
\bibfield{author}{\bibinfo{person}{Paramveer~S. Dhillon},
  \bibinfo{person}{Partha~Pratim Talukdar}, {and} \bibinfo{person}{Koby
  Crammer}.} \bibinfo{year}{2010}\natexlab{}.
\newblock \showarticletitle{Learning Better Data Representation Using
  Inference-Driven Metric Learning}. In \bibinfo{booktitle}{\emph{Proceedings
  of the ACL 2010 Conference Short Papers}} (Uppsala, Sweden)
  \emph{(\bibinfo{series}{ACLShort '10})}. \bibinfo{publisher}{Association for
  Computational Linguistics}, \bibinfo{address}{USA},
  \bibinfo{pages}{377–381}.
\newblock


\bibitem[\protect\citeauthoryear{Dong, Chawla, and Swami}{Dong
  et~al\mbox{.}}{2017}]%
        {dong2017metapath2vec}
\bibfield{author}{\bibinfo{person}{Yuxiao Dong}, \bibinfo{person}{Nitesh~V
  Chawla}, {and} \bibinfo{person}{Ananthram Swami}.}
  \bibinfo{year}{2017}\natexlab{}.
\newblock \showarticletitle{Metapath2vec: Scalable representation learning for
  heterogeneous networks}. In \bibinfo{booktitle}{\emph{Proceedings of the 23rd
  ACM SIGKDD international conference on knowledge discovery and data mining}}.
  \bibinfo{publisher}{Association for Computing Machinery},
  \bibinfo{address}{New York, NY, USA}, \bibinfo{pages}{135--144}.
\newblock
\urldef\tempurl%
\url{https://doi.org/10.1145/3097983.3098036}
\showDOI{\tempurl}


\bibitem[\protect\citeauthoryear{Du, Wang, Song, Lu, and Wang}{Du
  et~al\mbox{.}}{2018}]%
        {du2018dynamic}
\bibfield{author}{\bibinfo{person}{Lun Du}, \bibinfo{person}{Yun Wang},
  \bibinfo{person}{Guojie Song}, \bibinfo{person}{Zhicong Lu}, {and}
  \bibinfo{person}{Junshan Wang}.} \bibinfo{year}{2018}\natexlab{}.
\newblock \showarticletitle{Dynamic Network Embedding: An Extended Approach for
  Skip-gram based Network Embedding.}. In \bibinfo{booktitle}{\emph{IJCAI}},
  Vol.~\bibinfo{volume}{2018}. \bibinfo{publisher}{AAAI Press},
  \bibinfo{address}{Stockholm, Sweden}, \bibinfo{pages}{2086--2092}.
\newblock


\bibitem[\protect\citeauthoryear{Duvenaud, Maclaurin, Aguilera-Iparraguirre,
  G{\'o}mez-Bombarelli, Hirzel, Aspuru-Guzik, and Adams}{Duvenaud
  et~al\mbox{.}}{2015}]%
        {duvenaud2015convolutional}
\bibfield{author}{\bibinfo{person}{David Duvenaud}, \bibinfo{person}{Dougal
  Maclaurin}, \bibinfo{person}{Jorge Aguilera-Iparraguirre},
  \bibinfo{person}{Rafael G{\'o}mez-Bombarelli}, \bibinfo{person}{Timothy
  Hirzel}, \bibinfo{person}{Al{\'a}n Aspuru-Guzik}, {and}
  \bibinfo{person}{Ryan~P Adams}.} \bibinfo{year}{2015}\natexlab{}.
\newblock \bibinfo{title}{Convolutional networks on graphs for learning
  molecular fingerprints}.
\newblock
\newblock


\bibitem[\protect\citeauthoryear{Dwivedi, Joshi, Laurent, Bengio, and
  Bresson}{Dwivedi et~al\mbox{.}}{2020}]%
        {dwivedi2020benchmarking}
\bibfield{author}{\bibinfo{person}{Vijay~Prakash Dwivedi},
  \bibinfo{person}{Chaitanya~K Joshi}, \bibinfo{person}{Thomas Laurent},
  \bibinfo{person}{Yoshua Bengio}, {and} \bibinfo{person}{Xavier Bresson}.}
  \bibinfo{year}{2020}\natexlab{}.
\newblock \bibinfo{title}{Benchmarking graph neural networks}.
\newblock
\newblock


\bibitem[\protect\citeauthoryear{Feng, Zhang, and Shi}{Feng
  et~al\mbox{.}}{2020}]%
        {feng2020dpddi}
\bibfield{author}{\bibinfo{person}{Yue-Hua Feng}, \bibinfo{person}{Shao-Wu
  Zhang}, {and} \bibinfo{person}{Jian-Yu Shi}.}
  \bibinfo{year}{2020}\natexlab{}.
\newblock \showarticletitle{DPDDI: a deep predictor for drug-drug
  interactions}.
\newblock \bibinfo{journal}{\emph{BMC bioinformatics}} \bibinfo{volume}{21},
  \bibinfo{number}{1} (\bibinfo{year}{2020}), \bibinfo{pages}{1--15}.
\newblock


\bibitem[\protect\citeauthoryear{Fey and Lenssen}{Fey and Lenssen}{2019}]%
        {feyLenssen2019}
\bibfield{author}{\bibinfo{person}{Matthias Fey} {and} \bibinfo{person}{Jan~E.
  Lenssen}.} \bibinfo{year}{2019}\natexlab{}.
\newblock \showarticletitle{Fast Graph Representation Learning with {PyTorch
  Geometric}}. In \bibinfo{booktitle}{\emph{ICLR Workshop on Representation
  Learning on Graphs and Manifolds}}.
\newblock


\bibitem[\protect\citeauthoryear{Fornito, Zalesky, and Breakspear}{Fornito
  et~al\mbox{.}}{2013}]%
        {fornito2013graph}
\bibfield{author}{\bibinfo{person}{Alex Fornito}, \bibinfo{person}{Andrew
  Zalesky}, {and} \bibinfo{person}{Michael Breakspear}.}
  \bibinfo{year}{2013}\natexlab{}.
\newblock \showarticletitle{Graph analysis of the human connectome: promise,
  progress, and pitfalls}.
\newblock \bibinfo{journal}{\emph{Neuroimage}}  \bibinfo{volume}{80}
  (\bibinfo{year}{2013}), \bibinfo{pages}{426--444}.
\newblock


\bibitem[\protect\citeauthoryear{Fouss, Pirotte, Renders, and Saerens}{Fouss
  et~al\mbox{.}}{2007}]%
        {fouss2007random}
\bibfield{author}{\bibinfo{person}{Francois Fouss}, \bibinfo{person}{Alain
  Pirotte}, \bibinfo{person}{Jean-Michel Renders}, {and} \bibinfo{person}{Marco
  Saerens}.} \bibinfo{year}{2007}\natexlab{}.
\newblock \showarticletitle{Random-walk computation of similarities between
  nodes of a graph with application to collaborative recommendation}.
\newblock \bibinfo{journal}{\emph{IEEE Transactions on knowledge and data
  engineering}} \bibinfo{volume}{19}, \bibinfo{number}{3}
  (\bibinfo{year}{2007}), \bibinfo{pages}{355--369}.
\newblock


\bibitem[\protect\citeauthoryear{Fout}{Fout}{2017}]%
        {fout2017protein}
\bibfield{author}{\bibinfo{person}{Alex~M Fout}.}
  \bibinfo{year}{2017}\natexlab{}.
\newblock \emph{\bibinfo{title}{Protein interface prediction using graph
  convolutional networks}}.
\newblock \bibinfo{thesistype}{Ph.D. Dissertation}. \bibinfo{school}{Colorado
  State University}.
\newblock


\bibitem[\protect\citeauthoryear{Gao, Yang, Zhou, Wu, Pan, and Hu}{Gao
  et~al\mbox{.}}{2018}]%
        {gao2018active}
\bibfield{author}{\bibinfo{person}{Li Gao}, \bibinfo{person}{Hong Yang},
  \bibinfo{person}{Chuan Zhou}, \bibinfo{person}{Jia Wu},
  \bibinfo{person}{Shirui Pan}, {and} \bibinfo{person}{Yue Hu}.}
  \bibinfo{year}{2018}\natexlab{}.
\newblock \showarticletitle{Active discriminative network representation
  learning}. In \bibinfo{booktitle}{\emph{IJCAI International Joint Conference
  on Artificial Intelligence}}. \bibinfo{publisher}{AAAI Press},
  \bibinfo{address}{Stockholm, Sweden}, \bibinfo{pages}{2142–2148}.
\newblock


\bibitem[\protect\citeauthoryear{Garg, Jegelka, and Jaakkola}{Garg
  et~al\mbox{.}}{2020}]%
        {garg2020generalization}
\bibfield{author}{\bibinfo{person}{Vikas Garg}, \bibinfo{person}{Stefanie
  Jegelka}, {and} \bibinfo{person}{Tommi Jaakkola}.}
  \bibinfo{year}{2020}\natexlab{}.
\newblock \showarticletitle{Generalization and Representational Limits of Graph
  Neural Networks}.
\newblock \bibinfo{journal}{\emph{Proceedings of the 37th International
  Conference on Machine Learning}}  \bibinfo{volume}{119}
  (\bibinfo{date}{13--18 Jul} \bibinfo{year}{2020}),
  \bibinfo{pages}{3419--3430}.
\newblock
\urldef\tempurl%
\url{http://proceedings.mlr.press/v119/garg20c.html}
\showURL{%
\tempurl}


\bibitem[\protect\citeauthoryear{Giles, Bollacker, and Lawrence}{Giles
  et~al\mbox{.}}{1998}]%
        {giles1998citeseer}
\bibfield{author}{\bibinfo{person}{C.~Lee Giles}, \bibinfo{person}{Kurt~D.
  Bollacker}, {and} \bibinfo{person}{Steve Lawrence}.}
  \bibinfo{year}{1998}\natexlab{}.
\newblock \showarticletitle{CiteSeer: An Automatic Citation Indexing System}.
  In \bibinfo{booktitle}{\emph{Proceedings of the Third ACM Conference on
  Digital Libraries}} (Pittsburgh, Pennsylvania, USA)
  \emph{(\bibinfo{series}{DL '98})}. \bibinfo{publisher}{Association for
  Computing Machinery}, \bibinfo{address}{New York, NY, USA},
  \bibinfo{pages}{89–98}.
\newblock
\showISBNx{0897919653}
\urldef\tempurl%
\url{https://doi.org/10.1145/276675.276685}
\showDOI{\tempurl}


\bibitem[\protect\citeauthoryear{Gilmer, Schoenholz, Riley, Vinyals, and
  Dahl}{Gilmer et~al\mbox{.}}{2017}]%
        {gilmer2017neural}
\bibfield{author}{\bibinfo{person}{Justin Gilmer}, \bibinfo{person}{Samuel~S
  Schoenholz}, \bibinfo{person}{Patrick~F Riley}, \bibinfo{person}{Oriol
  Vinyals}, {and} \bibinfo{person}{George~E Dahl}.}
  \bibinfo{year}{2017}\natexlab{}.
\newblock \showarticletitle{Neural message passing for quantum chemistry}.
\newblock \bibinfo{journal}{\emph{Proceedings of the 34th International
  Conference on Machine Learning}}  \bibinfo{volume}{70}
  (\bibinfo{year}{2017}), \bibinfo{pages}{1263--1272}.
\newblock


\bibitem[\protect\citeauthoryear{Godwin*, Keck*, Battaglia, Bapst, Kipf, Li,
  Stachenfeld, Veli\v{c}kovi\'{c}, and Sanchez-Gonzalez}{Godwin*
  et~al\mbox{.}}{2020}]%
        {jraph2020github}
\bibfield{author}{\bibinfo{person}{Jonathan Godwin*}, \bibinfo{person}{Thomas
  Keck*}, \bibinfo{person}{Peter Battaglia}, \bibinfo{person}{Victor Bapst},
  \bibinfo{person}{Thomas Kipf}, \bibinfo{person}{Yujia Li},
  \bibinfo{person}{Kimberly Stachenfeld}, \bibinfo{person}{Petar
  Veli\v{c}kovi\'{c}}, {and} \bibinfo{person}{Alvaro Sanchez-Gonzalez}.}
  \bibinfo{year}{2020}\natexlab{}.
\newblock \bibinfo{booktitle}{\emph{{J}raph: {A} library for graph neural
  networks in jax.}}
\newblock
\urldef\tempurl%
\url{http://github.com/deepmind/jraph}
\showURL{%
\tempurl}


\bibitem[\protect\citeauthoryear{Gori, Monfardini, and Scarselli}{Gori
  et~al\mbox{.}}{2005}]%
        {gori2005new}
\bibfield{author}{\bibinfo{person}{Marco Gori}, \bibinfo{person}{Gabriele
  Monfardini}, {and} \bibinfo{person}{Franco Scarselli}.}
  \bibinfo{year}{2005}\natexlab{}.
\newblock \showarticletitle{A new model for learning in graph domains}. In
  \bibinfo{booktitle}{\emph{Proceedings. 2005 IEEE International Joint
  Conference on Neural Networks, 2005.}}, Vol.~\bibinfo{volume}{2}. IEEE,
  \bibinfo{publisher}{IEEE}, \bibinfo{address}{Montreal, QC, Canada},
  \bibinfo{pages}{729--734}.
\newblock


\bibitem[\protect\citeauthoryear{Grover and Leskovec}{Grover and
  Leskovec}{2016}]%
        {grover2016node2vec}
\bibfield{author}{\bibinfo{person}{Aditya Grover} {and} \bibinfo{person}{Jure
  Leskovec}.} \bibinfo{year}{2016}\natexlab{}.
\newblock \showarticletitle{node2vec: Scalable feature learning for networks}.
  In \bibinfo{booktitle}{\emph{Proceedings of the 22nd ACM SIGKDD international
  conference on Knowledge discovery and data mining}}.
  \bibinfo{publisher}{Association for Computing Machinery},
  \bibinfo{address}{New York, NY, USA}, \bibinfo{pages}{855--864}.
\newblock


\bibitem[\protect\citeauthoryear{Guan, Zhang, Li, Chang, Zhang, Qin, Jiang,
  Wang, and Zhu}{Guan et~al\mbox{.}}{2021}]%
        {guan2021autogl}
\bibfield{author}{\bibinfo{person}{Chaoyu Guan}, \bibinfo{person}{Ziwei Zhang},
  \bibinfo{person}{Haoyang Li}, \bibinfo{person}{Heng Chang},
  \bibinfo{person}{Zeyang Zhang}, \bibinfo{person}{Yijian Qin},
  \bibinfo{person}{Jiyan Jiang}, \bibinfo{person}{Xin Wang}, {and}
  \bibinfo{person}{Wenwu Zhu}.} \bibinfo{year}{2021}\natexlab{}.
\newblock \showarticletitle{Auto{GL}: A Library for Automated Graph Learning}.
  In \bibinfo{booktitle}{\emph{ICLR 2021 Workshop on Geometrical and
  Topological Representation Learning}}.
\newblock
\urldef\tempurl%
\url{https://openreview.net/forum?id=0yHwpLeInDn}
\showURL{%
\tempurl}


\bibitem[\protect\citeauthoryear{Hamilton, Ying, and Leskovec}{Hamilton
  et~al\mbox{.}}{2017a}]%
        {hamilton2017inductive}
\bibfield{author}{\bibinfo{person}{William~L Hamilton}, \bibinfo{person}{Rex
  Ying}, {and} \bibinfo{person}{Jure Leskovec}.}
  \bibinfo{year}{2017}\natexlab{a}.
\newblock \showarticletitle{Inductive representation learning on large graphs}.
\newblock \bibinfo{journal}{\emph{Proceedings of the 31st International
  Conference on Neural Information Processing Systems}}  \bibinfo{volume}{30}
  (\bibinfo{year}{2017}), \bibinfo{pages}{1025--1035}.
\newblock


\bibitem[\protect\citeauthoryear{Hamilton, Ying, and Leskovec}{Hamilton
  et~al\mbox{.}}{2017b}]%
        {hamilton2017representation}
\bibfield{author}{\bibinfo{person}{William~L Hamilton}, \bibinfo{person}{Rex
  Ying}, {and} \bibinfo{person}{Jure Leskovec}.}
  \bibinfo{year}{2017}\natexlab{b}.
\newblock \bibinfo{title}{Representation learning on graphs: Methods and
  applications}.
\newblock
\newblock


\bibitem[\protect\citeauthoryear{Hassani and Khasahmadi}{Hassani and
  Khasahmadi}{2020}]%
        {hassani2020contrastive}
\bibfield{author}{\bibinfo{person}{Kaveh Hassani} {and}
  \bibinfo{person}{Amir~Hosein Khasahmadi}.} \bibinfo{year}{2020}\natexlab{}.
\newblock \showarticletitle{Contrastive multi-view representation learning on
  graphs}.
\newblock \bibinfo{journal}{\emph{Proceedings of the 37th International
  Conference on Machine Learning}}  \bibinfo{volume}{119}
  (\bibinfo{year}{2020}), \bibinfo{pages}{4116--4126}.
\newblock


\bibitem[\protect\citeauthoryear{He, Fan, Wu, Xie, and Girshick}{He
  et~al\mbox{.}}{2020}]%
        {he2020momentum}
\bibfield{author}{\bibinfo{person}{Kaiming He}, \bibinfo{person}{Haoqi Fan},
  \bibinfo{person}{Yuxin Wu}, \bibinfo{person}{Saining Xie}, {and}
  \bibinfo{person}{Ross Girshick}.} \bibinfo{year}{2020}\natexlab{}.
\newblock \showarticletitle{Momentum contrast for unsupervised visual
  representation learning}. In \bibinfo{booktitle}{\emph{Proceedings of the
  IEEE/CVF Conference on Computer Vision and Pattern Recognition}}.
  \bibinfo{publisher}{IEEE}, \bibinfo{address}{Seattle, WA, USA},
  \bibinfo{pages}{9729--9738}.
\newblock


\bibitem[\protect\citeauthoryear{He and McAuley}{He and McAuley}{2016}]%
        {he2016ups}
\bibfield{author}{\bibinfo{person}{Ruining He} {and} \bibinfo{person}{Julian
  McAuley}.} \bibinfo{year}{2016}\natexlab{}.
\newblock \showarticletitle{Ups and downs: Modeling the visual evolution of
  fashion trends with one-class collaborative filtering}. In
  \bibinfo{booktitle}{\emph{proceedings of the 25th international conference on
  world wide web}}. \bibinfo{publisher}{International World Wide Web
  Conferences Steering Committee}, \bibinfo{address}{Republic and Canton of
  Geneva, CHE}, \bibinfo{pages}{507--517}.
\newblock


\bibitem[\protect\citeauthoryear{He, Packer, and McAuley}{He
  et~al\mbox{.}}{2016}]%
        {he2016learning}
\bibfield{author}{\bibinfo{person}{Ruining He}, \bibinfo{person}{Charles
  Packer}, {and} \bibinfo{person}{Julian McAuley}.}
  \bibinfo{year}{2016}\natexlab{}.
\newblock \showarticletitle{Learning compatibility across categories for
  heterogeneous item recommendation}. In \bibinfo{booktitle}{\emph{2016 IEEE
  16th International Conference on Data Mining (ICDM)}}. IEEE,
  \bibinfo{publisher}{IEEE}, \bibinfo{address}{Barcelona, Spain},
  \bibinfo{pages}{937--942}.
\newblock


\bibitem[\protect\citeauthoryear{Hjelm, Fedorov, Lavoie-Marchildon, Grewal,
  Bachman, Trischler, and Bengio}{Hjelm et~al\mbox{.}}{2018}]%
        {hjelm2018learning}
\bibfield{author}{\bibinfo{person}{R~Devon Hjelm}, \bibinfo{person}{Alex
  Fedorov}, \bibinfo{person}{Samuel Lavoie-Marchildon}, \bibinfo{person}{Karan
  Grewal}, \bibinfo{person}{Phil Bachman}, \bibinfo{person}{Adam Trischler},
  {and} \bibinfo{person}{Yoshua Bengio}.} \bibinfo{year}{2018}\natexlab{}.
\newblock \bibinfo{title}{Learning deep representations by mutual information
  estimation and maximization}.
\newblock
\newblock


\bibitem[\protect\citeauthoryear{Hu, Xiong, Qu, Yuan, C{\^o}t{\'e}, Liu, and
  Tang}{Hu et~al\mbox{.}}{2020c}]%
        {hu2020graph}
\bibfield{author}{\bibinfo{person}{Shengding Hu}, \bibinfo{person}{Zheng
  Xiong}, \bibinfo{person}{Meng Qu}, \bibinfo{person}{Xingdi Yuan},
  \bibinfo{person}{Marc-Alexandre C{\^o}t{\'e}}, \bibinfo{person}{Zhiyuan Liu},
  {and} \bibinfo{person}{Jian Tang}.} \bibinfo{year}{2020}\natexlab{c}.
\newblock \bibinfo{title}{Graph policy network for transferable active learning
  on graphs}.
\newblock
\newblock


\bibitem[\protect\citeauthoryear{Hu, Fey, Zitnik, Dong, Ren, Liu, Catasta, and
  Leskovec}{Hu et~al\mbox{.}}{2020b}]%
        {hu2020open}
\bibfield{author}{\bibinfo{person}{Weihua Hu}, \bibinfo{person}{Matthias Fey},
  \bibinfo{person}{Marinka Zitnik}, \bibinfo{person}{Yuxiao Dong},
  \bibinfo{person}{Hongyu Ren}, \bibinfo{person}{Bowen Liu},
  \bibinfo{person}{Michele Catasta}, {and} \bibinfo{person}{Jure Leskovec}.}
  \bibinfo{year}{2020}\natexlab{b}.
\newblock \bibinfo{title}{Open graph benchmark: Datasets for machine learning
  on graphs}.
\newblock
\newblock


\bibitem[\protect\citeauthoryear{Hu, Liu, Gomes, Zitnik, Liang, Pande, and
  Leskovec}{Hu et~al\mbox{.}}{2019b}]%
        {hu2019strategies}
\bibfield{author}{\bibinfo{person}{Weihua Hu}, \bibinfo{person}{Bowen Liu},
  \bibinfo{person}{Joseph Gomes}, \bibinfo{person}{Marinka Zitnik},
  \bibinfo{person}{Percy Liang}, \bibinfo{person}{Vijay Pande}, {and}
  \bibinfo{person}{Jure Leskovec}.} \bibinfo{year}{2019}\natexlab{b}.
\newblock \bibinfo{title}{Strategies for pre-training graph neural networks}.
\newblock
\newblock


\bibitem[\protect\citeauthoryear{Hu, Dong, Wang, Chang, and Sun}{Hu
  et~al\mbox{.}}{2020a}]%
        {hu2020gpt}
\bibfield{author}{\bibinfo{person}{Ziniu Hu}, \bibinfo{person}{Yuxiao Dong},
  \bibinfo{person}{Kuansan Wang}, \bibinfo{person}{Kai-Wei Chang}, {and}
  \bibinfo{person}{Yizhou Sun}.} \bibinfo{year}{2020}\natexlab{a}.
\newblock \showarticletitle{Gpt-gnn: Generative pre-training of graph neural
  networks}. In \bibinfo{booktitle}{\emph{Proceedings of the 26th ACM SIGKDD
  International Conference on Knowledge Discovery \& Data Mining}}.
  \bibinfo{publisher}{Association for Computing Machinery},
  \bibinfo{address}{New York, NY, USA}, \bibinfo{pages}{1857--1867}.
\newblock


\bibitem[\protect\citeauthoryear{Hu, Fan, Chen, Chang, and Sun}{Hu
  et~al\mbox{.}}{2019a}]%
        {hu2019pre}
\bibfield{author}{\bibinfo{person}{Ziniu Hu}, \bibinfo{person}{Changjun Fan},
  \bibinfo{person}{Ting Chen}, \bibinfo{person}{Kai-Wei Chang}, {and}
  \bibinfo{person}{Yizhou Sun}.} \bibinfo{year}{2019}\natexlab{a}.
\newblock \bibinfo{title}{Pre-training graph neural networks for generic
  structural feature extraction}.
\newblock
\newblock


\bibitem[\protect\citeauthoryear{Jamasb, Lio, and Blundell}{Jamasb
  et~al\mbox{.}}{2020}]%
        {Jamasb2020}
\bibfield{author}{\bibinfo{person}{Arian~Rokkum Jamasb},
  \bibinfo{person}{Pietro Lio}, {and} \bibinfo{person}{Tom Blundell}.}
  \bibinfo{year}{2020}\natexlab{}.
\newblock \showarticletitle{Graphein - a Python Library for Geometric Deep
  Learning and Network Analysis on Protein Structures}.
\newblock  (\bibinfo{date}{July} \bibinfo{year}{2020}).
\newblock
\urldef\tempurl%
\url{https://doi.org/10.1101/2020.07.15.204701}
\showDOI{\tempurl}


\bibitem[\protect\citeauthoryear{Jiao, Xiong, Zhang, Zhang, Zhang, and
  Zhu}{Jiao et~al\mbox{.}}{2020}]%
        {jiao2020sub}
\bibfield{author}{\bibinfo{person}{Yizhu Jiao}, \bibinfo{person}{Yun Xiong},
  \bibinfo{person}{Jiawei Zhang}, \bibinfo{person}{Yao Zhang},
  \bibinfo{person}{Tianqi Zhang}, {and} \bibinfo{person}{Yangyong Zhu}.}
  \bibinfo{year}{2020}\natexlab{}.
\newblock \showarticletitle{Sub-graph contrast for scalable self-supervised
  graph representation learning}. In \bibinfo{booktitle}{\emph{2020 IEEE
  International Conference on Data Mining (ICDM)}}. IEEE,
  \bibinfo{publisher}{IEEE}, \bibinfo{address}{Sorrento, Italy},
  \bibinfo{pages}{222--231}.
\newblock


\bibitem[\protect\citeauthoryear{Jin, Zheng, Li, Gong, Zhou, and Pan}{Jin
  et~al\mbox{.}}{2021b}]%
        {jin2021multi}
\bibfield{author}{\bibinfo{person}{Ming Jin}, \bibinfo{person}{Yizhen Zheng},
  \bibinfo{person}{Yuan-Fang Li}, \bibinfo{person}{Chen Gong},
  \bibinfo{person}{Chuan Zhou}, {and} \bibinfo{person}{Shirui Pan}.}
  \bibinfo{year}{2021}\natexlab{b}.
\newblock \bibinfo{title}{Multi-Scale Contrastive Siamese Networks for
  Self-Supervised Graph Representation Learning}.
\newblock
\newblock


\bibitem[\protect\citeauthoryear{Jin, Derr, Liu, Wang, Wang, Liu, and Tang}{Jin
  et~al\mbox{.}}{2020}]%
        {jin2020self}
\bibfield{author}{\bibinfo{person}{Wei Jin}, \bibinfo{person}{Tyler Derr},
  \bibinfo{person}{Haochen Liu}, \bibinfo{person}{Yiqi Wang},
  \bibinfo{person}{Suhang Wang}, \bibinfo{person}{Zitao Liu}, {and}
  \bibinfo{person}{Jiliang Tang}.} \bibinfo{year}{2020}\natexlab{}.
\newblock \bibinfo{title}{Self-supervised learning on graphs: Deep insights and
  new direction}.
\newblock
\newblock


\bibitem[\protect\citeauthoryear{Jin, Liu, Zhao, Ma, Shah, and Tang}{Jin
  et~al\mbox{.}}{2021a}]%
        {jin2021automated}
\bibfield{author}{\bibinfo{person}{Wei Jin}, \bibinfo{person}{Xiaorui Liu},
  \bibinfo{person}{Xiangyu Zhao}, \bibinfo{person}{Yao Ma},
  \bibinfo{person}{Neil Shah}, {and} \bibinfo{person}{Jiliang Tang}.}
  \bibinfo{year}{2021}\natexlab{a}.
\newblock \bibinfo{title}{Automated Self-Supervised Learning for Graphs}.
\newblock
\newblock


\bibitem[\protect\citeauthoryear{Johnson, Gupta, and Fei-Fei}{Johnson
  et~al\mbox{.}}{2018}]%
        {johnson2018image}
\bibfield{author}{\bibinfo{person}{Justin Johnson}, \bibinfo{person}{Agrim
  Gupta}, {and} \bibinfo{person}{Li Fei-Fei}.} \bibinfo{year}{2018}\natexlab{}.
\newblock \showarticletitle{Image generation from scene graphs}. In
  \bibinfo{booktitle}{\emph{Proceedings of the IEEE conference on computer
  vision and pattern recognition}}. \bibinfo{publisher}{IEEE},
  \bibinfo{address}{Salt Lake City, UT, USA}, \bibinfo{pages}{1219--1228}.
\newblock


\bibitem[\protect\citeauthoryear{Kang, Kumar, Ye, Li, and Doermann}{Kang
  et~al\mbox{.}}{2014}]%
        {kang2014convolutional}
\bibfield{author}{\bibinfo{person}{Le Kang}, \bibinfo{person}{Jayant Kumar},
  \bibinfo{person}{Peng Ye}, \bibinfo{person}{Yi Li}, {and}
  \bibinfo{person}{David Doermann}.} \bibinfo{year}{2014}\natexlab{}.
\newblock \showarticletitle{Convolutional neural networks for document image
  classification}. In \bibinfo{booktitle}{\emph{2014 22nd International
  Conference on Pattern Recognition}}. IEEE, \bibinfo{publisher}{IEEE},
  \bibinfo{address}{Stockholm, Sweden}, \bibinfo{pages}{3168--3172}.
\newblock


\bibitem[\protect\citeauthoryear{Keriven and Peyr{\'e}}{Keriven and
  Peyr{\'e}}{2019}]%
        {keriven2019universal}
\bibfield{author}{\bibinfo{person}{Nicolas Keriven} {and}
  \bibinfo{person}{Gabriel Peyr{\'e}}.} \bibinfo{year}{2019}\natexlab{}.
\newblock \showarticletitle{Universal invariant and equivariant graph neural
  networks}.
\newblock \bibinfo{journal}{\emph{Advances in Neural Information Processing
  Systems}}  \bibinfo{volume}{32} (\bibinfo{year}{2019}),
  \bibinfo{pages}{7092--7101}.
\newblock


\bibitem[\protect\citeauthoryear{Kim and Oh}{Kim and Oh}{2020}]%
        {kim2020find}
\bibfield{author}{\bibinfo{person}{Dongkwan Kim} {and} \bibinfo{person}{Alice
  Oh}.} \bibinfo{year}{2020}\natexlab{}.
\newblock \showarticletitle{How to find your friendly neighborhood: Graph
  attention design with self-supervision}. In
  \bibinfo{booktitle}{\emph{International Conference on Learning
  Representations}}. \bibinfo{publisher}{ICLR}, \bibinfo{address}{Vienna,
  Austria}, \bibinfo{pages}{1--25}.
\newblock


\bibitem[\protect\citeauthoryear{Kipf and Welling}{Kipf and Welling}{2016a}]%
        {kipf2016semi}
\bibfield{author}{\bibinfo{person}{Thomas~N Kipf} {and} \bibinfo{person}{Max
  Welling}.} \bibinfo{year}{2016}\natexlab{a}.
\newblock \bibinfo{title}{Semi-supervised classification with graph
  convolutional networks}.
\newblock
\newblock


\bibitem[\protect\citeauthoryear{Kipf and Welling}{Kipf and Welling}{2016b}]%
        {kipf2016variational}
\bibfield{author}{\bibinfo{person}{Thomas~N Kipf} {and} \bibinfo{person}{Max
  Welling}.} \bibinfo{year}{2016}\natexlab{b}.
\newblock \bibinfo{title}{Variational graph auto-encoders}.
\newblock
\newblock


\bibitem[\protect\citeauthoryear{Knyazev, Taylor, and Amer}{Knyazev
  et~al\mbox{.}}{2019}]%
        {knyazev2019understanding}
\bibfield{author}{\bibinfo{person}{Boris Knyazev}, \bibinfo{person}{Graham~W
  Taylor}, {and} \bibinfo{person}{Mohamed~R Amer}.}
  \bibinfo{year}{2019}\natexlab{}.
\newblock \bibinfo{title}{Understanding attention and generalization in graph
  neural networks}.
\newblock
\newblock


\bibitem[\protect\citeauthoryear{Kumar, Hamilton, Leskovec, and Jurafsky}{Kumar
  et~al\mbox{.}}{2018}]%
        {kumar2018community}
\bibfield{author}{\bibinfo{person}{Srijan Kumar}, \bibinfo{person}{William~L
  Hamilton}, \bibinfo{person}{Jure Leskovec}, {and} \bibinfo{person}{Dan
  Jurafsky}.} \bibinfo{year}{2018}\natexlab{}.
\newblock \showarticletitle{Community interaction and conflict on the web}. In
  \bibinfo{booktitle}{\emph{Proceedings of the 2018 World Wide Web Conference
  on World Wide Web}}. International World Wide Web Conferences Steering
  Committee, \bibinfo{pages}{933--943}.
\newblock


\bibitem[\protect\citeauthoryear{Kumar, Zhang, and Leskovec}{Kumar
  et~al\mbox{.}}{2019}]%
        {kumar2019predicting}
\bibfield{author}{\bibinfo{person}{Srijan Kumar}, \bibinfo{person}{Xikun
  Zhang}, {and} \bibinfo{person}{Jure Leskovec}.}
  \bibinfo{year}{2019}\natexlab{}.
\newblock \showarticletitle{Predicting dynamic embedding trajectory in temporal
  interaction networks}. In \bibinfo{booktitle}{\emph{Proceedings of the 25th
  ACM SIGKDD International Conference on Knowledge Discovery \& Data Mining}}.
  ACM, \bibinfo{pages}{1269--1278}.
\newblock


\bibitem[\protect\citeauthoryear{Leskovec, Adamic, and Huberman}{Leskovec
  et~al\mbox{.}}{2007}]%
        {leskovec2007dynamics}
\bibfield{author}{\bibinfo{person}{Jure Leskovec}, \bibinfo{person}{Lada~A
  Adamic}, {and} \bibinfo{person}{Bernardo~A Huberman}.}
  \bibinfo{year}{2007}\natexlab{}.
\newblock \showarticletitle{The dynamics of viral marketing}.
\newblock \bibinfo{journal}{\emph{ACM Transactions on the Web (TWEB)}}
  \bibinfo{volume}{1}, \bibinfo{number}{1} (\bibinfo{year}{2007}),
  \bibinfo{pages}{5--es}.
\newblock


\bibitem[\protect\citeauthoryear{Leskovec, Huttenlocher, and
  Kleinberg}{Leskovec et~al\mbox{.}}{2010}]%
        {leskovec2010signed}
\bibfield{author}{\bibinfo{person}{Jure Leskovec}, \bibinfo{person}{Daniel
  Huttenlocher}, {and} \bibinfo{person}{Jon Kleinberg}.}
  \bibinfo{year}{2010}\natexlab{}.
\newblock \showarticletitle{Signed networks in social media}. In
  \bibinfo{booktitle}{\emph{Proceedings of the SIGCHI conference on human
  factors in computing systems}}. \bibinfo{pages}{1361--1370}.
\newblock


\bibitem[\protect\citeauthoryear{Leskovec, Kleinberg, and Faloutsos}{Leskovec
  et~al\mbox{.}}{2005}]%
        {leskovec2005graphs}
\bibfield{author}{\bibinfo{person}{Jure Leskovec}, \bibinfo{person}{Jon
  Kleinberg}, {and} \bibinfo{person}{Christos Faloutsos}.}
  \bibinfo{year}{2005}\natexlab{}.
\newblock \showarticletitle{Graphs over time: densification laws, shrinking
  diameters and possible explanations}. In
  \bibinfo{booktitle}{\emph{Proceedings of the eleventh ACM SIGKDD
  international conference on Knowledge discovery in data mining}}.
  \bibinfo{pages}{177--187}.
\newblock


\bibitem[\protect\citeauthoryear{Leskovec and Krevl}{Leskovec and
  Krevl}{2014}]%
        {snapnets}
\bibfield{author}{\bibinfo{person}{Jure Leskovec} {and} \bibinfo{person}{Andrej
  Krevl}.} \bibinfo{year}{2014}\natexlab{}.
\newblock \bibinfo{title}{{SNAP Datasets}: {Stanford} Large Network Dataset
  Collection}.
\newblock \bibinfo{howpublished}{\url{http://snap.stanford.edu/data}}.
\newblock


\bibitem[\protect\citeauthoryear{Leskovec, Lang, Dasgupta, and
  Mahoney}{Leskovec et~al\mbox{.}}{2008}]%
        {leskovec2008community}
\bibfield{author}{\bibinfo{person}{J Leskovec}, \bibinfo{person}{KJ Lang},
  \bibinfo{person}{A Dasgupta}, {and} \bibinfo{person}{MW Mahoney}.}
  \bibinfo{year}{2008}\natexlab{}.
\newblock \showarticletitle{Community structure in large networks: Natural
  cluster sizes and the absence of large well-defined clusters. ArXiv}.
\newblock \bibinfo{journal}{\emph{arXiv preprint arXiv:0810.1355}}
  (\bibinfo{year}{2008}).
\newblock


\bibitem[\protect\citeauthoryear{Leskovec, Lang, Dasgupta, and
  Mahoney}{Leskovec et~al\mbox{.}}{2009}]%
        {leskovec2009community}
\bibfield{author}{\bibinfo{person}{Jure Leskovec}, \bibinfo{person}{Kevin~J
  Lang}, \bibinfo{person}{Anirban Dasgupta}, {and} \bibinfo{person}{Michael~W
  Mahoney}.} \bibinfo{year}{2009}\natexlab{}.
\newblock \showarticletitle{Community structure in large networks: Natural
  cluster sizes and the absence of large well-defined clusters}.
\newblock \bibinfo{journal}{\emph{Internet Mathematics}} \bibinfo{volume}{6},
  \bibinfo{number}{1} (\bibinfo{year}{2009}), \bibinfo{pages}{29--123}.
\newblock


\bibitem[\protect\citeauthoryear{Levie, Huang, Bucci, Bronstein, and
  Kutyniok}{Levie et~al\mbox{.}}{2019}]%
        {levie2019transferability}
\bibfield{author}{\bibinfo{person}{Ron Levie}, \bibinfo{person}{Wei Huang},
  \bibinfo{person}{Lorenzo Bucci}, \bibinfo{person}{Michael~M Bronstein}, {and}
  \bibinfo{person}{Gitta Kutyniok}.} \bibinfo{year}{2019}\natexlab{}.
\newblock \bibinfo{title}{Transferability of spectral graph convolutional
  neural networks}.
\newblock
\newblock


\bibitem[\protect\citeauthoryear{Lewis, Agam, Argamon, Frieder, Grossman, and
  Heard}{Lewis et~al\mbox{.}}{2006}]%
        {lewis2006building}
\bibfield{author}{\bibinfo{person}{David Lewis}, \bibinfo{person}{Gady Agam},
  \bibinfo{person}{Shlomo Argamon}, \bibinfo{person}{Ophir Frieder},
  \bibinfo{person}{David Grossman}, {and} \bibinfo{person}{Jefferson Heard}.}
  \bibinfo{year}{2006}\natexlab{}.
\newblock \showarticletitle{Building a test collection for complex document
  information processing}. In \bibinfo{booktitle}{\emph{Proceedings of the 29th
  annual international ACM SIGIR conference on Research and development in
  information retrieval}}. \bibinfo{publisher}{Association for Computing
  Machinery}, \bibinfo{address}{New York, NY, USA}, \bibinfo{pages}{665--666}.
\newblock


\bibitem[\protect\citeauthoryear{Li, Yan, Li, Qu, and Radev}{Li
  et~al\mbox{.}}{2021b}]%
        {li2021unsupervised}
\bibfield{author}{\bibinfo{person}{Irene Li}, \bibinfo{person}{Vanessa Yan},
  \bibinfo{person}{Tianxiao Li}, \bibinfo{person}{Rihao Qu}, {and}
  \bibinfo{person}{Dragomir Radev}.} \bibinfo{year}{2021}\natexlab{b}.
\newblock \bibinfo{title}{Unsupervised Cross-Domain Prerequisite Chain Learning
  using Variational Graph Autoencoders}.
\newblock
\newblock


\bibitem[\protect\citeauthoryear{Li and Zitnik}{Li and Zitnik}{2021}]%
        {li2021deep}
\bibfield{author}{\bibinfo{person}{Michelle~M Li} {and}
  \bibinfo{person}{Marinka Zitnik}.} \bibinfo{year}{2021}\natexlab{}.
\newblock \bibinfo{title}{Deep Contextual Learners for Protein Networks}.
\newblock
\newblock


\bibitem[\protect\citeauthoryear{Li, Han, and Wu}{Li et~al\mbox{.}}{2018a}]%
        {li2018deeper}
\bibfield{author}{\bibinfo{person}{Qimai Li}, \bibinfo{person}{Zhichao Han},
  {and} \bibinfo{person}{Xiao-Ming Wu}.} \bibinfo{year}{2018}\natexlab{a}.
\newblock \showarticletitle{Deeper insights into graph convolutional networks
  for semi-supervised learning}. In \bibinfo{booktitle}{\emph{Thirty-Second
  AAAI conference on artificial intelligence}}. \bibinfo{publisher}{AAAI
  Press}, \bibinfo{address}{New Orleans, USA}, \bibinfo{pages}{3538--3545}.
\newblock


\bibitem[\protect\citeauthoryear{Li, Xu, Wang, and Zhong}{Li
  et~al\mbox{.}}{2021a}]%
        {li2021self}
\bibfield{author}{\bibinfo{person}{Shucheng Li}, \bibinfo{person}{Fengyuan Xu},
  \bibinfo{person}{Runchuan Wang}, {and} \bibinfo{person}{Sheng Zhong}.}
  \bibinfo{year}{2021}\natexlab{a}.
\newblock \bibinfo{title}{Self-supervised Incremental Deep Graph Learning for
  Ethereum Phishing Scam Detection}.
\newblock
\newblock


\bibitem[\protect\citeauthoryear{Li, Jin, Xu, and Tang}{Li
  et~al\mbox{.}}{2020a}]%
        {li2020deeprobust}
\bibfield{author}{\bibinfo{person}{Yaxin Li}, \bibinfo{person}{Wei Jin},
  \bibinfo{person}{Han Xu}, {and} \bibinfo{person}{Jiliang Tang}.}
  \bibinfo{year}{2020}\natexlab{a}.
\newblock \showarticletitle{Deeprobust: A pytorch library for adversarial
  attacks and defenses}.
\newblock \bibinfo{journal}{\emph{arXiv preprint arXiv:2005.06149}}
  (\bibinfo{year}{2020}).
\newblock


\bibitem[\protect\citeauthoryear{Li, Ouyang, Zhou, Shi, Zhang, and Wang}{Li
  et~al\mbox{.}}{2018b}]%
        {li2018factorizable}
\bibfield{author}{\bibinfo{person}{Yikang Li}, \bibinfo{person}{Wanli Ouyang},
  \bibinfo{person}{Bolei Zhou}, \bibinfo{person}{Jianping Shi},
  \bibinfo{person}{Chao Zhang}, {and} \bibinfo{person}{Xiaogang Wang}.}
  \bibinfo{year}{2018}\natexlab{b}.
\newblock \showarticletitle{Factorizable Net: An Efficient Subgraph-Based
  Framework for Scene Graph Generation}. In \bibinfo{booktitle}{\emph{Computer
  Vision -- ECCV 2018}}, \bibfield{editor}{\bibinfo{person}{Vittorio Ferrari},
  \bibinfo{person}{Martial Hebert}, \bibinfo{person}{Cristian Sminchisescu},
  {and} \bibinfo{person}{Yair Weiss}} (Eds.). \bibinfo{publisher}{Springer
  International Publishing}, \bibinfo{address}{Cham},
  \bibinfo{pages}{346--363}.
\newblock
\showISBNx{978-3-030-01246-5}


\bibitem[\protect\citeauthoryear{Li, Vinyals, Dyer, Pascanu, and Battaglia}{Li
  et~al\mbox{.}}{2018c}]%
        {li2018learning}
\bibfield{author}{\bibinfo{person}{Yujia Li}, \bibinfo{person}{Oriol Vinyals},
  \bibinfo{person}{Chris Dyer}, \bibinfo{person}{Razvan Pascanu}, {and}
  \bibinfo{person}{Peter Battaglia}.} \bibinfo{year}{2018}\natexlab{c}.
\newblock \bibinfo{title}{Learning deep generative models of graphs}.
\newblock
\newblock


\bibitem[\protect\citeauthoryear{Li, Yu, Shahabi, and Liu}{Li
  et~al\mbox{.}}{2017}]%
        {li2017diffusion}
\bibfield{author}{\bibinfo{person}{Yaguang Li}, \bibinfo{person}{Rose Yu},
  \bibinfo{person}{Cyrus Shahabi}, {and} \bibinfo{person}{Yan Liu}.}
  \bibinfo{year}{2017}\natexlab{}.
\newblock \bibinfo{title}{Diffusion convolutional recurrent neural network:
  Data-driven traffic forecasting}.
\newblock
\newblock


\bibitem[\protect\citeauthoryear{Li, Kumar, Headden, Yin, Wei, Zhang, and
  Yang}{Li et~al\mbox{.}}{2020b}]%
        {li-etal-2020-learn}
\bibfield{author}{\bibinfo{person}{Zheng Li}, \bibinfo{person}{Mukul Kumar},
  \bibinfo{person}{William Headden}, \bibinfo{person}{Bing Yin},
  \bibinfo{person}{Ying Wei}, \bibinfo{person}{Yu Zhang}, {and}
  \bibinfo{person}{Qiang Yang}.} \bibinfo{year}{2020}\natexlab{b}.
\newblock \showarticletitle{Learn to Cross-lingual Transfer with Meta Graph
  Learning Across Heterogeneous Languages}. In
  \bibinfo{booktitle}{\emph{Proceedings of the 2020 Conference on Empirical
  Methods in Natural Language Processing (EMNLP)}}.
  \bibinfo{publisher}{Association for Computational Linguistics},
  \bibinfo{address}{Online}, \bibinfo{pages}{2290--2301}.
\newblock


\bibitem[\protect\citeauthoryear{Lin, Cai, and Lin}{Lin et~al\mbox{.}}{2021a}]%
        {lin2021multi}
\bibfield{author}{\bibinfo{person}{Jinke Lin}, \bibinfo{person}{Qingling Cai},
  {and} \bibinfo{person}{Manying Lin}.} \bibinfo{year}{2021}\natexlab{a}.
\newblock \showarticletitle{Multi-Label Classification of Fundus Images With
  Graph Convolutional Network and Self-Supervised Learning}.
\newblock \bibinfo{journal}{\emph{IEEE Signal Processing Letters}}
  \bibinfo{volume}{28} (\bibinfo{year}{2021}), \bibinfo{pages}{454--458}.
\newblock


\bibitem[\protect\citeauthoryear{Lin, Zhu, Shu, Zhu, Ye, Shi, Min, Li, Yuan,
  and Shao}{Lin et~al\mbox{.}}{2021b}]%
        {lin2021altered}
\bibfield{author}{\bibinfo{person}{Qi Lin}, \bibinfo{person}{Fei-Ying Zhu},
  \bibinfo{person}{Yong-Qiang Shu}, \bibinfo{person}{Pei-Wen Zhu},
  \bibinfo{person}{Lei Ye}, \bibinfo{person}{Wen-Qing Shi},
  \bibinfo{person}{You-Lan Min}, \bibinfo{person}{Biao Li},
  \bibinfo{person}{Qing Yuan}, {and} \bibinfo{person}{Yi Shao}.}
  \bibinfo{year}{2021}\natexlab{b}.
\newblock \showarticletitle{Altered brain network centrality in middle-aged
  patients with retinitis pigmentosa: A resting-state functional magnetic
  resonance imaging study}.
\newblock \bibinfo{journal}{\emph{Brain and Behavior}} \bibinfo{volume}{11},
  \bibinfo{number}{2} (\bibinfo{year}{2021}), \bibinfo{pages}{e01983}.
\newblock


\bibitem[\protect\citeauthoryear{Linmei, Yang, Shi, Ji, and Li}{Linmei
  et~al\mbox{.}}{2019}]%
        {linmei-etal-2019-heterogeneous}
\bibfield{author}{\bibinfo{person}{Hu Linmei}, \bibinfo{person}{Tianchi Yang},
  \bibinfo{person}{Chuan Shi}, \bibinfo{person}{Houye Ji}, {and}
  \bibinfo{person}{Xiaoli Li}.} \bibinfo{year}{2019}\natexlab{}.
\newblock \showarticletitle{Heterogeneous Graph Attention Networks for
  Semi-supervised Short Text Classification}. In
  \bibinfo{booktitle}{\emph{Proceedings of the 2019 Conference on Empirical
  Methods in Natural Language Processing and the 9th International Joint
  Conference on Natural Language Processing (EMNLP-IJCNLP)}}.
  \bibinfo{publisher}{Association for Computational Linguistics},
  \bibinfo{address}{Hong Kong, China}, \bibinfo{pages}{4821--4830}.
\newblock


\bibitem[\protect\citeauthoryear{Liu, Tan, Li, Yang, Zhou, and Hu}{Liu
  et~al\mbox{.}}{2019}]%
        {liu2019single}
\bibfield{author}{\bibinfo{person}{Ninghao Liu}, \bibinfo{person}{Qiaoyu Tan},
  \bibinfo{person}{Yuening Li}, \bibinfo{person}{Hongxia Yang},
  \bibinfo{person}{Jingren Zhou}, {and} \bibinfo{person}{Xia Hu}.}
  \bibinfo{year}{2019}\natexlab{}.
\newblock \showarticletitle{Is a single vector enough? exploring node polysemy
  for network embedding}. In \bibinfo{booktitle}{\emph{Proceedings of the 25th
  ACM SIGKDD International Conference on Knowledge Discovery \& Data Mining}}.
  \bibinfo{publisher}{Association for Computing Machinery},
  \bibinfo{address}{New York, NY, USA}, \bibinfo{pages}{932--940}.
\newblock


\bibitem[\protect\citeauthoryear{Liu, Zhang, Hou, Mian, Wang, Zhang, and
  Tang}{Liu et~al\mbox{.}}{2021b}]%
        {liu2021self}
\bibfield{author}{\bibinfo{person}{Xiao Liu}, \bibinfo{person}{Fanjin Zhang},
  \bibinfo{person}{Zhenyu Hou}, \bibinfo{person}{Li Mian},
  \bibinfo{person}{Zhaoyu Wang}, \bibinfo{person}{Jing Zhang}, {and}
  \bibinfo{person}{Jie Tang}.} \bibinfo{year}{2021}\natexlab{b}.
\newblock \showarticletitle{Self-supervised learning: Generative or
  contrastive}.
\newblock \bibinfo{journal}{\emph{IEEE Transactions on Knowledge and Data
  Engineering}} \bibinfo{volume}{Early Access}, \bibinfo{number}{Early Access}
  (\bibinfo{year}{2021}), \bibinfo{pages}{1--1}.
\newblock
\urldef\tempurl%
\url{https://doi.org/10.1109/TKDE.2021.3090866}
\showDOI{\tempurl}


\bibitem[\protect\citeauthoryear{Liu, Pan, Jin, Zhou, Xia, and Yu}{Liu
  et~al\mbox{.}}{2021a}]%
        {liu2021graph}
\bibfield{author}{\bibinfo{person}{Yixin Liu}, \bibinfo{person}{Shirui Pan},
  \bibinfo{person}{Ming Jin}, \bibinfo{person}{Chuan Zhou},
  \bibinfo{person}{Feng Xia}, {and} \bibinfo{person}{Philip~S Yu}.}
  \bibinfo{year}{2021}\natexlab{a}.
\newblock \bibinfo{title}{Graph self-supervised learning: A survey}.
\newblock
\newblock


\bibitem[\protect\citeauthoryear{Loukas}{Loukas}{2019}]%
        {loukas2019graph}
\bibfield{author}{\bibinfo{person}{Andreas Loukas}.}
  \bibinfo{year}{2019}\natexlab{}.
\newblock \bibinfo{title}{What graph neural networks cannot learn: depth vs
  width}.
\newblock
\newblock


\bibitem[\protect\citeauthoryear{Maiya}{Maiya}{2020}]%
        {maiya2020ktrain}
\bibfield{author}{\bibinfo{person}{Arun~S. Maiya}.}
  \bibinfo{year}{2020}\natexlab{}.
\newblock \showarticletitle{ktrain: A Low-Code Library for Augmented Machine
  Learning}.
\newblock \bibinfo{journal}{\emph{arXiv preprint arXiv:2004.10703}}
  (\bibinfo{year}{2020}).
\newblock
\showeprint[arxiv]{2004.10703}~[cs.LG]


\bibitem[\protect\citeauthoryear{Manessi and Rozza}{Manessi and Rozza}{2021}]%
        {manessi2021graph}
\bibfield{author}{\bibinfo{person}{Franco Manessi} {and}
  \bibinfo{person}{Alessandro Rozza}.} \bibinfo{year}{2021}\natexlab{}.
\newblock \showarticletitle{Graph-based neural network models with multiple
  self-supervised auxiliary tasks}.
\newblock \bibinfo{journal}{\emph{Pattern Recognition Letters}}
  \bibinfo{volume}{148} (\bibinfo{year}{2021}), \bibinfo{pages}{15--21}.
\newblock


\bibitem[\protect\citeauthoryear{Marcheggiani and Titov}{Marcheggiani and
  Titov}{2017}]%
        {marcheggiani2017encoding}
\bibfield{author}{\bibinfo{person}{Diego Marcheggiani} {and}
  \bibinfo{person}{Ivan Titov}.} \bibinfo{year}{2017}\natexlab{}.
\newblock \bibinfo{title}{Encoding sentences with graph convolutional networks
  for semantic role labeling}.
\newblock
\newblock


\bibitem[\protect\citeauthoryear{Maron, Ben-Hamu, Shamir, and Lipman}{Maron
  et~al\mbox{.}}{2018}]%
        {maron2018invariant}
\bibfield{author}{\bibinfo{person}{Haggai Maron}, \bibinfo{person}{Heli
  Ben-Hamu}, \bibinfo{person}{Nadav Shamir}, {and} \bibinfo{person}{Yaron
  Lipman}.} \bibinfo{year}{2018}\natexlab{}.
\newblock \bibinfo{title}{Invariant and equivariant graph networks}.
\newblock
\newblock


\bibitem[\protect\citeauthoryear{Maron, Fetaya, Segol, and Lipman}{Maron
  et~al\mbox{.}}{2019}]%
        {maron2019universality}
\bibfield{author}{\bibinfo{person}{Haggai Maron}, \bibinfo{person}{Ethan
  Fetaya}, \bibinfo{person}{Nimrod Segol}, {and} \bibinfo{person}{Yaron
  Lipman}.} \bibinfo{year}{2019}\natexlab{}.
\newblock \showarticletitle{On the universality of invariant networks}. In
  \bibinfo{booktitle}{\emph{International conference on machine learning}}.
  PMLR, \bibinfo{pages}{4363--4371}.
\newblock


\bibitem[\protect\citeauthoryear{McAuley, Pandey, and Leskovec}{McAuley
  et~al\mbox{.}}{2015a}]%
        {mcauley2015inferring}
\bibfield{author}{\bibinfo{person}{Julian McAuley}, \bibinfo{person}{Rahul
  Pandey}, {and} \bibinfo{person}{Jure Leskovec}.}
  \bibinfo{year}{2015}\natexlab{a}.
\newblock \showarticletitle{Inferring networks of substitutable and
  complementary products}. In \bibinfo{booktitle}{\emph{Proceedings of the 21th
  ACM SIGKDD international conference on knowledge discovery and data mining}}.
  \bibinfo{publisher}{Association for Computing Machinery},
  \bibinfo{address}{New York, NY, USA}, \bibinfo{pages}{785--794}.
\newblock


\bibitem[\protect\citeauthoryear{McAuley, Targett, Shi, and Van
  Den~Hengel}{McAuley et~al\mbox{.}}{2015b}]%
        {mcauley2015image}
\bibfield{author}{\bibinfo{person}{Julian McAuley},
  \bibinfo{person}{Christopher Targett}, \bibinfo{person}{Qinfeng Shi}, {and}
  \bibinfo{person}{Anton Van Den~Hengel}.} \bibinfo{year}{2015}\natexlab{b}.
\newblock \showarticletitle{Image-based recommendations on styles and
  substitutes}. In \bibinfo{booktitle}{\emph{Proceedings of the 38th
  international ACM SIGIR conference on research and development in information
  retrieval}}. \bibinfo{pages}{43--52}.
\newblock


\bibitem[\protect\citeauthoryear{McAuley and Leskovec}{McAuley and
  Leskovec}{2012}]%
        {mcauley2012learning}
\bibfield{author}{\bibinfo{person}{Julian~J McAuley} {and}
  \bibinfo{person}{Jure Leskovec}.} \bibinfo{year}{2012}\natexlab{}.
\newblock \showarticletitle{Learning to discover social circles in ego
  networks.}. In \bibinfo{booktitle}{\emph{NIPS}}, Vol.~\bibinfo{volume}{2012}.
  Citeseer, \bibinfo{pages}{548--56}.
\newblock


\bibitem[\protect\citeauthoryear{McCallum, Nigam, Rennie, and Seymore}{McCallum
  et~al\mbox{.}}{2000}]%
        {mccallum2000automating}
\bibfield{author}{\bibinfo{person}{Andrew~Kachites McCallum},
  \bibinfo{person}{Kamal Nigam}, \bibinfo{person}{Jason Rennie}, {and}
  \bibinfo{person}{Kristie Seymore}.} \bibinfo{year}{2000}\natexlab{}.
\newblock \showarticletitle{Automating the construction of internet portals
  with machine learning}.
\newblock \bibinfo{journal}{\emph{Information Retrieval}} \bibinfo{volume}{3},
  \bibinfo{number}{2} (\bibinfo{year}{2000}), \bibinfo{pages}{127--163}.
\newblock


\bibitem[\protect\citeauthoryear{McPherson, Smith-Lovin, and Cook}{McPherson
  et~al\mbox{.}}{2001}]%
        {mcpherson2001birds}
\bibfield{author}{\bibinfo{person}{Miller McPherson}, \bibinfo{person}{Lynn
  Smith-Lovin}, {and} \bibinfo{person}{James~M Cook}.}
  \bibinfo{year}{2001}\natexlab{}.
\newblock \showarticletitle{Birds of a feather: Homophily in social networks}.
\newblock \bibinfo{journal}{\emph{Annual review of sociology}}
  \bibinfo{volume}{27}, \bibinfo{number}{1} (\bibinfo{year}{2001}),
  \bibinfo{pages}{415--444}.
\newblock


\bibitem[\protect\citeauthoryear{Mislove, Koppula, Gummadi, Druschel, and
  Bhattacharjee}{Mislove et~al\mbox{.}}{2008}]%
        {mislove2008flickr}
\bibfield{author}{\bibinfo{person}{Alan Mislove}, \bibinfo{person}{Hema~Swetha
  Koppula}, \bibinfo{person}{Krishna~P. Gummadi}, \bibinfo{person}{Peter
  Druschel}, {and} \bibinfo{person}{Bobby Bhattacharjee}.}
  \bibinfo{year}{2008}\natexlab{}.
\newblock \showarticletitle{Growth of the Flickr Social Network}. In
  \bibinfo{booktitle}{\emph{Proceedings of the First Workshop on Online Social
  Networks}} (Seattle, WA, USA) \emph{(\bibinfo{series}{WOSN '08})}.
  \bibinfo{publisher}{Association for Computing Machinery},
  \bibinfo{address}{New York, NY, USA}, \bibinfo{pages}{25–30}.
\newblock
\showISBNx{9781605581828}
\urldef\tempurl%
\url{https://doi.org/10.1145/1397735.1397742}
\showDOI{\tempurl}


\bibitem[\protect\citeauthoryear{Monti, Bronstein, and Bresson}{Monti
  et~al\mbox{.}}{2017}]%
        {monti2017geometric}
\bibfield{author}{\bibinfo{person}{Federico Monti}, \bibinfo{person}{Michael~M
  Bronstein}, {and} \bibinfo{person}{Xavier Bresson}.}
  \bibinfo{year}{2017}\natexlab{}.
\newblock \bibinfo{title}{Geometric matrix completion with recurrent
  multi-graph neural networks}.
\newblock
\newblock


\bibitem[\protect\citeauthoryear{Morris, Ritzert, Fey, Hamilton, Lenssen,
  Rattan, and Grohe}{Morris et~al\mbox{.}}{2019}]%
        {morris2019weisfeiler}
\bibfield{author}{\bibinfo{person}{Christopher Morris}, \bibinfo{person}{Martin
  Ritzert}, \bibinfo{person}{Matthias Fey}, \bibinfo{person}{William~L
  Hamilton}, \bibinfo{person}{Jan~Eric Lenssen}, \bibinfo{person}{Gaurav
  Rattan}, {and} \bibinfo{person}{Martin Grohe}.}
  \bibinfo{year}{2019}\natexlab{}.
\newblock \showarticletitle{Weisfeiler and leman go neural: Higher-order graph
  neural networks}.
\newblock \bibinfo{journal}{\emph{Proceedings of the AAAI Conference on
  Artificial Intelligence}} \bibinfo{volume}{33}, \bibinfo{number}{01}
  (\bibinfo{year}{2019}), \bibinfo{pages}{4602--4609}.
\newblock


\bibitem[\protect\citeauthoryear{Newman}{Newman}{2005}]%
        {newman2005measure}
\bibfield{author}{\bibinfo{person}{Mark~EJ Newman}.}
  \bibinfo{year}{2005}\natexlab{}.
\newblock \showarticletitle{A measure of betweenness centrality based on random
  walks}.
\newblock \bibinfo{journal}{\emph{Social networks}} \bibinfo{volume}{27},
  \bibinfo{number}{1} (\bibinfo{year}{2005}), \bibinfo{pages}{39--54}.
\newblock


\bibitem[\protect\citeauthoryear{Nt and Maehara}{Nt and Maehara}{2019}]%
        {nt2019revisiting}
\bibfield{author}{\bibinfo{person}{Hoang Nt} {and} \bibinfo{person}{Takanori
  Maehara}.} \bibinfo{year}{2019}\natexlab{}.
\newblock \bibinfo{title}{Revisiting graph neural networks: All we have is
  low-pass filters}.
\newblock
\newblock


\bibitem[\protect\citeauthoryear{Okuda, Satoh, Sato, and Kidawara}{Okuda
  et~al\mbox{.}}{2021}]%
        {okuda2021unsupervised}
\bibfield{author}{\bibinfo{person}{Makoto Okuda}, \bibinfo{person}{Shin’ichi
  Satoh}, \bibinfo{person}{Yoichi Sato}, {and} \bibinfo{person}{Yutaka
  Kidawara}.} \bibinfo{year}{2021}\natexlab{}.
\newblock \showarticletitle{Unsupervised Common Particular Object Discovery and
  Localization by Analyzing a Match Graph}. In \bibinfo{booktitle}{\emph{ICASSP
  2021-2021 IEEE International Conference on Acoustics, Speech and Signal
  Processing (ICASSP)}}. IEEE, \bibinfo{pages}{1540--1544}.
\newblock


\bibitem[\protect\citeauthoryear{Oono and Suzuki}{Oono and Suzuki}{2019}]%
        {oono2019graph}
\bibfield{author}{\bibinfo{person}{Kenta Oono} {and} \bibinfo{person}{Taiji
  Suzuki}.} \bibinfo{year}{2019}\natexlab{}.
\newblock \bibinfo{title}{Graph neural networks exponentially lose expressive
  power for node classification}.
\newblock
\newblock


\bibitem[\protect\citeauthoryear{Oord, Li, and Vinyals}{Oord
  et~al\mbox{.}}{2018}]%
        {oord2018representation}
\bibfield{author}{\bibinfo{person}{Aaron van~den Oord}, \bibinfo{person}{Yazhe
  Li}, {and} \bibinfo{person}{Oriol Vinyals}.} \bibinfo{year}{2018}\natexlab{}.
\newblock \bibinfo{title}{Representation learning with contrastive predictive
  coding}.
\newblock
\newblock


\bibitem[\protect\citeauthoryear{Opolka, Solomon, Cangea, Veli{\v{c}}kovi{\'c},
  Li{\`o}, and Hjelm}{Opolka et~al\mbox{.}}{2019}]%
        {opolka2019spatio}
\bibfield{author}{\bibinfo{person}{Felix~L Opolka}, \bibinfo{person}{Aaron
  Solomon}, \bibinfo{person}{C{\u{a}}t{\u{a}}lina Cangea},
  \bibinfo{person}{Petar Veli{\v{c}}kovi{\'c}}, \bibinfo{person}{Pietro
  Li{\`o}}, {and} \bibinfo{person}{R~Devon Hjelm}.}
  \bibinfo{year}{2019}\natexlab{}.
\newblock \bibinfo{title}{Spatio-temporal deep graph infomax}.
\newblock
\newblock


\bibitem[\protect\citeauthoryear{Ou, Cui, Pei, Zhang, and Zhu}{Ou
  et~al\mbox{.}}{2016}]%
        {ou2016asymmetric}
\bibfield{author}{\bibinfo{person}{Mingdong Ou}, \bibinfo{person}{Peng Cui},
  \bibinfo{person}{Jian Pei}, \bibinfo{person}{Ziwei Zhang}, {and}
  \bibinfo{person}{Wenwu Zhu}.} \bibinfo{year}{2016}\natexlab{}.
\newblock \showarticletitle{Asymmetric transitivity preserving graph
  embedding}. In \bibinfo{booktitle}{\emph{Proceedings of the 22nd ACM SIGKDD
  international conference on Knowledge discovery and data mining}}.
  \bibinfo{publisher}{Association for Computing Machinery},
  \bibinfo{address}{New York, NY, USA}, \bibinfo{pages}{1105--1114}.
\newblock


\bibitem[\protect\citeauthoryear{Ozaki, Shimbo, Komachi, and Matsumoto}{Ozaki
  et~al\mbox{.}}{2011}]%
        {ozaki2011using}
\bibfield{author}{\bibinfo{person}{Kohei Ozaki}, \bibinfo{person}{Masashi
  Shimbo}, \bibinfo{person}{Mamoru Komachi}, {and} \bibinfo{person}{Yuji
  Matsumoto}.} \bibinfo{year}{2011}\natexlab{}.
\newblock \showarticletitle{Using the mutual k-nearest neighbor graphs for
  semi-supervised classification on natural language data}. In
  \bibinfo{booktitle}{\emph{Proceedings of the fifteenth conference on
  computational natural language learning}}. \bibinfo{pages}{154--162}.
\newblock


\bibitem[\protect\citeauthoryear{PaddlePaddle}{PaddlePaddle}{2021}]%
        {src3}
\bibfield{author}{\bibinfo{person}{PaddlePaddle}.}
  \bibinfo{year}{2021}\natexlab{}.
\newblock \bibinfo{title}{{PGL}}.
\newblock
\newblock
\urldef\tempurl%
\url{https://github.com/PaddlePaddle/PGL}
\showURL{%
\tempurl}


\bibitem[\protect\citeauthoryear{Page, Brin, Motwani, and Winograd}{Page
  et~al\mbox{.}}{1999}]%
        {page1999pagerank}
\bibfield{author}{\bibinfo{person}{Lawrence Page}, \bibinfo{person}{Sergey
  Brin}, \bibinfo{person}{Rajeev Motwani}, {and} \bibinfo{person}{Terry
  Winograd}.} \bibinfo{year}{1999}\natexlab{}.
\newblock \bibinfo{booktitle}{\emph{The PageRank citation ranking: Bringing
  order to the web.}}
\newblock \bibinfo{type}{{T}echnical {R}eport}. \bibinfo{institution}{Stanford
  InfoLab}.
\newblock


\bibitem[\protect\citeauthoryear{Pan, Hu, Fung, Long, Jiang, and Zhang}{Pan
  et~al\mbox{.}}{2019}]%
        {pan2019learning}
\bibfield{author}{\bibinfo{person}{Shirui Pan}, \bibinfo{person}{Ruiqi Hu},
  \bibinfo{person}{Sai-fu Fung}, \bibinfo{person}{Guodong Long},
  \bibinfo{person}{Jing Jiang}, {and} \bibinfo{person}{Chengqi Zhang}.}
  \bibinfo{year}{2019}\natexlab{}.
\newblock \showarticletitle{Learning graph embedding with adversarial training
  methods}.
\newblock \bibinfo{journal}{\emph{IEEE transactions on cybernetics}}
  \bibinfo{volume}{50}, \bibinfo{number}{6} (\bibinfo{year}{2019}),
  \bibinfo{pages}{2475--2487}.
\newblock


\bibitem[\protect\citeauthoryear{Pan, Hu, Long, Jiang, Yao, and Zhang}{Pan
  et~al\mbox{.}}{2018}]%
        {PanAdversarially}
\bibfield{author}{\bibinfo{person}{Shirui Pan}, \bibinfo{person}{Ruiqi Hu},
  \bibinfo{person}{Guodong Long}, \bibinfo{person}{Jing Jiang},
  \bibinfo{person}{Lina Yao}, {and} \bibinfo{person}{Chengqi Zhang}.}
  \bibinfo{year}{2018}\natexlab{}.
\newblock \showarticletitle{Adversarially Regularized Graph Autoencoder for
  Graph Embedding}. In \bibinfo{booktitle}{\emph{Proceedings of the
  Twenty-Seventh International Joint Conference on Artificial Intelligence,
  {IJCAI-18}}}. \bibinfo{publisher}{International Joint Conferences on
  Artificial Intelligence Organization}, \bibinfo{pages}{2609--2615}.
\newblock


\bibitem[\protect\citeauthoryear{Paranjape, Benson, and Leskovec}{Paranjape
  et~al\mbox{.}}{2017}]%
        {paranjape2017motifs}
\bibfield{author}{\bibinfo{person}{Ashwin Paranjape}, \bibinfo{person}{Austin~R
  Benson}, {and} \bibinfo{person}{Jure Leskovec}.}
  \bibinfo{year}{2017}\natexlab{}.
\newblock \showarticletitle{Motifs in temporal networks}. In
  \bibinfo{booktitle}{\emph{Proceedings of the tenth ACM international
  conference on web search and data mining}}. \bibinfo{pages}{601--610}.
\newblock


\bibitem[\protect\citeauthoryear{Park, Kim, Han, and Yu}{Park
  et~al\mbox{.}}{2020}]%
        {park2020unsupervised}
\bibfield{author}{\bibinfo{person}{Chanyoung Park}, \bibinfo{person}{Donghyun
  Kim}, \bibinfo{person}{Jiawei Han}, {and} \bibinfo{person}{Hwanjo Yu}.}
  \bibinfo{year}{2020}\natexlab{}.
\newblock \showarticletitle{Unsupervised attributed multiplex network
  embedding}. In \bibinfo{booktitle}{\emph{Proceedings of the AAAI Conference
  on Artificial Intelligence}}, Vol.~\bibinfo{volume}{34}.
  \bibinfo{pages}{5371--5378}.
\newblock


\bibitem[\protect\citeauthoryear{Park, Cho, Chang, and Choi}{Park
  et~al\mbox{.}}{2021}]%
        {park2021unsupervised}
\bibfield{author}{\bibinfo{person}{Jiwoong Park}, \bibinfo{person}{Junho Cho},
  \bibinfo{person}{Hyung~Jin Chang}, {and} \bibinfo{person}{Jin~Young Choi}.}
  \bibinfo{year}{2021}\natexlab{}.
\newblock \showarticletitle{Unsupervised Hyperbolic Representation Learning via
  Message Passing Auto-Encoders}. In \bibinfo{booktitle}{\emph{Proceedings of
  the IEEE/CVF Conference on Computer Vision and Pattern Recognition}}.
  \bibinfo{pages}{5516--5526}.
\newblock


\bibitem[\protect\citeauthoryear{Park, Lee, Chang, Lee, and Choi}{Park
  et~al\mbox{.}}{2019}]%
        {park2019symmetric}
\bibfield{author}{\bibinfo{person}{Jiwoong Park}, \bibinfo{person}{Minsik Lee},
  \bibinfo{person}{Hyung~Jin Chang}, \bibinfo{person}{Kyuewang Lee}, {and}
  \bibinfo{person}{Jin~Young Choi}.} \bibinfo{year}{2019}\natexlab{}.
\newblock \showarticletitle{Symmetric graph convolutional autoencoder for
  unsupervised graph representation learning}. In
  \bibinfo{booktitle}{\emph{Proceedings of the IEEE/CVF International
  Conference on Computer Vision}}. \bibinfo{pages}{6519--6528}.
\newblock


\bibitem[\protect\citeauthoryear{Peng, Dong, Luo, Wu, and Zheng}{Peng
  et~al\mbox{.}}{2020a}]%
        {peng2020self}
\bibfield{author}{\bibinfo{person}{Zhen Peng}, \bibinfo{person}{Yixiang Dong},
  \bibinfo{person}{Minnan Luo}, \bibinfo{person}{Xiao-Ming Wu}, {and}
  \bibinfo{person}{Qinghua Zheng}.} \bibinfo{year}{2020}\natexlab{a}.
\newblock \bibinfo{title}{Self-supervised graph representation learning via
  global context prediction}.
\newblock
\newblock


\bibitem[\protect\citeauthoryear{Peng, Huang, Luo, Zheng, Rong, Xu, and
  Huang}{Peng et~al\mbox{.}}{2020b}]%
        {peng2020graph}
\bibfield{author}{\bibinfo{person}{Zhen Peng}, \bibinfo{person}{Wenbing Huang},
  \bibinfo{person}{Minnan Luo}, \bibinfo{person}{Qinghua Zheng},
  \bibinfo{person}{Yu Rong}, \bibinfo{person}{Tingyang Xu}, {and}
  \bibinfo{person}{Junzhou Huang}.} \bibinfo{year}{2020}\natexlab{b}.
\newblock \showarticletitle{Graph representation learning via graphical mutual
  information maximization}. In \bibinfo{booktitle}{\emph{Proceedings of The
  Web Conference 2020}}. \bibinfo{pages}{259--270}.
\newblock


\bibitem[\protect\citeauthoryear{Perozzi, Al-Rfou, and Skiena}{Perozzi
  et~al\mbox{.}}{2014}]%
        {perozzi2014deepwalk}
\bibfield{author}{\bibinfo{person}{Bryan Perozzi}, \bibinfo{person}{Rami
  Al-Rfou}, {and} \bibinfo{person}{Steven Skiena}.}
  \bibinfo{year}{2014}\natexlab{}.
\newblock \showarticletitle{Deepwalk: Online learning of social
  representations}. In \bibinfo{booktitle}{\emph{Proceedings of the 20th ACM
  SIGKDD international conference on Knowledge discovery and data mining}}.
  \bibinfo{publisher}{Association for Computing Machinery},
  \bibinfo{address}{New York, NY, USA}, \bibinfo{pages}{701--710}.
\newblock


\bibitem[\protect\citeauthoryear{Prakash and Nithya}{Prakash and
  Nithya}{2014}]%
        {prakash2014survey}
\bibfield{author}{\bibinfo{person}{V~Jothi Prakash} {and}
  \bibinfo{person}{Dr~LM Nithya}.} \bibinfo{year}{2014}\natexlab{}.
\newblock \bibinfo{title}{A survey on semi-supervised learning techniques}.
\newblock
\newblock


\bibitem[\protect\citeauthoryear{Qiu, Chen, Dong, Zhang, Yang, Ding, Wang, and
  Tang}{Qiu et~al\mbox{.}}{2020}]%
        {qiu2020gcc}
\bibfield{author}{\bibinfo{person}{Jiezhong Qiu}, \bibinfo{person}{Qibin Chen},
  \bibinfo{person}{Yuxiao Dong}, \bibinfo{person}{Jing Zhang},
  \bibinfo{person}{Hongxia Yang}, \bibinfo{person}{Ming Ding},
  \bibinfo{person}{Kuansan Wang}, {and} \bibinfo{person}{Jie Tang}.}
  \bibinfo{year}{2020}\natexlab{}.
\newblock \showarticletitle{Gcc: Graph contrastive coding for graph neural
  network pre-training}. In \bibinfo{booktitle}{\emph{Proceedings of the 26th
  ACM SIGKDD International Conference on Knowledge Discovery \& Data Mining}}.
  \bibinfo{publisher}{Association for Computing Machinery},
  \bibinfo{address}{New York, NY, USA}, \bibinfo{pages}{1150--1160}.
\newblock


\bibitem[\protect\citeauthoryear{Qiu, Tang, Ma, Dong, Wang, and Tang}{Qiu
  et~al\mbox{.}}{2018}]%
        {qiu2018deepinf}
\bibfield{author}{\bibinfo{person}{Jiezhong Qiu}, \bibinfo{person}{Jian Tang},
  \bibinfo{person}{Hao Ma}, \bibinfo{person}{Yuxiao Dong},
  \bibinfo{person}{Kuansan Wang}, {and} \bibinfo{person}{Jie Tang}.}
  \bibinfo{year}{2018}\natexlab{}.
\newblock \showarticletitle{Deepinf: Social influence prediction with deep
  learning}. In \bibinfo{booktitle}{\emph{Proceedings of the 24th ACM SIGKDD
  International Conference on Knowledge Discovery \& Data Mining}}.
  \bibinfo{publisher}{Association for Computing Machinery},
  \bibinfo{address}{New York, NY, USA}, \bibinfo{pages}{2110--2119}.
\newblock


\bibitem[\protect\citeauthoryear{Ren, Liu, Huang, Dai, Bo, and Zhang}{Ren
  et~al\mbox{.}}{2020}]%
        {ren2020hdgi}
\bibfield{author}{\bibinfo{person}{Yuxiang Ren}, \bibinfo{person}{Bo Liu},
  \bibinfo{person}{Chao Huang}, \bibinfo{person}{Peng Dai},
  \bibinfo{person}{Liefeng Bo}, {and} \bibinfo{person}{Jiawei Zhang}.}
  \bibinfo{year}{2020}\natexlab{}.
\newblock \showarticletitle{HDGI: An Unsupervised Graph Neural Network for
  Representation Learning in Heterogeneous Graph}. In
  \bibinfo{booktitle}{\emph{AAAI Workshop}}.
\newblock


\bibitem[\protect\citeauthoryear{Rohban and Rabiee}{Rohban and Rabiee}{2012}]%
        {rohban2012supervised}
\bibfield{author}{\bibinfo{person}{Mohammad~Hossein Rohban} {and}
  \bibinfo{person}{Hamid~R Rabiee}.} \bibinfo{year}{2012}\natexlab{}.
\newblock \showarticletitle{Supervised neighborhood graph construction for
  semi-supervised classification}.
\newblock \bibinfo{journal}{\emph{Pattern Recognition}} \bibinfo{volume}{45},
  \bibinfo{number}{4} (\bibinfo{year}{2012}), \bibinfo{pages}{1363--1372}.
\newblock


\bibitem[\protect\citeauthoryear{Rong, Bian, Xu, Xie, Wei, Huang, and
  Huang}{Rong et~al\mbox{.}}{2020}]%
        {rong2020self}
\bibfield{author}{\bibinfo{person}{Yu Rong}, \bibinfo{person}{Yatao Bian},
  \bibinfo{person}{Tingyang Xu}, \bibinfo{person}{Weiyang Xie},
  \bibinfo{person}{Ying Wei}, \bibinfo{person}{Wenbing Huang}, {and}
  \bibinfo{person}{Junzhou Huang}.} \bibinfo{year}{2020}\natexlab{}.
\newblock \bibinfo{title}{Self-supervised graph transformer on large-scale
  molecular data}.
\newblock
\newblock


\bibitem[\protect\citeauthoryear{Roweis and Saul}{Roweis and Saul}{2000}]%
        {roweis2000nonlinear}
\bibfield{author}{\bibinfo{person}{Sam~T Roweis} {and}
  \bibinfo{person}{Lawrence~K Saul}.} \bibinfo{year}{2000}\natexlab{}.
\newblock \showarticletitle{Nonlinear dimensionality reduction by locally
  linear embedding}.
\newblock \bibinfo{journal}{\emph{science}} \bibinfo{volume}{290},
  \bibinfo{number}{5500} (\bibinfo{year}{2000}), \bibinfo{pages}{2323--2326}.
\newblock


\bibitem[\protect\citeauthoryear{Rozemberczki, Scherer, He, Panagopoulos,
  Riedel, Astefanoaei, Kiss, Beres, Lopez, Collignon, and Sarkar}{Rozemberczki
  et~al\mbox{.}}{2021}]%
        {rozemberczki2021pytorch}
\bibfield{author}{\bibinfo{person}{Benedek Rozemberczki}, \bibinfo{person}{Paul
  Scherer}, \bibinfo{person}{Yixuan He}, \bibinfo{person}{George Panagopoulos},
  \bibinfo{person}{Alexander Riedel}, \bibinfo{person}{Maria Astefanoaei},
  \bibinfo{person}{Oliver Kiss}, \bibinfo{person}{Ferenc Beres},
  \bibinfo{person}{Guzman Lopez}, \bibinfo{person}{Nicolas Collignon}, {and}
  \bibinfo{person}{Rik Sarkar}.} \bibinfo{year}{2021}\natexlab{}.
\newblock \bibinfo{title}{{PyTorch Geometric Temporal: Spatiotemporal Signal
  Processing with Neural Machine Learning Models}}.
\newblock
\newblock
\showeprint{arXiv:2104.07788}


\bibitem[\protect\citeauthoryear{Ruiz, Chamon, and Ribeiro}{Ruiz
  et~al\mbox{.}}{2020}]%
        {ruiz2020graphon}
\bibfield{author}{\bibinfo{person}{Luana Ruiz}, \bibinfo{person}{Luiz Chamon},
  {and} \bibinfo{person}{Alejandro Ribeiro}.} \bibinfo{year}{2020}\natexlab{}.
\newblock \showarticletitle{Graphon neural networks and the transferability of
  graph neural networks}.
\newblock \bibinfo{journal}{\emph{Advances in Neural Information Processing
  Systems}}  \bibinfo{volume}{33} (\bibinfo{year}{2020}).
\newblock


\bibitem[\protect\citeauthoryear{Sato}{Sato}{2020}]%
        {Sato}
\bibfield{author}{\bibinfo{person}{Ryoma Sato}.}
  \bibinfo{year}{2020}\natexlab{}.
\newblock \bibinfo{title}{A Survey on The Expressive Power of Graph Neural
  Networks}.
\newblock
\newblock
\showeprint[arxiv]{2003.04078}~[cs.LG]


\bibitem[\protect\citeauthoryear{Scarselli, Gori, Tsoi, Hagenbuchner, and
  Monfardini}{Scarselli et~al\mbox{.}}{2008}]%
        {scarselli2008graph}
\bibfield{author}{\bibinfo{person}{Franco Scarselli}, \bibinfo{person}{Marco
  Gori}, \bibinfo{person}{Ah~Chung Tsoi}, \bibinfo{person}{Markus
  Hagenbuchner}, {and} \bibinfo{person}{Gabriele Monfardini}.}
  \bibinfo{year}{2008}\natexlab{}.
\newblock \showarticletitle{The graph neural network model}.
\newblock \bibinfo{journal}{\emph{IEEE transactions on neural networks}}
  \bibinfo{volume}{20}, \bibinfo{number}{1} (\bibinfo{year}{2008}),
  \bibinfo{pages}{61--80}.
\newblock


\bibitem[\protect\citeauthoryear{Scarselli, Tsoi, and Hagenbuchner}{Scarselli
  et~al\mbox{.}}{2018}]%
        {scarselli2018vapnik}
\bibfield{author}{\bibinfo{person}{Franco Scarselli}, \bibinfo{person}{Ah~Chung
  Tsoi}, {and} \bibinfo{person}{Markus Hagenbuchner}.}
  \bibinfo{year}{2018}\natexlab{}.
\newblock \showarticletitle{The Vapnik--Chervonenkis dimension of graph and
  recursive neural networks}.
\newblock \bibinfo{journal}{\emph{Neural Networks}}  \bibinfo{volume}{108}
  (\bibinfo{year}{2018}), \bibinfo{pages}{248--259}.
\newblock


\bibitem[\protect\citeauthoryear{Sen, Namata, Bilgic, Getoor, Galligher, and
  Eliassi-Rad}{Sen et~al\mbox{.}}{2008}]%
        {sen2008collective}
\bibfield{author}{\bibinfo{person}{Prithviraj Sen}, \bibinfo{person}{Galileo
  Namata}, \bibinfo{person}{Mustafa Bilgic}, \bibinfo{person}{Lise Getoor},
  \bibinfo{person}{Brian Galligher}, {and} \bibinfo{person}{Tina Eliassi-Rad}.}
  \bibinfo{year}{2008}\natexlab{}.
\newblock \showarticletitle{Collective classification in network data}.
\newblock \bibinfo{journal}{\emph{AI magazine}} \bibinfo{volume}{29},
  \bibinfo{number}{3} (\bibinfo{year}{2008}), \bibinfo{pages}{93--93}.
\newblock


\bibitem[\protect\citeauthoryear{SeongokRyu}{SeongokRyu}{2021}]%
        {src8}
\bibfield{author}{\bibinfo{person}{SeongokRyu}.}
  \bibinfo{year}{2021}\natexlab{}.
\newblock \bibinfo{title}{{Graph-neural-networks}}.
\newblock
\newblock
\urldef\tempurl%
\url{https://github.com/SeongokRyu/Graph-neural-networks}
\showURL{%
\tempurl}


\bibitem[\protect\citeauthoryear{Shchur, Mumme, Bojchevski, and
  G{\"u}nnemann}{Shchur et~al\mbox{.}}{2018}]%
        {shchur2018pitfalls}
\bibfield{author}{\bibinfo{person}{Oleksandr Shchur},
  \bibinfo{person}{Maximilian Mumme}, \bibinfo{person}{Aleksandar Bojchevski},
  {and} \bibinfo{person}{Stephan G{\"u}nnemann}.}
  \bibinfo{year}{2018}\natexlab{}.
\newblock \bibinfo{title}{Pitfalls of graph neural network evaluation}.
\newblock
\newblock


\bibitem[\protect\citeauthoryear{Spitzer}{Spitzer}{2013}]%
        {spitzer2013principles}
\bibfield{author}{\bibinfo{person}{Frank Spitzer}.}
  \bibinfo{year}{2013}\natexlab{}.
\newblock \bibinfo{booktitle}{\emph{Principles of random walk}}.
  Vol.~\bibinfo{volume}{34}.
\newblock \bibinfo{publisher}{Springer Science \& Business Media}.
\newblock


\bibitem[\protect\citeauthoryear{Subramonian}{Subramonian}{2021}]%
        {Subramonian}
\bibfield{author}{\bibinfo{person}{Arjun Subramonian}.}
  \bibinfo{year}{2021}\natexlab{}.
\newblock \showarticletitle{{MOTIF-Driven Contrastive Learning of Graph
  Representations}}.
\newblock \bibinfo{journal}{\emph{AAAI}} \bibinfo{volume}{35},
  \bibinfo{number}{18} (\bibinfo{date}{May} \bibinfo{year}{2021}),
  \bibinfo{pages}{15980--15981}.
\newblock
\showISSN{2374-3468}
\urldef\tempurl%
\url{https://ojs.aaai.org/index.php/AAAI/article/view/17986}
\showURL{%
\tempurl}


\bibitem[\protect\citeauthoryear{Sun, Hoffmann, Verma, and Tang}{Sun
  et~al\mbox{.}}{2019}]%
        {sun2019infograph}
\bibfield{author}{\bibinfo{person}{Fan-Yun Sun}, \bibinfo{person}{Jordan
  Hoffmann}, \bibinfo{person}{Vikas Verma}, {and} \bibinfo{person}{Jian Tang}.}
  \bibinfo{year}{2019}\natexlab{}.
\newblock \bibinfo{title}{Infograph: Unsupervised and semi-supervised
  graph-level representation learning via mutual information maximization}.
\newblock
\newblock


\bibitem[\protect\citeauthoryear{Sun, Lin, and Zhu}{Sun et~al\mbox{.}}{2020}]%
        {sun2020multi}
\bibfield{author}{\bibinfo{person}{Ke Sun}, \bibinfo{person}{Zhouchen Lin},
  {and} \bibinfo{person}{Zhanxing Zhu}.} \bibinfo{year}{2020}\natexlab{}.
\newblock \showarticletitle{Multi-stage self-supervised learning for graph
  convolutional networks on graphs with few labeled nodes}. In
  \bibinfo{booktitle}{\emph{Proceedings of the AAAI Conference on Artificial
  Intelligence}}, Vol.~\bibinfo{volume}{34}. \bibinfo{pages}{5892--5899}.
\newblock


\bibitem[\protect\citeauthoryear{Sun, Li, Peng, Wu, Ning, Yu, and He}{Sun
  et~al\mbox{.}}{2021}]%
        {sun2021sugar}
\bibfield{author}{\bibinfo{person}{Qingyun Sun}, \bibinfo{person}{Jianxin Li},
  \bibinfo{person}{Hao Peng}, \bibinfo{person}{Jia Wu},
  \bibinfo{person}{Yuanxing Ning}, \bibinfo{person}{Philip~S Yu}, {and}
  \bibinfo{person}{Lifang He}.} \bibinfo{year}{2021}\natexlab{}.
\newblock \showarticletitle{SUGAR: Subgraph neural network with reinforcement
  pooling and self-supervised mutual information mechanism}. In
  \bibinfo{booktitle}{\emph{Proceedings of the Web Conference 2021}}.
  \bibinfo{pages}{2081--2091}.
\newblock


\bibitem[\protect\citeauthoryear{svjan5}{svjan5}{2021}]%
        {src13}
\bibfield{author}{\bibinfo{person}{svjan5}.} \bibinfo{year}{2021}\natexlab{}.
\newblock \bibinfo{title}{{GNNs-for-NLP}}.
\newblock
\newblock
\urldef\tempurl%
\url{https://github.com/svjan5/GNNs-for-NLP}
\showURL{%
\tempurl}


\bibitem[\protect\citeauthoryear{Taheri, Gimpel, and Berger-Wolf}{Taheri
  et~al\mbox{.}}{2019}]%
        {taheri2018learning}
\bibfield{author}{\bibinfo{person}{Aynaz Taheri}, \bibinfo{person}{Kevin
  Gimpel}, {and} \bibinfo{person}{Tanya Berger-Wolf}.}
  \bibinfo{year}{2019}\natexlab{}.
\newblock \showarticletitle{Learning to Represent the Evolution of Dynamic
  Graphs with Recurrent Models}. In \bibinfo{booktitle}{\emph{Companion
  Proceedings of The 2019 World Wide Web Conference}} (San Francisco, USA)
  \emph{(\bibinfo{series}{WWW '19})}. \bibinfo{publisher}{Association for
  Computing Machinery}, \bibinfo{address}{New York, NY, USA},
  \bibinfo{pages}{301–307}.
\newblock
\showISBNx{9781450366755}
\urldef\tempurl%
\url{https://doi.org/10.1145/3308560.3316581}
\showDOI{\tempurl}


\bibitem[\protect\citeauthoryear{Tang and Liu}{Tang and Liu}{2012}]%
        {tang2012unsupervised}
\bibfield{author}{\bibinfo{person}{Jiliang Tang} {and} \bibinfo{person}{Huan
  Liu}.} \bibinfo{year}{2012}\natexlab{}.
\newblock \showarticletitle{Unsupervised feature selection for linked social
  media data}. In \bibinfo{booktitle}{\emph{Proceedings of the 18th ACM SIGKDD
  international conference on Knowledge discovery and data mining}}.
  \bibinfo{pages}{904--912}.
\newblock


\bibitem[\protect\citeauthoryear{Tang, Qu, and Mei}{Tang
  et~al\mbox{.}}{2015a}]%
        {tang2015pte}
\bibfield{author}{\bibinfo{person}{Jian Tang}, \bibinfo{person}{Meng Qu}, {and}
  \bibinfo{person}{Qiaozhu Mei}.} \bibinfo{year}{2015}\natexlab{a}.
\newblock \showarticletitle{Pte: Predictive text embedding through large-scale
  heterogeneous text networks}. In \bibinfo{booktitle}{\emph{Proceedings of the
  21th ACM SIGKDD international conference on knowledge discovery and data
  mining}}. \bibinfo{publisher}{Association for Computing Machinery},
  \bibinfo{address}{New York, NY, USA}, \bibinfo{pages}{1165--1174}.
\newblock


\bibitem[\protect\citeauthoryear{Tang, Qu, Wang, Zhang, Yan, and Mei}{Tang
  et~al\mbox{.}}{2015b}]%
        {tang2015line}
\bibfield{author}{\bibinfo{person}{Jian Tang}, \bibinfo{person}{Meng Qu},
  \bibinfo{person}{Mingzhe Wang}, \bibinfo{person}{Ming Zhang},
  \bibinfo{person}{Jun Yan}, {and} \bibinfo{person}{Qiaozhu Mei}.}
  \bibinfo{year}{2015}\natexlab{b}.
\newblock \showarticletitle{Line: Large-scale information network embedding}.
  In \bibinfo{booktitle}{\emph{Proceedings of the 24th international conference
  on world wide web}}. \bibinfo{publisher}{International World Wide Web
  Conferences Steering Committee}, \bibinfo{address}{Republic and Canton of
  Geneva, CHE}, \bibinfo{pages}{1067--1077}.
\newblock


\bibitem[\protect\citeauthoryear{Tang, Zhang, Yao, Li, Zhang, and Su}{Tang
  et~al\mbox{.}}{2008}]%
        {tang2008dblp}
\bibfield{author}{\bibinfo{person}{Jie Tang}, \bibinfo{person}{Jing Zhang},
  \bibinfo{person}{Limin Yao}, \bibinfo{person}{Juanzi Li}, \bibinfo{person}{Li
  Zhang}, {and} \bibinfo{person}{Zhong Su}.} \bibinfo{year}{2008}\natexlab{}.
\newblock \showarticletitle{ArnetMiner: Extraction and Mining of Academic
  Social Networks}. In \bibinfo{booktitle}{\emph{Proceedings of the 14th ACM
  SIGKDD International Conference on Knowledge Discovery and Data Mining}} (Las
  Vegas, Nevada, USA) \emph{(\bibinfo{series}{KDD '08})}.
  \bibinfo{publisher}{Association for Computing Machinery},
  \bibinfo{address}{New York, NY, USA}, \bibinfo{pages}{990–998}.
\newblock
\showISBNx{9781605581934}
\urldef\tempurl%
\url{https://doi.org/10.1145/1401890.1402008}
\showDOI{\tempurl}


\bibitem[\protect\citeauthoryear{Thudm}{Thudm}{2021}]%
        {src7}
\bibfield{author}{\bibinfo{person}{Thudm}.} \bibinfo{year}{2021}\natexlab{}.
\newblock \bibinfo{title}{{cogdl}}.
\newblock
\newblock
\urldef\tempurl%
\url{https://github.com/THUDM/cogdl}
\showURL{%
\tempurl}


\bibitem[\protect\citeauthoryear{thunlp}{thunlp}{2021}]%
        {src6}
\bibfield{author}{\bibinfo{person}{thunlp}.} \bibinfo{year}{2021}\natexlab{}.
\newblock \bibinfo{title}{{OpenNE}}.
\newblock
\newblock
\urldef\tempurl%
\url{https://github.com/thunlp/OpenNE/tree/pytorch}
\showURL{%
\tempurl}


\bibitem[\protect\citeauthoryear{Tu, Cui, Wang, Yu, and Zhu}{Tu
  et~al\mbox{.}}{2018}]%
        {tu2018deep}
\bibfield{author}{\bibinfo{person}{Ke Tu}, \bibinfo{person}{Peng Cui},
  \bibinfo{person}{Xiao Wang}, \bibinfo{person}{Philip~S Yu}, {and}
  \bibinfo{person}{Wenwu Zhu}.} \bibinfo{year}{2018}\natexlab{}.
\newblock \showarticletitle{Deep recursive network embedding with regular
  equivalence}. In \bibinfo{booktitle}{\emph{Proceedings of the 24th ACM SIGKDD
  international conference on knowledge discovery \& data mining}}.
  \bibinfo{publisher}{Association for Computing Machinery},
  \bibinfo{address}{New York, NY, USA}, \bibinfo{pages}{2357--2366}.
\newblock


\bibitem[\protect\citeauthoryear{Van~Engelen and Hoos}{Van~Engelen and
  Hoos}{2020}]%
        {van2020survey}
\bibfield{author}{\bibinfo{person}{Jesper~E Van~Engelen} {and}
  \bibinfo{person}{Holger~H Hoos}.} \bibinfo{year}{2020}\natexlab{}.
\newblock \showarticletitle{A survey on semi-supervised learning}.
\newblock \bibinfo{journal}{\emph{Machine Learning}} \bibinfo{volume}{109},
  \bibinfo{number}{2} (\bibinfo{year}{2020}), \bibinfo{pages}{373--440}.
\newblock


\bibitem[\protect\citeauthoryear{Velickovic, Fedus, Hamilton, Li{\`o}, Bengio,
  and Hjelm}{Velickovic et~al\mbox{.}}{2019}]%
        {velickovic2019deep}
\bibfield{author}{\bibinfo{person}{Petar Velickovic}, \bibinfo{person}{William
  Fedus}, \bibinfo{person}{William~L Hamilton}, \bibinfo{person}{Pietro
  Li{\`o}}, \bibinfo{person}{Yoshua Bengio}, {and} \bibinfo{person}{R~Devon
  Hjelm}.} \bibinfo{year}{2019}\natexlab{}.
\newblock \showarticletitle{Deep Graph Infomax.}
\newblock \bibinfo{journal}{\emph{ICLR (Poster)}} \bibinfo{volume}{2},
  \bibinfo{number}{3} (\bibinfo{year}{2019}), \bibinfo{pages}{4}.
\newblock


\bibitem[\protect\citeauthoryear{Verma and Zhang}{Verma and Zhang}{2019}]%
        {verma2019stability}
\bibfield{author}{\bibinfo{person}{Saurabh Verma} {and} \bibinfo{person}{Zhi-Li
  Zhang}.} \bibinfo{year}{2019}\natexlab{}.
\newblock \showarticletitle{Stability and generalization of graph convolutional
  neural networks}. In \bibinfo{booktitle}{\emph{Proceedings of the 25th ACM
  SIGKDD International Conference on Knowledge Discovery \& Data Mining}}.
  \bibinfo{publisher}{Association for Computing Machinery},
  \bibinfo{address}{New York, NY, USA}, \bibinfo{pages}{1539--1548}.
\newblock


\bibitem[\protect\citeauthoryear{Wan, Pan, Yang, and Gong}{Wan
  et~al\mbox{.}}{2020}]%
        {wan2020contrastive}
\bibfield{author}{\bibinfo{person}{Sheng Wan}, \bibinfo{person}{Shirui Pan},
  \bibinfo{person}{Jian Yang}, {and} \bibinfo{person}{Chen Gong}.}
  \bibinfo{year}{2020}\natexlab{}.
\newblock \bibinfo{title}{Contrastive and generative graph convolutional
  networks for graph-based semi-supervised learning}.
\newblock
\newblock


\bibitem[\protect\citeauthoryear{Wang, Pan, Long, Zhu, and Jiang}{Wang
  et~al\mbox{.}}{2017}]%
        {wang2017mgae}
\bibfield{author}{\bibinfo{person}{Chun Wang}, \bibinfo{person}{Shirui Pan},
  \bibinfo{person}{Guodong Long}, \bibinfo{person}{Xingquan Zhu}, {and}
  \bibinfo{person}{Jing Jiang}.} \bibinfo{year}{2017}\natexlab{}.
\newblock \showarticletitle{Mgae: Marginalized graph autoencoder for graph
  clustering}. In \bibinfo{booktitle}{\emph{Proceedings of the 2017 ACM on
  Conference on Information and Knowledge Management}}.
  \bibinfo{pages}{889--898}.
\newblock


\bibitem[\protect\citeauthoryear{Wang, Cui, and Zhu}{Wang
  et~al\mbox{.}}{2016}]%
        {wang2016structural}
\bibfield{author}{\bibinfo{person}{Daixin Wang}, \bibinfo{person}{Peng Cui},
  {and} \bibinfo{person}{Wenwu Zhu}.} \bibinfo{year}{2016}\natexlab{}.
\newblock \showarticletitle{Structural deep network embedding}. In
  \bibinfo{booktitle}{\emph{Proceedings of the 22nd ACM SIGKDD international
  conference on Knowledge discovery and data mining}}.
  \bibinfo{publisher}{Association for Computing Machinery},
  \bibinfo{address}{New York, NY, USA}, \bibinfo{pages}{1225--1234}.
\newblock


\bibitem[\protect\citeauthoryear{Wang, Xu, Liu, Lian, Chen, Du, Wu, and
  Su}{Wang et~al\mbox{.}}{2019b}]%
        {wang2019mcne}
\bibfield{author}{\bibinfo{person}{Hao Wang}, \bibinfo{person}{Tong Xu},
  \bibinfo{person}{Qi Liu}, \bibinfo{person}{Defu Lian},
  \bibinfo{person}{Enhong Chen}, \bibinfo{person}{Dongfang Du},
  \bibinfo{person}{Han Wu}, {and} \bibinfo{person}{Wen Su}.}
  \bibinfo{year}{2019}\natexlab{b}.
\newblock \showarticletitle{MCNE: an end-to-end framework for learning multiple
  conditional network representations of social network}. In
  \bibinfo{booktitle}{\emph{Proceedings of the 25th ACM SIGKDD International
  Conference on Knowledge Discovery \& Data Mining}}.
  \bibinfo{publisher}{Association for Computing Machinery},
  \bibinfo{address}{New York, NY, USA}, \bibinfo{pages}{1064--1072}.
\newblock


\bibitem[\protect\citeauthoryear{Wang, Zheng, Ye, Gan, Li, Song, Zhou, Ma, Yu,
  Gai, Xiao, He, Karypis, Li, and Zhang}{Wang et~al\mbox{.}}{2021b}]%
        {src5}
\bibfield{author}{\bibinfo{person}{Minjie Wang}, \bibinfo{person}{Da Zheng},
  \bibinfo{person}{Zihao Ye}, \bibinfo{person}{Quan Gan},
  \bibinfo{person}{Mufei Li}, \bibinfo{person}{Xiang Song},
  \bibinfo{person}{Jinjing Zhou}, \bibinfo{person}{Chao Ma},
  \bibinfo{person}{Lingfan Yu}, \bibinfo{person}{Yu Gai},
  \bibinfo{person}{Tianjun Xiao}, \bibinfo{person}{Tong He},
  \bibinfo{person}{George Karypis}, \bibinfo{person}{Jinyang Li}, {and}
  \bibinfo{person}{Zheng Zhang}.} \bibinfo{year}{2021}\natexlab{b}.
\newblock \bibinfo{title}{{DGL}}.
\newblock
\newblock
\urldef\tempurl%
\url{https://github.com/dmlc/dgl}
\showURL{%
\tempurl}


\bibitem[\protect\citeauthoryear{Wang, Agarwal, Ham, Choudhury, and Reddy}{Wang
  et~al\mbox{.}}{2021a}]%
        {wang2021self}
\bibfield{author}{\bibinfo{person}{Ping Wang}, \bibinfo{person}{Khushbu
  Agarwal}, \bibinfo{person}{Colby Ham}, \bibinfo{person}{Sutanay Choudhury},
  {and} \bibinfo{person}{Chandan~K Reddy}.} \bibinfo{year}{2021}\natexlab{a}.
\newblock \showarticletitle{Self-Supervised Learning of Contextual Embeddings
  for Link Prediction in Heterogeneous Networks}. In
  \bibinfo{booktitle}{\emph{Proceedings of the Web Conference 2021}}.
  \bibinfo{pages}{2946--2957}.
\newblock


\bibitem[\protect\citeauthoryear{Wang, Fass, Stern, Luo, and Chodera}{Wang
  et~al\mbox{.}}{2019a}]%
        {wang2019graph}
\bibfield{author}{\bibinfo{person}{Yuanqing Wang}, \bibinfo{person}{Josh Fass},
  \bibinfo{person}{Chaya~D Stern}, \bibinfo{person}{Kun Luo}, {and}
  \bibinfo{person}{John Chodera}.} \bibinfo{year}{2019}\natexlab{a}.
\newblock \showarticletitle{Graph nets for partial charge prediction}.
\newblock \bibinfo{journal}{\emph{arXiv preprint arXiv:1909.07903}}
  (\bibinfo{year}{2019}).
\newblock


\bibitem[\protect\citeauthoryear{Wang, Jiang, Ren, Tang, and Yin}{Wang
  et~al\mbox{.}}{2018}]%
        {wang2018path}
\bibfield{author}{\bibinfo{person}{Zihan Wang}, \bibinfo{person}{Ziheng Jiang},
  \bibinfo{person}{Zhaochun Ren}, \bibinfo{person}{Jiliang Tang}, {and}
  \bibinfo{person}{Dawei Yin}.} \bibinfo{year}{2018}\natexlab{}.
\newblock \showarticletitle{A path-constrained framework for discriminating
  substitutable and complementary products in e-commerce}. In
  \bibinfo{booktitle}{\emph{Proceedings of the Eleventh ACM International
  Conference on Web Search and Data Mining}}. \bibinfo{pages}{619--627}.
\newblock


\bibitem[\protect\citeauthoryear{Wink, Tijms, Ten~Kate, Raspor, de~Munck,
  Altena, Ecay-Torres, Clerigue, Estanga, Garcia-Sebastian, et~al\mbox{.}}{Wink
  et~al\mbox{.}}{2018}]%
        {wink2018functional}
\bibfield{author}{\bibinfo{person}{Alle~Meije Wink}, \bibinfo{person}{Betty~M
  Tijms}, \bibinfo{person}{Mara Ten~Kate}, \bibinfo{person}{Eva Raspor},
  \bibinfo{person}{Jan~C de Munck}, \bibinfo{person}{Ellemarije Altena},
  \bibinfo{person}{Mirian Ecay-Torres}, \bibinfo{person}{Montserrat Clerigue},
  \bibinfo{person}{Ainara Estanga}, \bibinfo{person}{Maite Garcia-Sebastian},
  {et~al\mbox{.}}} \bibinfo{year}{2018}\natexlab{}.
\newblock \showarticletitle{Functional brain network centrality is related to
  APOE genotype in cognitively normal elderly}.
\newblock \bibinfo{journal}{\emph{Brain and behavior}} \bibinfo{volume}{8},
  \bibinfo{number}{9} (\bibinfo{year}{2018}), \bibinfo{pages}{e01080}.
\newblock


\bibitem[\protect\citeauthoryear{Wu, Souza, Zhang, Fifty, Yu, and
  Weinberger}{Wu et~al\mbox{.}}{2019}]%
        {wu2019simplifying}
\bibfield{author}{\bibinfo{person}{Felix Wu}, \bibinfo{person}{Amauri Souza},
  \bibinfo{person}{Tianyi Zhang}, \bibinfo{person}{Christopher Fifty},
  \bibinfo{person}{Tao Yu}, {and} \bibinfo{person}{Kilian Weinberger}.}
  \bibinfo{year}{2019}\natexlab{}.
\newblock \showarticletitle{Simplifying graph convolutional networks}.
\newblock \bibinfo{journal}{\emph{Proceedings of the 36th International
  Conference on Machine Learning}}  \bibinfo{volume}{97}
  (\bibinfo{year}{2019}), \bibinfo{pages}{6861--6871}.
\newblock


\bibitem[\protect\citeauthoryear{Wu, Wang, Feng, He, Chen, Lian, and Xie}{Wu
  et~al\mbox{.}}{2021b}]%
        {wu2021self}
\bibfield{author}{\bibinfo{person}{Jiancan Wu}, \bibinfo{person}{Xiang Wang},
  \bibinfo{person}{Fuli Feng}, \bibinfo{person}{Xiangnan He},
  \bibinfo{person}{Liang Chen}, \bibinfo{person}{Jianxun Lian}, {and}
  \bibinfo{person}{Xing Xie}.} \bibinfo{year}{2021}\natexlab{b}.
\newblock \showarticletitle{Self-supervised graph learning for recommendation}.
  In \bibinfo{booktitle}{\emph{Proceedings of the 44th International ACM SIGIR
  Conference on Research and Development in Information Retrieval}}.
  \bibinfo{pages}{726--735}.
\newblock


\bibitem[\protect\citeauthoryear{Wu and Cheng}{Wu and Cheng}{2021}]%
        {wu2021deepened}
\bibfield{author}{\bibinfo{person}{Xinxing Wu} {and} \bibinfo{person}{Qiang
  Cheng}.} \bibinfo{year}{2021}\natexlab{}.
\newblock \bibinfo{title}{Deepened Graph Auto-Encoders Help Stabilize and
  Enhance Link Prediction}.
\newblock
\newblock


\bibitem[\protect\citeauthoryear{Wu, Pan, Chen, Long, Zhang, and Yu}{Wu
  et~al\mbox{.}}{2021a}]%
        {Wu}
\bibfield{author}{\bibinfo{person}{Zonghan Wu}, \bibinfo{person}{Shirui Pan},
  \bibinfo{person}{Fengwen Chen}, \bibinfo{person}{Guodong Long},
  \bibinfo{person}{Chengqi Zhang}, {and} \bibinfo{person}{Philip~S. Yu}.}
  \bibinfo{year}{2021}\natexlab{a}.
\newblock \showarticletitle{A Comprehensive Survey on Graph Neural Networks}.
\newblock \bibinfo{journal}{\emph{IEEE Transactions on Neural Networks and
  Learning Systems}} \bibinfo{volume}{32}, \bibinfo{number}{1}
  (\bibinfo{year}{2021}), \bibinfo{pages}{4--24}.
\newblock
\urldef\tempurl%
\url{https://doi.org/10.1109/TNNLS.2020.2978386}
\showDOI{\tempurl}


\bibitem[\protect\citeauthoryear{Xu, Shen, Cao, Qiu, and Cheng}{Xu
  et~al\mbox{.}}{2019}]%
        {xu2019graph}
\bibfield{author}{\bibinfo{person}{Bingbing Xu}, \bibinfo{person}{Huawei Shen},
  \bibinfo{person}{Qi Cao}, \bibinfo{person}{Yunqi Qiu}, {and}
  \bibinfo{person}{Xueqi Cheng}.} \bibinfo{year}{2019}\natexlab{}.
\newblock \bibinfo{title}{Graph wavelet neural network}.
\newblock
\newblock


\bibitem[\protect\citeauthoryear{Xu, Hu, Leskovec, and Jegelka}{Xu
  et~al\mbox{.}}{2018}]%
        {xu2018powerful}
\bibfield{author}{\bibinfo{person}{Keyulu Xu}, \bibinfo{person}{Weihua Hu},
  \bibinfo{person}{Jure Leskovec}, {and} \bibinfo{person}{Stefanie Jegelka}.}
  \bibinfo{year}{2018}\natexlab{}.
\newblock \bibinfo{title}{How powerful are graph neural networks?}
\newblock
\newblock


\bibitem[\protect\citeauthoryear{Xu, Li, Yu, Ge, Shi, Li, Liang, Lin, Zhang,
  and Shao}{Xu et~al\mbox{.}}{2020}]%
        {xu2020altered}
\bibfield{author}{\bibinfo{person}{Qian-Hui Xu}, \bibinfo{person}{Qiu-Yu Li},
  \bibinfo{person}{Kang Yu}, \bibinfo{person}{Qian-Ming Ge},
  \bibinfo{person}{Wen-Qing Shi}, \bibinfo{person}{Biao Li},
  \bibinfo{person}{Rong-Bin Liang}, \bibinfo{person}{Qi Lin},
  \bibinfo{person}{Yu-Qing Zhang}, {and} \bibinfo{person}{Yi Shao}.}
  \bibinfo{year}{2020}\natexlab{}.
\newblock \showarticletitle{Altered Brain Network Centrality in Patients with
  Diabetic Optic Neuropathy: A Resting-State FMRI Study}.
\newblock \bibinfo{journal}{\emph{Endocrine Practice}} \bibinfo{volume}{26},
  \bibinfo{number}{12} (\bibinfo{year}{2020}), \bibinfo{pages}{1399--1405}.
\newblock


\bibitem[\protect\citeauthoryear{Yang and Leskovec}{Yang and Leskovec}{2015}]%
        {yang2015defining}
\bibfield{author}{\bibinfo{person}{Jaewon Yang} {and} \bibinfo{person}{Jure
  Leskovec}.} \bibinfo{year}{2015}\natexlab{}.
\newblock \showarticletitle{Defining and evaluating network communities based
  on ground-truth}.
\newblock \bibinfo{journal}{\emph{Knowledge and Information Systems}}
  \bibinfo{volume}{42}, \bibinfo{number}{1} (\bibinfo{year}{2015}),
  \bibinfo{pages}{181--213}.
\newblock


\bibitem[\protect\citeauthoryear{Yang, Gu, Wang, Cao, Zhai, Jin, and Guo}{Yang
  et~al\mbox{.}}{2020}]%
        {yang2020toward}
\bibfield{author}{\bibinfo{person}{Liang Yang}, \bibinfo{person}{Junhua Gu},
  \bibinfo{person}{Chuan Wang}, \bibinfo{person}{Xiaochun Cao},
  \bibinfo{person}{Lu Zhai}, \bibinfo{person}{Di Jin}, {and}
  \bibinfo{person}{Yuanfang Guo}.} \bibinfo{year}{2020}\natexlab{}.
\newblock \showarticletitle{Toward Unsupervised Graph Neural Network:
  Interactive Clustering and Embedding via Optimal Transport}. In
  \bibinfo{booktitle}{\emph{2020 IEEE International Conference on Data Mining
  (ICDM)}}. IEEE, \bibinfo{pages}{1358--1363}.
\newblock


\bibitem[\protect\citeauthoryear{Yang, Cohen, and Salakhudinov}{Yang
  et~al\mbox{.}}{2016}]%
        {yang2016revisiting}
\bibfield{author}{\bibinfo{person}{Zhilin Yang}, \bibinfo{person}{William
  Cohen}, {and} \bibinfo{person}{Ruslan Salakhudinov}.}
  \bibinfo{year}{2016}\natexlab{}.
\newblock \showarticletitle{Revisiting semi-supervised learning with graph
  embeddings}.
\newblock \bibinfo{journal}{\emph{Proceedings of The 33rd International
  Conference on Machine Learning}}  \bibinfo{volume}{48}
  (\bibinfo{year}{2016}), \bibinfo{pages}{40--48}.
\newblock


\bibitem[\protect\citeauthoryear{Yao, Mao, and Luo}{Yao et~al\mbox{.}}{2019}]%
        {yao2019graph}
\bibfield{author}{\bibinfo{person}{Liang Yao}, \bibinfo{person}{Chengsheng
  Mao}, {and} \bibinfo{person}{Yuan Luo}.} \bibinfo{year}{2019}\natexlab{}.
\newblock \showarticletitle{Graph convolutional networks for text
  classification}. In \bibinfo{booktitle}{\emph{Proceedings of the AAAI
  conference on artificial intelligence}}, Vol.~\bibinfo{volume}{33}.
  \bibinfo{pages}{7370--7377}.
\newblock


\bibitem[\protect\citeauthoryear{Ying, He, Chen, Eksombatchai, Hamilton, and
  Leskovec}{Ying et~al\mbox{.}}{2018}]%
        {ying2018graph}
\bibfield{author}{\bibinfo{person}{Rex Ying}, \bibinfo{person}{Ruining He},
  \bibinfo{person}{Kaifeng Chen}, \bibinfo{person}{Pong Eksombatchai},
  \bibinfo{person}{William~L Hamilton}, {and} \bibinfo{person}{Jure Leskovec}.}
  \bibinfo{year}{2018}\natexlab{}.
\newblock \showarticletitle{Graph convolutional neural networks for web-scale
  recommender systems}. In \bibinfo{booktitle}{\emph{Proceedings of the 24th
  ACM SIGKDD International Conference on Knowledge Discovery \& Data Mining}}.
  \bibinfo{publisher}{Association for Computing Machinery},
  \bibinfo{address}{New York, NY, USA}, \bibinfo{pages}{974--983}.
\newblock


\bibitem[\protect\citeauthoryear{You, Ying, and Leskovec}{You
  et~al\mbox{.}}{2020b}]%
        {you2020design}
\bibfield{author}{\bibinfo{person}{Jiaxuan You}, \bibinfo{person}{Zhitao Ying},
  {and} \bibinfo{person}{Jure Leskovec}.} \bibinfo{year}{2020}\natexlab{b}.
\newblock \showarticletitle{Design space for graph neural networks}.
\newblock \bibinfo{journal}{\emph{Advances in Neural Information Processing
  Systems}}  \bibinfo{volume}{33} (\bibinfo{year}{2020}).
\newblock


\bibitem[\protect\citeauthoryear{You, Chen, Wang, and Shen}{You
  et~al\mbox{.}}{2020a}]%
        {you2020does}
\bibfield{author}{\bibinfo{person}{Yuning You}, \bibinfo{person}{Tianlong
  Chen}, \bibinfo{person}{Zhangyang Wang}, {and} \bibinfo{person}{Yang Shen}.}
  \bibinfo{year}{2020}\natexlab{a}.
\newblock \showarticletitle{When does self-supervision help graph convolutional
  networks?}
\newblock \bibinfo{journal}{\emph{Proceedings of the 37th International
  Conference on Machine Learning}}  \bibinfo{volume}{119}
  (\bibinfo{year}{2020}), \bibinfo{pages}{10871--10880}.
\newblock


\bibitem[\protect\citeauthoryear{Yu, Yin, and Zhu}{Yu et~al\mbox{.}}{2017}]%
        {yu2017spatio}
\bibfield{author}{\bibinfo{person}{Bing Yu}, \bibinfo{person}{Haoteng Yin},
  {and} \bibinfo{person}{Zhanxing Zhu}.} \bibinfo{year}{2017}\natexlab{}.
\newblock \bibinfo{title}{Spatio-temporal graph convolutional networks: A deep
  learning framework for traffic forecasting}.
\newblock
\newblock


\bibitem[\protect\citeauthoryear{Yu, Lin, Yang, Shen, Lu, and Huang}{Yu
  et~al\mbox{.}}{2018}]%
        {yu2018generative}
\bibfield{author}{\bibinfo{person}{Jiahui Yu}, \bibinfo{person}{Zhe Lin},
  \bibinfo{person}{Jimei Yang}, \bibinfo{person}{Xiaohui Shen},
  \bibinfo{person}{Xin Lu}, {and} \bibinfo{person}{Thomas~S Huang}.}
  \bibinfo{year}{2018}\natexlab{}.
\newblock \showarticletitle{Generative image inpainting with contextual
  attention}. In \bibinfo{booktitle}{\emph{Proceedings of the IEEE conference
  on computer vision and pattern recognition}}. \bibinfo{publisher}{IEEE},
  \bibinfo{address}{Salt Lake City, UT, USA}, \bibinfo{pages}{5505--5514}.
\newblock


\bibitem[\protect\citeauthoryear{Zachary}{Zachary}{1977}]%
        {zachary1977information}
\bibfield{author}{\bibinfo{person}{Wayne~W Zachary}.}
  \bibinfo{year}{1977}\natexlab{}.
\newblock \showarticletitle{An information flow model for conflict and fission
  in small groups}.
\newblock \bibinfo{journal}{\emph{Journal of anthropological research}}
  (\bibinfo{year}{1977}), \bibinfo{pages}{452--473}.
\newblock


\bibitem[\protect\citeauthoryear{Zeng and Xie}{Zeng and Xie}{2020}]%
        {zeng2020contrastive}
\bibfield{author}{\bibinfo{person}{Jiaqi Zeng} {and} \bibinfo{person}{Pengtao
  Xie}.} \bibinfo{year}{2020}\natexlab{}.
\newblock \bibinfo{title}{Contrastive self-supervised learning for graph
  classification}.
\newblock
\newblock


\bibitem[\protect\citeauthoryear{Zhang, Yu, and Zhang}{Zhang
  et~al\mbox{.}}{2020c}]%
        {zhang2020community}
\bibfield{author}{\bibinfo{person}{Binbin Zhang}, \bibinfo{person}{Zhizhi Yu},
  {and} \bibinfo{person}{Weixiong Zhang}.} \bibinfo{year}{2020}\natexlab{c}.
\newblock \showarticletitle{Community-centric graph convolutional network for
  unsupervised community detection}. In \bibinfo{booktitle}{\emph{IJCAI}}.
  \bibinfo{pages}{551--556}.
\newblock


\bibitem[\protect\citeauthoryear{Zhang, Lin, Liu, Zhou, Tang, Liang, and
  Xing}{Zhang et~al\mbox{.}}{2020a}]%
        {zhang2020iterative}
\bibfield{author}{\bibinfo{person}{Hanlin Zhang}, \bibinfo{person}{Shuai Lin},
  \bibinfo{person}{Weiyang Liu}, \bibinfo{person}{Pan Zhou},
  \bibinfo{person}{Jian Tang}, \bibinfo{person}{Xiaodan Liang}, {and}
  \bibinfo{person}{Eric~P Xing}.} \bibinfo{year}{2020}\natexlab{a}.
\newblock \bibinfo{title}{Iterative graph self-distillation}.
\newblock
\newblock


\bibitem[\protect\citeauthoryear{Zhang, Shi, Xie, Ma, King, and Yeung}{Zhang
  et~al\mbox{.}}{2018}]%
        {zhang2018gaan}
\bibfield{author}{\bibinfo{person}{Jiani Zhang}, \bibinfo{person}{Xingjian
  Shi}, \bibinfo{person}{Junyuan Xie}, \bibinfo{person}{Hao Ma},
  \bibinfo{person}{Irwin King}, {and} \bibinfo{person}{Dit-Yan Yeung}.}
  \bibinfo{year}{2018}\natexlab{}.
\newblock \bibinfo{title}{Gaan: Gated attention networks for learning on large
  and spatiotemporal graphs}.
\newblock
\newblock


\bibitem[\protect\citeauthoryear{Zhang, Zhang, Xia, and Sun}{Zhang
  et~al\mbox{.}}{2020d}]%
        {zhang2020graph}
\bibfield{author}{\bibinfo{person}{Jiawei Zhang}, \bibinfo{person}{Haopeng
  Zhang}, \bibinfo{person}{Congying Xia}, {and} \bibinfo{person}{Li Sun}.}
  \bibinfo{year}{2020}\natexlab{d}.
\newblock \bibinfo{title}{Graph-bert: Only attention is needed for learning
  graph representations}.
\newblock
\newblock


\bibitem[\protect\citeauthoryear{Zhang, Wang, Li, Zhu, Shen, Li, Lu, Shah, and
  Bennamoun}{Zhang et~al\mbox{.}}{2020b}]%
        {zhang2020structure}
\bibfield{author}{\bibinfo{person}{Liang Zhang}, \bibinfo{person}{Xudong Wang},
  \bibinfo{person}{Hongsheng Li}, \bibinfo{person}{Guangming Zhu},
  \bibinfo{person}{Peiyi Shen}, \bibinfo{person}{Ping Li},
  \bibinfo{person}{Xiaoyuan Lu}, \bibinfo{person}{Syed Afaq~Ali Shah}, {and}
  \bibinfo{person}{Mohammed Bennamoun}.} \bibinfo{year}{2020}\natexlab{b}.
\newblock \showarticletitle{Structure-feature based graph self-adaptive
  pooling}. In \bibinfo{booktitle}{\emph{Proceedings of The Web Conference
  2020}}. \bibinfo{pages}{3098--3104}.
\newblock


\bibitem[\protect\citeauthoryear{Zhang and Chen}{Zhang and Chen}{2018}]%
        {zhang2018link}
\bibfield{author}{\bibinfo{person}{Muhan Zhang} {and} \bibinfo{person}{Yixin
  Chen}.} \bibinfo{year}{2018}\natexlab{}.
\newblock \showarticletitle{Link prediction based on graph neural networks}.
\newblock \bibinfo{journal}{\emph{Advances in Neural Information Processing
  Systems}}  \bibinfo{volume}{31} (\bibinfo{year}{2018}),
  \bibinfo{pages}{5165--5175}.
\newblock


\bibitem[\protect\citeauthoryear{Zhang, Liu, Li, and Wu}{Zhang
  et~al\mbox{.}}{2019}]%
        {zhang2019attributed}
\bibfield{author}{\bibinfo{person}{Xiaotong Zhang}, \bibinfo{person}{Han Liu},
  \bibinfo{person}{Qimai Li}, {and} \bibinfo{person}{Xiao-Ming Wu}.}
  \bibinfo{year}{2019}\natexlab{}.
\newblock \bibinfo{title}{Attributed graph clustering via adaptive graph
  convolution}.
\newblock
\newblock


\bibitem[\protect\citeauthoryear{Zhou, Cui, Hu, Zhang, Yang, Liu, Wang, Li, and
  Sun}{Zhou et~al\mbox{.}}{2020}]%
        {zhou2020graph}
\bibfield{author}{\bibinfo{person}{Jie Zhou}, \bibinfo{person}{Ganqu Cui},
  \bibinfo{person}{Shengding Hu}, \bibinfo{person}{Zhengyan Zhang},
  \bibinfo{person}{Cheng Yang}, \bibinfo{person}{Zhiyuan Liu},
  \bibinfo{person}{Lifeng Wang}, \bibinfo{person}{Changcheng Li}, {and}
  \bibinfo{person}{Maosong Sun}.} \bibinfo{year}{2020}\natexlab{}.
\newblock \showarticletitle{Graph neural networks: A review of methods and
  applications}.
\newblock \bibinfo{journal}{\emph{AI Open}}  \bibinfo{volume}{1}
  (\bibinfo{year}{2020}), \bibinfo{pages}{57--81}.
\newblock


\bibitem[\protect\citeauthoryear{Zhu, Xu, Yu, Liu, Wu, and Wang}{Zhu
  et~al\mbox{.}}{2020}]%
        {zhu2020deep}
\bibfield{author}{\bibinfo{person}{Yanqiao Zhu}, \bibinfo{person}{Yichen Xu},
  \bibinfo{person}{Feng Yu}, \bibinfo{person}{Qiang Liu}, \bibinfo{person}{Shu
  Wu}, {and} \bibinfo{person}{Liang Wang}.} \bibinfo{year}{2020}\natexlab{}.
\newblock \bibinfo{title}{Deep graph contrastive representation learning}.
\newblock
\newblock


\bibitem[\protect\citeauthoryear{Zhu, Xu, Yu, Liu, Wu, and Wang}{Zhu
  et~al\mbox{.}}{2021}]%
        {zhu2021graph}
\bibfield{author}{\bibinfo{person}{Yanqiao Zhu}, \bibinfo{person}{Yichen Xu},
  \bibinfo{person}{Feng Yu}, \bibinfo{person}{Qiang Liu}, \bibinfo{person}{Shu
  Wu}, {and} \bibinfo{person}{Liang Wang}.} \bibinfo{year}{2021}\natexlab{}.
\newblock \showarticletitle{Graph contrastive learning with adaptive
  augmentation}. In \bibinfo{booktitle}{\emph{Proceedings of the Web Conference
  2021}}. \bibinfo{pages}{2069--2080}.
\newblock


\bibitem[\protect\citeauthoryear{Zhuang, Zhou, Gao, Yin, Lin, and Ma}{Zhuang
  et~al\mbox{.}}{2017}]%
        {zhuang2017label}
\bibfield{author}{\bibinfo{person}{Liansheng Zhuang}, \bibinfo{person}{Zihan
  Zhou}, \bibinfo{person}{Shenghua Gao}, \bibinfo{person}{Jingwen Yin},
  \bibinfo{person}{Zhouchen Lin}, {and} \bibinfo{person}{Yi Ma}.}
  \bibinfo{year}{2017}\natexlab{}.
\newblock \showarticletitle{Label information guided graph construction for
  semi-supervised learning}.
\newblock \bibinfo{journal}{\emph{IEEE Transactions on Image Processing}}
  \bibinfo{volume}{26}, \bibinfo{number}{9} (\bibinfo{year}{2017}),
  \bibinfo{pages}{4182--4192}.
\newblock


\bibitem[\protect\citeauthoryear{Zuo, Ehmke, Mennes, Imperati, Castellanos,
  Sporns, and Milham}{Zuo et~al\mbox{.}}{2012}]%
        {zuo2012network}
\bibfield{author}{\bibinfo{person}{Xi-Nian Zuo}, \bibinfo{person}{Ross Ehmke},
  \bibinfo{person}{Maarten Mennes}, \bibinfo{person}{Davide Imperati},
  \bibinfo{person}{F~Xavier Castellanos}, \bibinfo{person}{Olaf Sporns}, {and}
  \bibinfo{person}{Michael~P Milham}.} \bibinfo{year}{2012}\natexlab{}.
\newblock \showarticletitle{Network centrality in the human functional
  connectome}.
\newblock \bibinfo{journal}{\emph{Cerebral cortex}} \bibinfo{volume}{22},
  \bibinfo{number}{8} (\bibinfo{year}{2012}), \bibinfo{pages}{1862--1875}.
\newblock


\end{thebibliography}

\end{document}